\documentclass[acmsmall,screen]{acmart}

\AtBeginDocument{%
  \providecommand\BibTeX{{%
    \normalfont B\kern-0.5em{\scshape i\kern-0.25em b}\kern-0.8em\TeX}}}

\setcopyright{acmcopyright}
\copyrightyear{2022}
\acmYear{2022}
\acmDOI{10.1145/1122445.1122456}

\usepackage{graphicx} 

\usepackage{amssymb}
\usepackage{longtable}
\usepackage{setspace}
\usepackage{url}

\usepackage{bm}
\usepackage{tabularx}
\usepackage{xcolor}
\usepackage{multirow}
\usepackage{subfigure}
\usepackage{colortbl}
	
\definecolor{Gray}{gray}{0.9}

\usepackage{CJKutf8}

\usepackage{extarrows}
\usepackage{hyperref}
\usepackage{url}
\usepackage{booktabs}
\usepackage{enumitem}

\acmJournal{TWEB}
\acmVolume{37}
\acmNumber{4}
\acmMonth{4}

\begin{document}

\title{Type Information Utilized Event Detection via Multi-Channel GNNs in Electrical Power Systems}

\author{Qian Li}
\affiliation{%
  \institution{School of Computer Science and Engineering, Beihang University}
  \city{Haidian}
  \state{Beijing}
  \country{China}}
\email{liqian@act.buaa.edu.cn}

\author{Jianxin Li}
\authornote{Corresponding author}
\affiliation{%
  \institution{School of Computer Science and Engineering, Beihang University}
  \city{Haidian}
  \state{Beijing}
  \country{China}}
\email{lijx@act.buaa.edu.cn}

\author{Lihong Wang}
\affiliation{%
  \institution{National Computer Network Emergency Response Technical Team/Coordination Center of China}
  \state{Beijing}
  \country{China}}
\email{wlh@isc.org.cn}

\author{Cheng Ji}
\affiliation{%
  \institution{School of Computer Science and Engineering, Beihang University}
  \city{Haidian}
  \state{Beijing}
  \country{China}}
\email{jicheng@act.buaa.edu.cn}


\author{Yiming Hei}
\affiliation{%
  \institution{School of Cyber Science and Technology, Beihang University}
  \city{Haidian}
  \state{Beijing}
  \country{China}}
\email{black@buaa.edu.cn}

\author{Jiawei Sheng}
\affiliation{%
  \institution{Institute of Information Engineering, Chinese Academy of Sciences}
  \state{Beijing}
  \country{China}}
\email{shengjiawei@iie.ac.cn}

\author{Qingyun Sun}
\affiliation{%
  \institution{School of Computer Science and Engineering, Beihang University}
  \city{Haidian}
  \state{Beijing}
  \country{China}}
\email{sunqy@act.buaa.edu.cn}

\author{Shan Xue}
\affiliation{%
  \institution{School of Computing, Macquarie University}
  \state{Sydney}
  \country{Australia}}
\email{emma.xue@data61.csiro.au}




\author{Pengtao	Xie}
\affiliation{%
  \institution{Department of Electrical and Computer Engineering, UC San Diego}
  \country{United States}}
\email{p1xie@eng.ucsd.edu}

\renewcommand{\shortauthors}{Qian Li, et al.}

\begin{abstract}

Event detection in power systems aims to identify triggers and event types, which helps relevant personnel respond to emergencies promptly and facilitates the optimization of power supply strategies.
However, the limited length of short electrical record texts causes severe information sparsity, and numerous domain-specific terminologies of power systems makes it difficult to transfer knowledge from language models pre-trained on general-domain texts. 
Traditional event detection approaches primarily focus on the general domain and ignore these two problems in the power system domain.
To address the above issues, we propose a Multi-Channel graph neural network utilizing Type information for Event Detection in power systems, named \textbf{MC-TED}, leveraging a semantic channel and a topological channel to enrich information interaction from short texts.
Concretely, the semantic channel refines textual representations with semantic similarity, building the semantic information interaction among potential event-related words.
The topological channel generates a relation-type-aware graph modeling word dependencies, and a word-type-aware graph integrating part-of-speech tags.
To further reduce errors worsened by professional terminologies in type analysis, a type learning mechanism is designed for updating the representations of both the word type and relation type in the topological channel. 
In this way, the information sparsity and professional term occurrence problems can be alleviated by enabling interaction between topological and semantic information. 
Furthermore, to address the lack of labeled data in power systems, we built a Chinese event detection dataset based on electrical Power Event texts, named \textbf{PoE}.
In experiments, our model achieves compelling results not only on the PoE dataset, but on general-domain event detection datasets including ACE 2005 and MAVEN.
\end{abstract}


\maketitle

\section{Introduction}

Electrical power systems \cite{el1995electrical, arrillage1983computer} provide the whole electricity supply across the country to ensure the basic needs of people's livelihood. The application of automated monitoring and analysis on the device-related event reports in the electrical power systems can improve management efficiency, save labor cost and maintain power supplement stability \cite{DBLP:conf/iciot3/HussainAIS20, DBLP:conf/isgt/RighettoMCHF21}.
One critical step for this is Event Detection (ED) \cite{DBLP:conf/acl/FengHTJQL16, DBLP:conf/acl/SunHLL18a, DBLP:journals/fcsc/LiuPLSL20} and specifically, the electrical event detection. The electrical event detection task refers to identifying triggers from power records and correctly classifying the device-related events. 
Through precisely detecting electrical events, we can swiftly obtain essential information, e.g., latent device risks, and help execute subsequent analysis. 
Thus, we focus on the event detection task in the electrical power systems.

\begin{figure}[t]
    \centering
    \includegraphics[width=0.95\linewidth]{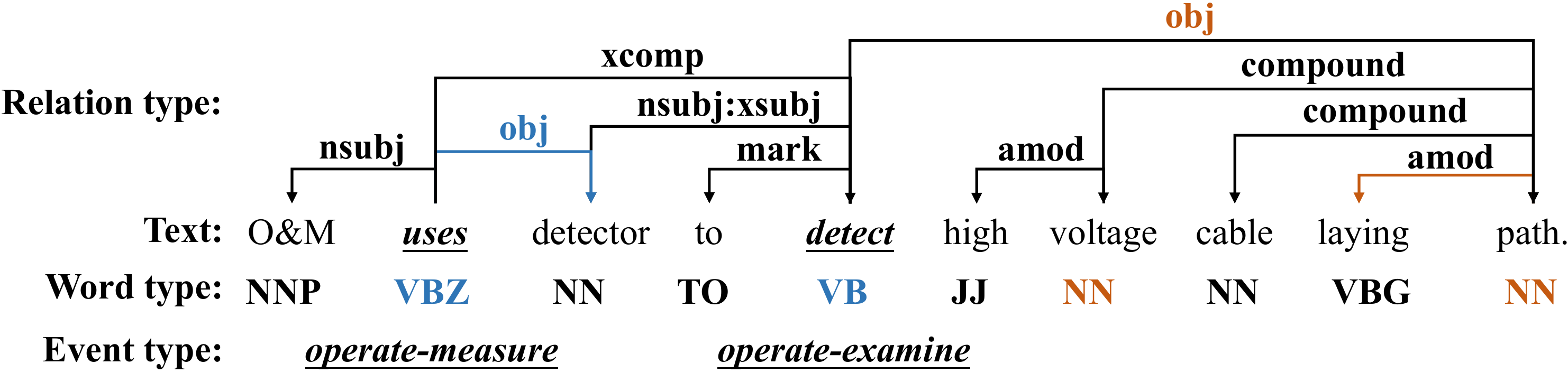}
    \caption{An example in our dataset. Word type means Part-of-Speech, and relation type means dependencies between words with relation. The triggers are ``uses'' and ``detect''.}
    \label{example}
\end{figure}

There are two main problems for electrical event detection: short texts and professional terminology.
Due to the advantage of short texts being easy to read and spread, device events are often described in short sentences for efficient communication in the electrical power systems.
However, the descriptions of short texts are inadequate, sparse, and irregular, which imposes higher requirements on feature learning.
Furthermore, electricity texts contain many terminologies, such as ``high voltage'' and ``laying path'' (shown in Fig. \ref{example}), which are rarely mentioned in general-domain texts. 


Event detection methods based on graph representation learning learn semantic structural information in text by composing graphs. Graph representation learning aims to map nodes, edges, and even the entire graph to a low-dimensional space. It maintains the structure and properties of the graph itself, which in turn serves the downstream graph data mining task. Homogeneous graphs can only characterize data with simple structure and semantic homogeneity, while textual composition graphs will contain multiple types of nodes and relations with rich semantic information. Heterogeneous graphs can fully portray such responsible nonlinear relationships with stronger representational and inference capabilities. Each type of node in a heterogeneous graph contains a variety of rich semantic information and complex interactions between nodes. Each type of semantic information also reflects only one aspect of the node's characteristics. In order to learn a more comprehensive node representation, the model needs to fuse multiple aspects of information to enhance the node representation.

Existing methods for power system event detection fail to address the above two problems. 
In recent years, researchers have proposed many methods for event detection and achieved promising results \cite{DBLP:conf/aaai/Liu00018, DBLP:conf/www/ZhengCCLC21, DBLP:conf/www/CaoPWDLY21}.
However, most of these methods focus on dealing with the long-tail distribution \cite{DBLP:conf/emnlp/WangWHJHLLLLZ20} of data and few-shot data \cite{DBLP:conf/www/ZhengCCLC21, DBLP:conf/acl/CongCYLWW21}, ignoring the above-mentioned problems in power systems. Existing methods can be categorized into two main paradigms: BERT-based methods \cite{DBLP:conf/naacl/WangHLSL19} and graph-based methods \cite{DBLP:conf/emnlp/YanJMGC19, DBLP:conf/emnlp/Cui0L0WS20, DBLP:conf/acl/LouLDZC20}. (1) BERT-based models perform well in general domain event detection due to the large amount of data available for pre-training. 
However, in the domain of power systems, some combinations of common words can have different meanings. For example, ``laying path'' means ``put something on a path'' in generation domain, but it means ``laying a route of cable'' in a power system. The distinction between general and electrical domains causes severe semantic confusion. This problem can be potentially resolved by pre-training models on labeled data in the power field, which takes time and experience.
(2) Graph-based models integrate various methods to design a graph with texts \cite{DBLP:conf/emnlp/Cui0L0WS20,DBLP:conf/acl/LouLDZC20, DBLP:journals/fgcs/MaoLPLHGHW21}. As shown in Fig. \ref{example}, we give an example of words, and relation is obtained by the Stanford CoreNLP tool \footnote{\url{http://corenlp.run/}}. It can bring a variety of interactive information among words.
Nevertheless, graphs constructed from short electricity texts contain few nodes and relations, which hinders the effectiveness of graph representation learning. 

To solve these problems, we propose a \underline{\textbf{M}}ulti-\underline{\textbf{C}}hannel GNN Utilizing \underline{\textbf{T}}ype Information for \underline{\textbf{E}}vent \underline{\textbf{D}}etection (\textbf{MC-TED}). 
It contains a semantic channel and a topological channel to fuse multiple sources of information to learn better node representations.
The semantic channel reflects the similarity of words through adding relation with lower cosine distance.
The topological channel constructs a relation-type-aware graph and a word-type-aware graph to learn more comprehensive node representations.
They consider the word type and relation type to portray nonlinear relationships with stronger representative and inference capabilities fully. 
Each type of word and relation contains a variety of rich semantic information and complex interaction between nodes. 
Furthermore, we use the Stanford CoreNLP tool to initialize the word and relation type, which is updated in training to mitigate the errors caused by Stanford CoreNLP. 
To address the lack of labeled data in the domain of the power systems, we build a \underline{\textbf{Po}}wer \underline{\textbf{E}}vent detection dataset (\textbf{PoE}).
Experimental results show that our approach can achieve excellent performance on both the PoE dataset and public event detection datasets, including the ACE 2005 \cite{DBLP:conf/lrec/DoddingtonMPRSW04} and MAVEN \cite{DBLP:conf/emnlp/WangWHJHLLLLZ20} datasets. We also demonstrate that our word type and relation type learning strategies are particularly beneficial for the word-type-aware graph and relation-type-aware graph.

This paper makes the following contributions. 
\begin{itemize}[leftmargin=*]
\item We design a novel multi-channel event detection model (MC-TED\footnote{The source code is available at \url{https://anonymous.4open.science/r/MC-TED-3C77}.}) utilizing word and relation type information for the power systems from semantic and topological channels with type utilized multi-channel GNNs to capture richer information. 
\item Topological channel designs a relation-type-aware graph to utilize the relation type and a node-type-aware graph considering the node type. The two types can be learnable to mitigate the errors caused by the CoreNLP syntactic analysis tools. 
\item We construct a Chinese event detection dataset -- PoE, in the domain of power systems, derived from power system records. Our model achieves state-of-the-art performance on the PoE as well as on the general-domain datasets including the ACE 2005 and MAVEN.
\end{itemize}

\section{Related Work}
The electrical power systems contain numerous event text, but engineers only focus on the core events regularly. It requires models to capture certain types of events, which is a schema-based event detection task \cite{DBLP:conf/emnlp/YanJMGC19, DBLP:conf/emnlp/Cui0L0WS20, DBLP:conf/acl/LouLDZC20, peng2022reinforced, 9927311}. Thus, event detection in power systems is to find triggers from the text and identify the event type corresponding to the triggers.

\subsection{Event Detection in Power Systems}

Large-scale new energy is integrated into the power grid system, which introduces strong uncertainty to the power system \cite{DBLP:journals/tsg/Jiang21}. It brings significant challenges to detect events in the modern power systems.
Traditional methods convert it to a constrained optimization problem \cite{smith2009event}, or non-convex optimization learning algorithms \cite{DBLP:journals/tsg/ZhouAS18}, etc.
To detect the abnormal events in the power systems using reduced phasor measurement units data, Liu et al. \cite{liu2019data} propose the unequal-interval reduction method with local outlier factor.
Zhou et al. \cite{DBLP:journals/tsg/ZhouAS18} propose a hidden structure semi-supervised machine for the event detection task. It incorporates information from labeled data and unlabeled data for the power distribution network.
$\text{HS}^{3} \text{M}$ \cite{DBLP:journals/tsg/ZhouAS18} combines unlabeled data and partly labeled data in a large margin learning objective, which is a classical method for event detection in the power systems. It incorporates partial knowledge for event detection. SS-LOF \cite{liu2019data} is an event detection algorithm in the power systems using reduced data and local outlier factor for detecting the events in power systems. It is employed to determine the region of the event source.
There are some graph-based methods for spatio-temporal solar irradiance forecasting \cite{khodayar2019convolutional}, and short-term wind speed prediction \cite{khodayar2018spatio}, etc.
To our best knowledge, there are few graph representation learning methods for the event detection task \cite{ozcanli2020deep, khodayar2020deep}.

\subsection{Graph-Based Event Detection}


In recent years, there are two main branches: BERT based models \cite{DBLP:conf/acl/DaganJVHCR18, DBLP:conf/naacl/WangHLSL19, DBLP:journals/taslp/LiPLWNWYW22} and graph based models \cite{DBLP:conf/emnlp/LaiNN20, DBLP:conf/sigir/LaiNND21, DBLP:journals/tkdd/PengLSYRYH21, DBLP:conf/ijcai/PengLGSNLY19, DBLP:journals/corr/abs-2205-12179}. 
The BERT-based methods learn a better context representation through self-attention mechanisms, and large-scale pre-train \cite{DBLP:conf/naacl/DevlinCLT19}. However, for the text with professional terms and short text, there are no apparent advantages.
The graph based methods commit to constructing syntactic dependency graph \cite{DBLP:conf/aaai/NguyenG18, DBLP:conf/emnlp/YanJMGC19, DBLP:conf/emnlp/Cui0L0WS20, li2022aiqoser}, and building event correlations \cite{DBLP:conf/acl/XieSZQD21, DBLP:journals/corr/abs-2104-15104}, etc.
There have been previous approaches to incorporate syntactic information into the Graph Convolutional Network (GCN) \cite{DBLP:conf/emnlp/YanJMGC19}, often ignoring dependency tag information to convey rich and valuable language knowledge to ED. 
Nguyen et al. \cite{DBLP:conf/aaai/NguyenG18} propose a GCN-based event detection model over syntactic dependency trees and entity mention-guided pooling. It operates a pooling over the graph-based convolution vectors of the current word and the entity mentioned in the sentences.
MOGANED \cite{DBLP:conf/emnlp/YanJMGC19} uses a dependency tree-based GCN with aggregative attention to explicitly model and aggregate multi-order syntactic representations in sentences. It models multi-order representations via graph attention network (GAT). It utilizes both first-order syntactic graphs and high-order syntactic graphs to explicitly model multi-order representations of candidate triggers.
These methods for ED fail to exploit the overall contextual importance of the words, which can be obtained via the dependency tree, to boost the performance. Lai et al. \cite{DBLP:journals/corr/abs-2010-14123} propose a novel gating mechanism to filter noisy information in the hidden vectors of the GCN models for ED based on the information from the candidate triggers. They also introduce a mechanism to achieve the contextual diversity for the gates and the importance of score consistency for the graphs and models in ED. They demonstrate how gating mechanisms, gate diversity, and graph structure can integrate syntactic information and improve the hidden vectors for ED models. GatedGCN \cite{DBLP:conf/emnlp/LaiNN20} is a gating mechanism to filter noisy information in the hidden vectors of the GCN models for ED. 
Lai et al. \cite{DBLP:conf/sigir/LaiNND21} present how transferring open-domain knowledge from word sense disambiguation and regulating representation based on pruned dependency graphs can improve few-shot learning on large-scale datasets. They propose a training signal derived from dependency graphs to regularize the representation learning. 

There have been previous approaches to incorporate syntactic information into the GCN, often ignoring dependency tag information to convey rich and valuable language knowledge to ED.
Cui et al. \cite{DBLP:conf/emnlp/Cui0L0WS20} propose an edge-enhanced graph convolutional network, which uses both syntactic structures and dependent label information to perform ED.
For the event correlations-based graph for event detection task, Xie et al. \cite{DBLP:conf/acl/XieSZQD21} formulate event detection task as a graph parsing problem, to overcome the inherent issues with existing trigger classification-based models. 
It can explicitly model the multiple event correlations and naturally utilize the rich information conveyed by event type and sub-type. 
Dutta et al. \cite{DBLP:journals/corr/abs-2104-15104} propose a framework for incorporating both dependencies and their labels using the graph transformer networks \cite{yun2019graph}. 
However, these methods cannot be completely applied to electrical power systems due to the characteristic of short text and professional terms.

\subsection{Topological Graph Learning}

Graph neural network~\cite{scarselli2008graph, zhou2020graph, DBLP:conf/www/CaoPWDLY21} can fully describe the complex nonlinear structure on the graph and has a more vigorous representation and reasoning ability \cite{zhou2020graph, DBLP:journals/tnn/WuPCLZY21}. 
Through topological graph learning and utilizing the heterogeneity, the high-level information can be mined to overcome the short text \cite{DBLP:journals/tkde/PengLWWGYLYH21, DBLP:conf/emnlp/HuYSJL19}. 
Graph representation learning and graph neural networks have gradually become the hot field of graph data mining. Traditional graph representation algorithms are mainly shallow models, such as Metapath2vec \cite{DBLP:conf/kdd/DongCS17} and HERec \cite{DBLP:journals/tkde/ShiHZY19} based on the random walk and Skip-Gram model. They cannot effectively capture complex nonlinear structures on the graph. Graph neural network aims to map nodes, edges, or even the whole graph to low-dimensional vector space and keep the structure and properties of the graph itself. It can fully describe the complex nonlinear structure on the graph and has a more vigorous representation and reasoning ability.
The spatial embedding fusion method based on graph convolution neural network mainly realizes the aggregation and updating of node information in a multi-layer network.
In recent years, multi-graph \cite{DBLP:conf/aaai/KhanB19, DBLP:conf/sdm/MaWAYT19, DBLP:conf/ijcai/HuangLYN20, DBLP:journals/bigdatama/LiaoZC21} and heterogeneous graph \cite{DBLP:conf/kdd/TangQM15,DBLP:journals/bigdatama/ZhangX21, DBLP:conf/emnlp/HuYSJL19, DBLP:conf/kdd/FanZHSHML19, DBLP:conf/www/WangJSWYCY19, DBLP:conf/www/HuDWS20} are effective method two mining high-order and hidden information.
Khan et al. \cite{DBLP:conf/aaai/KhanB19} propose a multi-graph convolutional network model of multi-view network to solve the embedding problem of the existing multivariate relational network model.
Ma et al. \cite{DBLP:conf/sdm/MaWAYT19} propose a graph convolution model of multi-dimensional networks to capture more abundant node-level information.
Considering that most of the existing methods are over-parameterized and limited to learning node representation, Vashishth et al. \cite{DBLP:conf/iclr/VashishthSNT20} propose a Composition-based multi-relational Graph Convolutional Networks (CompGCN) in which nodes and relations are embedded into the relational graph. 
Huang et al. \cite{DBLP:conf/ijcai/HuangLYN20} propose the framework of a multi-graph convolutional network by developing new convolution operators on multi-graph.
To solve the problem of short text classification, Hu et al. \cite{DBLP:conf/emnlp/HuYSJL19} convert text into a heterogeneous graph and design a dual-level heterogeneous graph attention network for learning text representation.

Heterogeneous graph neural networks can fully mine the complex structure and rich semantics to learn node representation and improve the performance of subsequent tasks.
PTE \cite{DBLP:conf/kdd/TangQM15} decomposes the text corpus into multiple heterogeneous text graphs and realizes the representation learning of text graphs through the joint decomposition of multiple graphs.
HAN \cite{DBLP:conf/www/WangJSWYCY19} is a heterogeneous graph neural network integrating hierarchical attention mechanism, which learns node representation by weighted fusion from node level and semantic level, respectively.
MEIRec \cite{DBLP:conf/kdd/FanZHSHML19} is a classical heterogeneous graph neural network, which can learn node representation by integrating rich semantic information provided by multi-element paths.
HGT \cite{DBLP:conf/www/HuDWS20} learns node representation based on heterogeneous graph neural network by aggregating heterogeneous relation triples.

\begin{figure*}[!htbp]
    \centering
    \includegraphics[width=\linewidth]{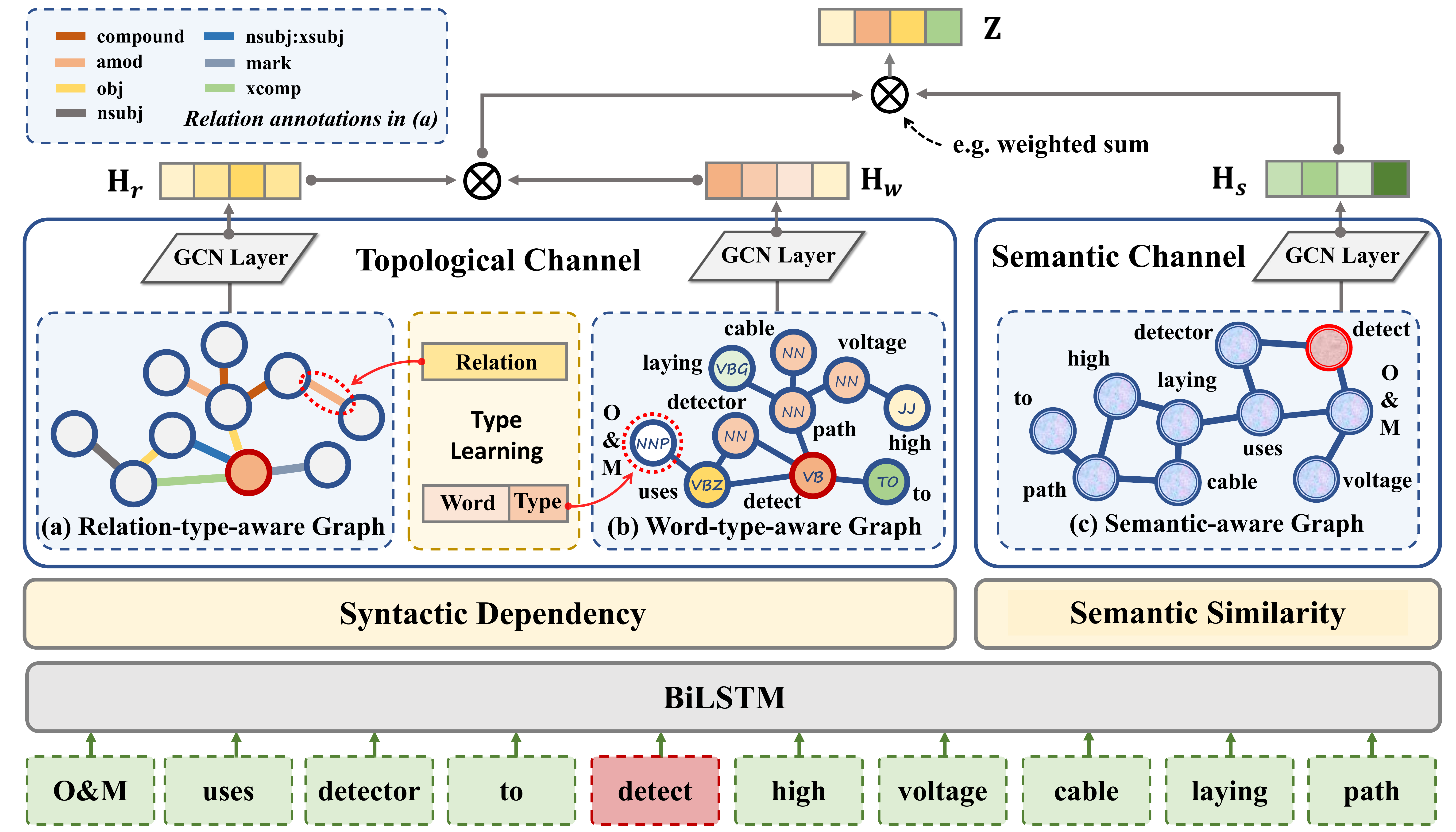}
    \caption{Our multi-channel event detection framework, MC-TED. It contains a topological channel and semantic channel. The topological channel includes two heterogeneous graphs utilizing word dependencies and part-of-speech tags. A relation-type-ware graph updates relation type, and a node-type-aware graph updates node type. The semantic channel refines the textual representations with semantic similarity. }
    \label{Framework}
\end{figure*}

\section{Framework}
We build a multi-channel event detection model for the power systems (MC-TED), as shown in Fig. \ref{Framework}. 
The inputs of our MC-TED model are the words obtained by the CoreNLP syntactic dependency analysis tool. 
All words are fed into a Bidirectional Long Short-Term Memory (BiLSTM) \cite{schuster1997bidirectional} to get their initial representations.
Then, the MC-TED model learns from the topological and semantic channels through the type utilized multi-channel GNN: 
\begin{itemize}[leftmargin=*]
    \item In the topological channel, we construct a syntactic graph based on the syntactic dependency analysis results generated by the Stanford CoreNLP tool. To consider the heterogeneity of relations and words, we further design a relation-type-aware graph and a word-type-aware graph, considering the relation types between words and the word types, respectively. In the two graphs, the representations of relation types and node types are both learnable. 
    \item In the semantic channel, we build a semantic-aware graph according to the semantic distance of the current word representations. The structure of the semantic-aware graph dynamically changes during the training process of the MC-TED. It reflects the similarity of words through adding relation with lower cosine distance. 
\end{itemize}

Through the topological and semantic channels, the MC-TED integrates multiple aspects of multiple sources of information (eg,. node and context) to learn better word representations through the type utilized multi-channel GNN. The type utilized multi-channel GNN considers the word type and relation type to portray nonlinear relationships with stronger representative and inference capabilities fully. 
At the final step of the MC-TED, the weighted average of the two channels is normalized to obtain the final representation of each word. It then detects triggers and event types in each sentence through a weighted mechanism to obtain the optimal weights.

\subsection{Word Feature Representation}

We use the CoreNLP syntactic analysis tool to divide a sentence $T$ into words ${[w_{1}, w_{2}, \dots, w_{N}]}$, where $N$ is the number of words in $T$. 
The word representation can be contained by the GloVe 100-dimensional word vectors \cite{DBLP:conf/emnlp/PenningtonSM14, 10.1145/3495162}. For our Chinese power event detection dataset PoE, a sentence needs to be firstly segmented through word segmentation tools.
The word with multiple characters is initialized randomly to keep dimensions consistent.
Then the word embeddings are fed into a BiLSTM to obtain the contextual representations $\mathbf{H}^0$.

\subsection{Type Utilized Multi-Channel GNNs}

Graph neural network \cite{DBLP:journals/tnn/WuPCLZY21} can mine the complex structure and rich semantics of the graph to learn node representation and improve the performance of subsequent tasks. 
In event detection, by modeling multiple syntactic dependency associations between words, GNN-based models can aggregate diverse interaction information to understand the multi-semantic expression of words.

To harness the great power of GNN, we design a multi-channel graph neural network for event detection through semantic and topological channels.
As illustrated in Fig.~\ref{Framework}, it contains: (1) \textbf{topological channel}, where a relation-type-aware graph (Fig. \ref{Framework} (a)) and a word-type-aware graph (Fig. \ref{Framework} (b)) are generated to achieve a more detailed and comprehensive description considering the semantics of word and relation types, and (2) \textbf{semantic channel}, where a semantic-aware graph is constructed based on the semantic similarity to refine the textual representations for the event detection.
With the two channels, MC-TED integrates multiple aspects of information to learn a more comprehensive word representation.

\subsection{Topological channel}
The text contains rich syntactic (topological) information, and the representations of words change in different contexts. 
To improve comprehensively word representation, we use the Stanford CoreNLP syntactic analysis tool to construct a syntactic graph where the nodes denote the words in a sentence and edges denote the dependencies between words, with the part-of-speech tags (e.g., noun and verb) as node types and the dependency classes (e.g., nominal subject and direct object) as relation types.
Formally, the syntactic graph is a form of heterogeneous graph $\mathcal{G}=(\mathcal{V}, \mathcal{E})$ with a word type mapping function $\psi_w : \mathcal{V} \to \mathcal{A}$ and a relation type mapping function $\psi_r : \mathcal{E} \to \mathcal{R}$, where $\mathcal{V}$ is the set of words and $\mathcal{E}$ is the set of relations, $\mathcal{A}$ is the set of word types and $\mathcal{R}$ is the set of relation types, and $|\mathcal{A}|+|\mathcal{R}|>2$.

\begin{figure}[t]
    \centering
    \subfigure[Different relation types between the same node type pair.]{
 \includegraphics[width=0.35\linewidth]{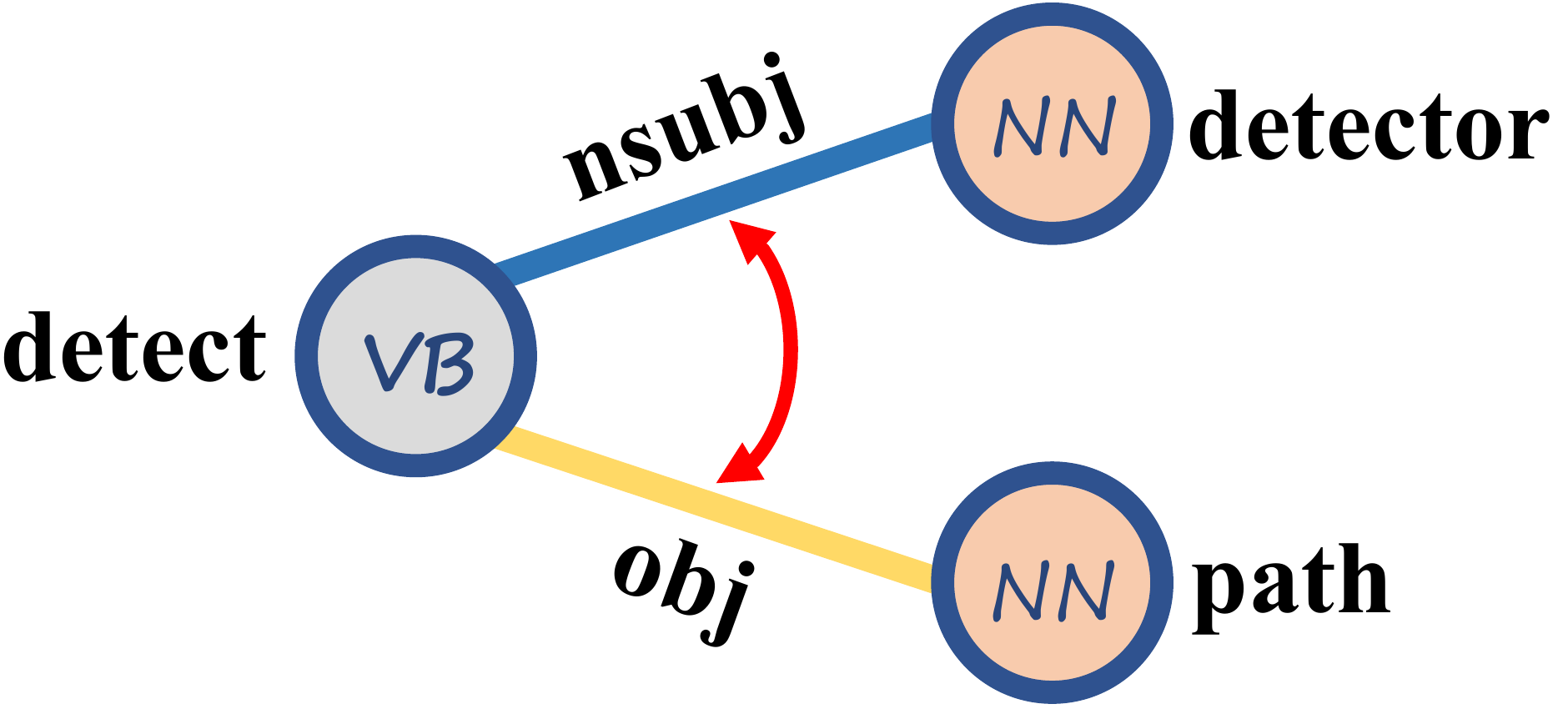}}
 \hspace{0.4in}
  \centering
 \subfigure[Different word types connected by the same relation type.]{
 \includegraphics[width=0.35\linewidth]{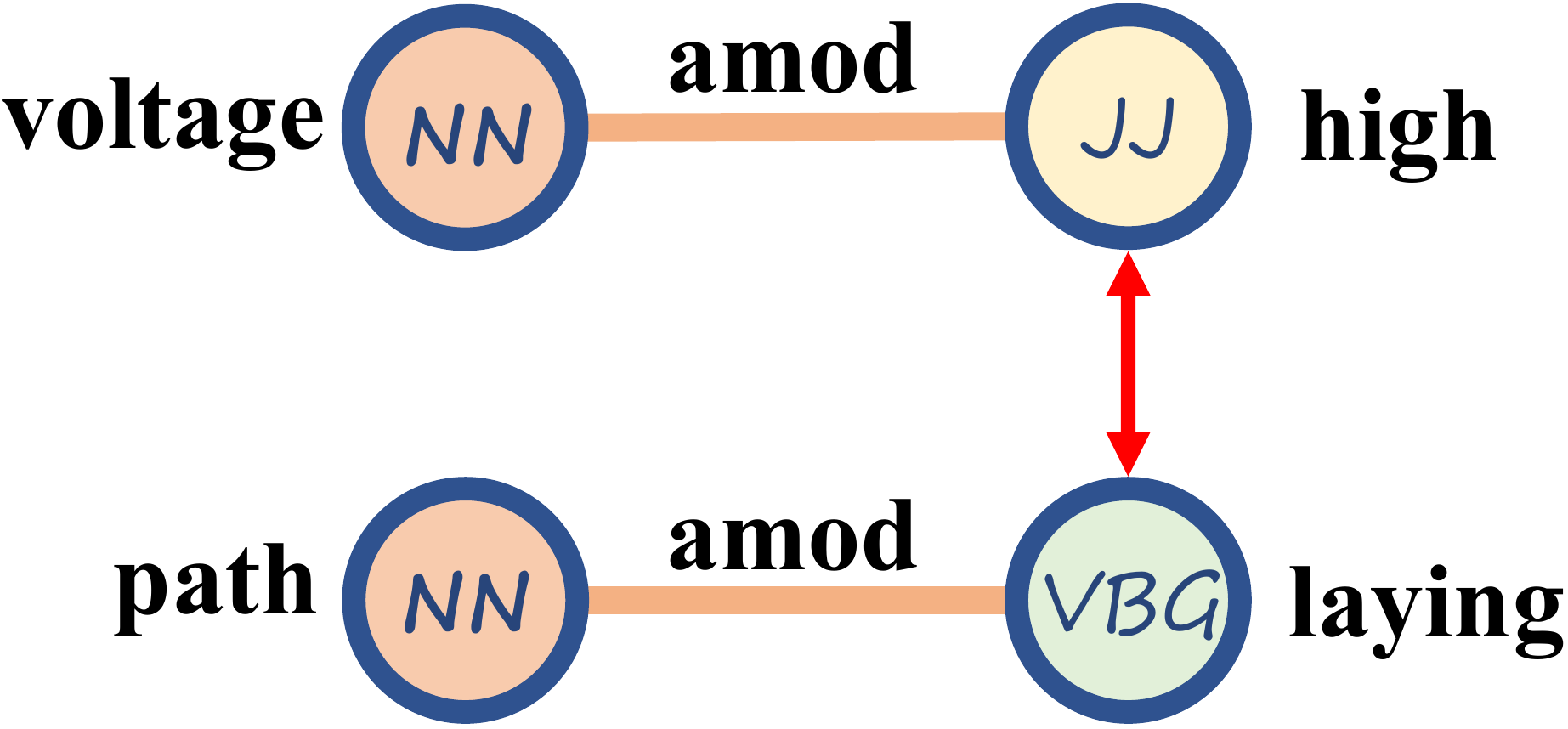}}
    \caption{The heterogeneity of relations and words. For one sentence, there are both multi-relationship types for the same node type as the left, and multi-node types for the same relation type as the right.}
    \label{graph}
\end{figure}

\textbf{Insights on different type information.}
Different from the common heterogeneous graphs, the syntactic graph contains two kinds of type information that cannot be substituted or deduced from each other, the word type and the relation type. It is noticed that there are two phenomenons in the syntactic graph: (1) some relations between two words with the same type belong to different types (i.e., ``$A_1\stackrel{R_1}{\longrightarrow}A_2$ and $A_1\stackrel{R_2}{\longrightarrow}A_2$''), and (2) relations with the same type may connect two words whose types are different (i.e., ``$A_1\stackrel{R}{\longrightarrow}A_2$ and $A_1\stackrel{R}{\longrightarrow}A_3$''). Taking the text in Fig.~\ref{example} as an example, as show in Fig.~\ref{graph}, the relation types between the words with type ``VB'' and type ``NN'' are from \{``nsubj'', ``obj''\}, and the relation with type ``amod'' connects two words with types \{``NN'', ``JJ''\} and \{``NN'', ``VBG''\}. 
Therefore, only obtaining relation type information or word type information is inadequate, as the two types cannot be substituted or deduced.

To address the above issue, we divide the syntactic graph into a relation-type-aware graph and a word-type-aware graph to utilize different type information for the type utilized multi-channel GNN.
Specifically, (1) the \textbf{relation-type-aware graph} has only one word type (i.e., $|\mathcal{A}|=1$) and multiple relation types (i.e., $|\mathcal{R}|>1$) to learn the topological information under the relation type constraint, 
and (2) the \textbf{word-type-aware graph} has multiple word types (i.e., $|\mathcal{A}|>1$) and only one relation type (i.e., $|\mathcal{R}|=1$) for word type learning.

Furthermore, the type information in syntactic is obtained by the CoreNLP syntactic analysis tool, which leads to type fault in some cases, especially for the electrical power system with numerous professional terminology (see an example in Section~\ref{case_study}).
We design a type learning mechanism in both the relation-type-aware graph and word-type-aware graph for the type utilized multi-channel GNNs to lessen this error caused by short text and professional terms.

\textbf{Relation-type-aware graph.}
To consider the heterogeneity of relations, we design a relation-type-aware graph $\mathcal{G}_{r}=(\mathcal{V}_{r}, \mathcal{E}_{r})$ with the mapping functions $\psi_{r,v}: \mathcal{V}_{r} \rightarrow \mathcal{A}_{r}$ and $\psi_{r,r}: \mathcal{E}_{r} \rightarrow \mathcal{R}_{r}$, where $|\mathcal{A}_{r}|=1$ and $|\mathcal{R}_r|>1$. 
The relation-type-aware graph does not introduce node type information that means all nodes are the same type. However, the relation types of this graph are different, which are obtained by the CoreNLP syntactic analysis tool. In this way, the relation-type-aware graph only focus on the relation types.
In order to utilize the information of relation types, we propose to learn the corresponding relation type vectors as the edge representations:
\begin{equation}
    \mathbf{E}_{ij} \gets \mathbf{R}_{\psi_{r,r}(e_{ij})}, i,j \in [1,N],
\end{equation}
where $\mathbf{E} \in \mathbb{R}^{N \times N \times d_r}$ is the representation of the edge $e$, and $\mathbf{R} \in \mathbb{R}^{|\mathcal{R}_r| \times d_r}$ is the representation of the edge type which is a learnable tensor in the type learning mechanism ($d_r$ is the dimension of the relation type representation between two nodes). 

In the relation-type-aware graph, the representations of edges and nodes in the $l$-th layer are
 \begin{equation}
\mathbf{E}^{l}, \mathbf{H}_r^{l}=\mathrm{GNN}(\mathbf{E}^{l-1}, \mathbf{H}_r^{l-1}),
\end{equation}
where $\mathbf{H}_r^{l-1}$ is the word (node) representations in layer $l-1$ and $\mathbf{H}_r^0$ is the output of BiLSTM in word feature representation. In this paper, we adapt a two-layer GCN and the edge representation is used as the weight of the edge during the message-passing process.

In order to make the model better learn the representation of nodes under different relationships, our relation-type-aware graph updates the edge representation in each layer, making the model focus on learning edge information. The representation of edge in the layer $l$ is updated as
 \begin{equation}
\mathbf{E}_{i,j}^l =\mathbf{W}_{r}\left[\mathbf{E}_{i,j}^{l-1} || \mathbf{h}_i^{l} || \mathbf{h}_j^{l}\right], i, j \in[1, N],
\end{equation}
where $||$ means the concatenation operator, $\mathbf{h}_i^{l}$ and $\mathbf{h}_j^{l}$ denote the representations of the two nodes $i$ and node $j$ connected by the edge $e_{i,j}$ in the $l_{th}$ layer, respectively, $\mathbf{W}_{r} \in \mathbb{R}^{(2 \times d+d_r) \times d_r}$ is a learnable transformation matrix ($d$ is the dimension of word representation).

\textbf{Word-type-aware graph.}
We next design a word-type-aware graph $\mathcal{G}_{w}=(\mathcal{V}_{w}, \mathcal{E}_{w})$ with the mapping functions $\psi_{w,v}: \mathcal{V}_{w} \rightarrow \mathcal{A}_{w}$ and $\psi_{w,r}: \mathcal{E}_{w} \rightarrow \mathcal{R}_{w}$, where $|\mathcal{A}_{w}|>1$ and $|\mathcal{R}_w|=1$.
The word-type-aware graph does not introduce the relation type and considers the word types obtained by the CoreNLP syntactic tool. The node embedding of this graph is the concatenation of word embedding and word type embedding. 
Thus, for word type learning, we propose to learn the corresponding word type vectors with the word representation:
\begin{equation}
    \mathbf{H}_w^l, \mathbf{A}^l = \mathrm{GNN}([ \mathbf{H}_w^{l-1} || \mathbf{A}^{l-1} ]),
\end{equation}
where $\mathbf{A} \in \mathbb{R}^{N \times d_w}$ is a learnable vector for the word type representation in the type learning mechanism ($d_w$ is the dimension of the word type representation).

The nodes and edges in the node-type-aware graph are the same as the relation-type-aware graph, but the difference is that the types of nodes in this graph are considered instead of the types of relations between nodes.
Our node-type-aware graph updates the representation for each node by aggregating the information from its neighbors through the adjacency matrix. It helps for learn more context information for each node.
In order to make the model better learn the representation of nodes under different node types, our node-type-aware graph channel only updates the node type representation, making the model focus on learning node information. 

\subsection{Semantic channel}
Topological channel can learn word representations based on syntactic dependencies from two dimensions: node type and relation type. In addition, we design a semantic channel to learn connections based on the semantic distance between words. It metrics the similarity of semantics among words.
The semantic channel can establish relationships between words that are far apart and grammatically independent but have similar meanings.

\textbf{Semantic-aware graph.}
Different from the topological channel, we establish links between the words based on the semantic similarity. We then generate edge weights based on word semantic embeddings over the corpus. The similarity score can be calculated through cosine distance.
If the similarity score exceeds a predefined threshold $\rho_{\text {sem }}$, it means that the two words have a semantic relationship in the current sentence. The edge weight of each pair of words can be obtained by
\begin{equation}
\alpha_{i,j}=\text{sim} (\mathbf{h}_{i}, \mathbf{h}_{j})=\begin{cases}\frac{\mathbf{h}_{i} \cdot \mathbf{h}_{j}}{||\mathbf{h}_{i}||\cdot|| \mathbf{h}_{j}||}, &\cos (\theta) \geq \rho_{\text {sem }}\\ 0, & \cos (\theta) < \rho_{\text {sem }}\end{cases},
\end{equation}
where $\alpha_{i,j}$ denotes the edge weight between words $i$ and $j$.

Note that the structure of the semantic-aware graph is dynamically changing in training process, because the edge weight depends on the current representations of words in each step of the training process.
Finally, given the current structure of the semantic-aware graph, a GNN is used for learning the representations of words $\mathbf{H}_s$.

\subsection{Event Detection}
Event detection is a fundamental task in the field of natural language processing, aiming at identifying trigger words from the text and classifying the text into corresponding event types. Although current methods and models for event detection have achieved increasingly good results, they are mostly trained based on generic domain datasets and often encounter the problem of insufficient annotated data when faced with event detection tasks in a specific domain.
In electrical power systems, device events are often described in short sentences for efficient communication and convenience for reading and spreading.
Thus, it requires the model to impose higher requirements on feature learning and fuse multiple aspects of information to enhance the node representation.

\begin{table}[t]
\caption{Statistics of datasets used in our work. Different from other datasets, our dataset is a Chinese dataset in electrical power systems.}
\centering
\renewcommand\arraystretch{1.2}
\resizebox{\linewidth}{!}{
\begin{tabular}{l|cccccccc}
\toprule
\textbf{Dataset} & \textbf{Field} &\textbf{Language} & \textbf{\#Events} & \textbf{\#Event Types} & \textbf{\#Sentences} & \textbf{\#Tokens} &\textbf{\#Average Sentence Length}   \\     
\midrule
MAVEN~\cite{DBLP:conf/emnlp/WangWHJHLLLLZ20} & General& English & 111,611 & 168 & 49,873 & 1,276,000 &25.5\\ 
ACE 2005~\cite{DBLP:conf/lrec/DoddingtonMPRSW04} & General& English  & 4,090 & 33  &15,789 & 303,000  &19.1\\ 
\textbf{PoE (Ours)} & \textbf{Power} & \textbf{Chinese}  & \textbf{5,124}  & \textbf{26} & \textbf{4,767} & \textbf{53,071}&\textbf{11.1} \\ 
\bottomrule
\end{tabular}
}
\label{datasetall}
\end{table}

\begin{table}[t]
\small
\caption{Event type and sub-type on PoE. It contains six event types and twenty-six sub-types. The defect event type consists of eleven sub-types and the measurement event type only has one sub-type.}
\centering
\renewcommand\arraystretch{1.2}
\resizebox{\linewidth}{!}{
\begin{tabular}{c|cc}
\toprule
\textbf{Number} & \textbf{Event Type}  & \textbf{Event Sub-type}       \\\midrule
1  & measurement & measurement      \\ 
2  & statistics  & total, ratio     \\
3  & requirement & stipulate, expect        \\
4  & operate     &  connect, disconnect, modify, check  \\
5  & happen      &  start, complete, cancel, analyze, implement, suffer    \\
\multirow{2}{*}{6}  & \multirow{2}{*}{defect}      & damage, carve, break, burn, placement, dislocation,  deformation, corrosion \\
  &  &  mismatch, shed, variant\\
   \bottomrule                                 
\end{tabular}
}

\label{datasetours}
\end{table}

We pass the representation of the topological channel and the semantic channel through a weighted mechanism to obtain the optimal weights of the three different graphs. The word representations in text are formulated as follows:
\begin{equation}
\mathbf{Z}=\lambda_{1} \mathbf{H}_{r} + \lambda_{2} \mathbf{H}_{w}  + \lambda_{3} \mathbf{H}_{s},
\end{equation}
where $\lambda_{1}, \lambda_{2}, \lambda_{3}$ are the parameters. We feed the representation of each word into a fully-connected network. It is followed by a softmax function to compute distribution $p(t \mid \mathbf{Z})$ for event type $t$:
\begin{equation}
p(t \mid \mathbf{Z})=\operatorname{softmax}(\mathbf{W}_{t} \mathbf{Z}+\mathbf{b}_{t}),
\end{equation}
where $\mathbf{W}_{t}$ maps the word representation $\mathbf{Z}$ to the feature score for each event type and $\mathbf{b}_{t}$ is a bias term. After softmax, the event label with the largest probability is chosen as the event classification result.
The loss function is formulated as follows:
\begin{equation}
\begin{aligned}
J(\theta)=-\sum_{i=1}^{N_{T}} \sum_{j=1}^{N_i} \log p(P_{j}^{t} \mid T_{i}, \theta),
\end{aligned}
\end{equation}
where $N_{T}$ is the number of sentences, $N_i$ is the number of words in the $i$-th sentence, $P_{j}^{t}$ is the event type $t$ in the $j\!-\!th$ word, and $T_{i}$ is the $i$-th sentence.

\section{Datasets and Settings}
\subsection{Datasets} \label{datasets}

We construct a Chinese event detection dataset in electrical power systems, called PoE. 
The text is device-related event reports in electrical power systems, written by professional.   Due to the advantage of short texts being easy to read and spread, device events are often described in short sentences for efficient communication in electrical power systems.  Furthermore, the event release in power system is usually short and contains numerous professional terms, which is convenient for relevant personnel to respond quickly.  Thus, the raw data are the short sentence-level data.
It describes daily operation of electrical equipment containing 763 event reports.
There are 4,767 sentences, 5,124 events, and 53,071 tokens, containing 26 sub-categories of events, as shown in Table \ref{datasetall}. To verify the effect of MC-TED on open datasets, we also conduct experiments on ACE 2005 and MAVEN datasets. The average sentence length are 11.1 tokens in PoE, which is 14.4 and 8.0 less than MAVEN and ACE 2005. 
Furthermore, compared with ACE 2005 and MAVEN datasets, PoE has a few None event type sentences, which is the most event type in other datasets.

\begin{table}[t]
\small
\caption{Statistical information on PoE. It is divided into three parts: train, valid and test sets. The partition rule is to make the ratio of sentence number, event sentence number, and candidate trigger close to 8:1:1.}
\centering
\renewcommand\arraystretch{1.2}
\resizebox{\linewidth}{!}{
\begin{tabular}{c|ccccc}
\toprule
\textbf{Dataset} & \textbf{\#Sentence Number} &\textbf{\#Event Sentence Number} & \textbf{\#Event Trigger}  & \textbf{\#No Event Trigger} &\textbf{\#Candidate Trigger}    \\\midrule
Train   & 3,813    & 3,521  & 4,095  & 1,728  & 5,823  \\
Valid   & 477   & 452  & 515   & 217   & 732  \\
Test    & 477  & 451  & 514   & 215   & 729 \\ \bottomrule        
\end{tabular}
}
\label{trainValid}
\end{table}

\begin{table}[t]
\small
\caption{Statistical information of co-occurrence events on PoE training data. \#One Type Number means that the number of sentences having only one event type or one sub-type. \#Co-occurrence Proportion means that the proportion of co-occurrence sentences.}
\centering
\renewcommand\arraystretch{1.2}
\resizebox{\linewidth}{!}{
\begin{tabular}{c|ccccc}
\toprule
\textbf{Dataset} & \textbf{\#One Type Number} &\textbf{\#Two Type Number} & \textbf{\#Three Type Number}  & \textbf{\#Four Type Number} &\textbf{\#Co-occurrence Proportion}    \\\midrule
Event Type   & 2675    & 423  & 282   & 141  & 0.24 \\
Event Sub-type   & 2289   & 634  & 352   & 246   & 0.35  \\ \bottomrule        
\end{tabular}
}
\label{co-occurrence}
\end{table}

Table \ref{co-occurrence} shows co-occurrence events information of the event type and sub-type detail of PoE on training data. The co-occurrence sentences means that a sentence belong to different event types or event sub-types. Our data set contains up to four co-occurrence event types. For the event type and event sub-type, most sentences on the PoE contain only one event type and sub-type, which is the same as the existing event detection data set. However, the PoE contains more multi-event sentences. It has 12\% sentences containing two different event types and 18\% sentences containing two different event sub-types. Thus, our data set is more suitable for studying the event type detection problem caused by event co-occurrence.


\begin{figure*}[t]
    \centering
        \subfigure[Sentence length distribution.]{
     \includegraphics[width=5.2cm]{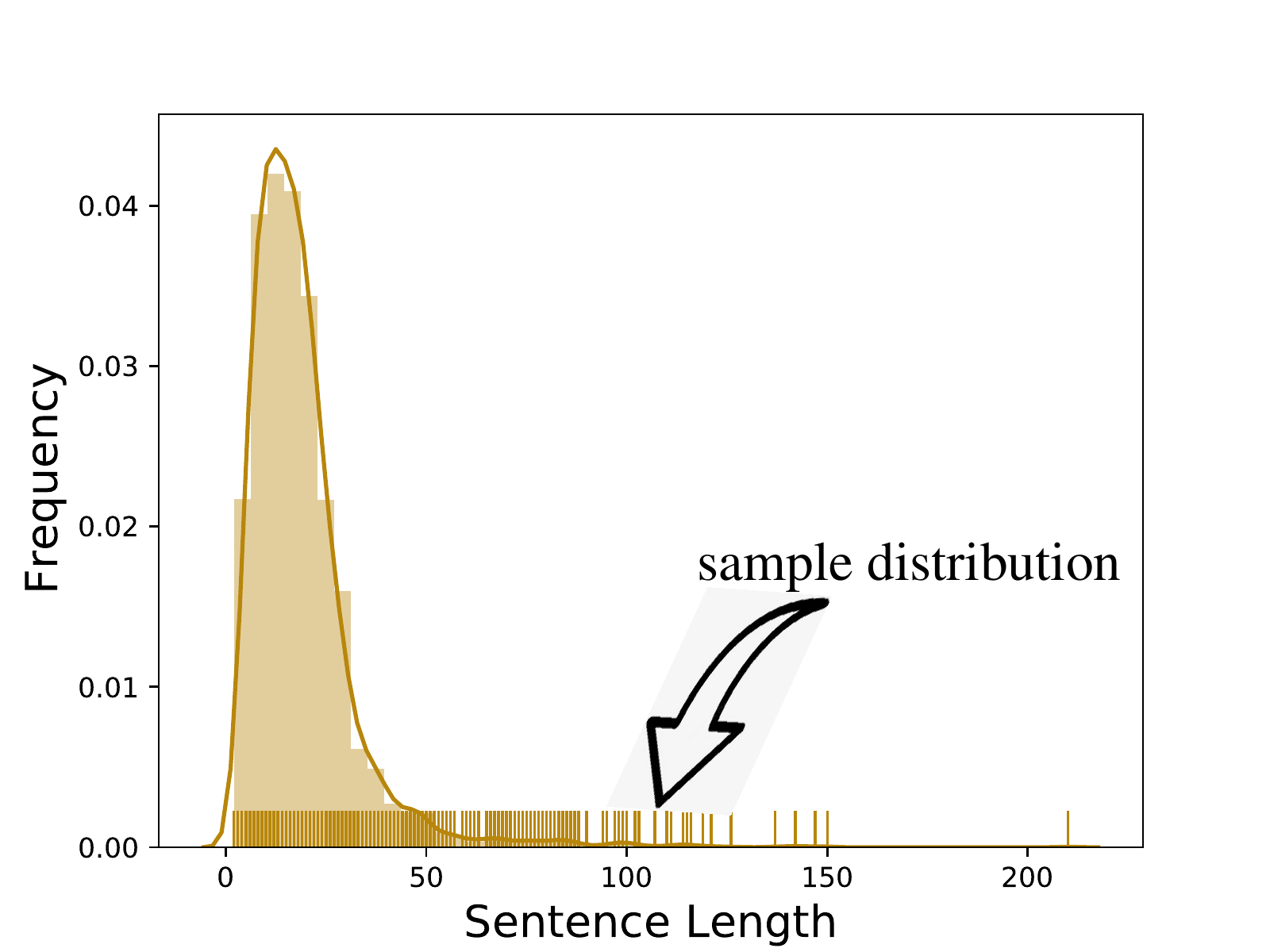}}
      \hspace{6mm}
    \centering
        \subfigure[Event type.]{
     \includegraphics[width=6.7cm]{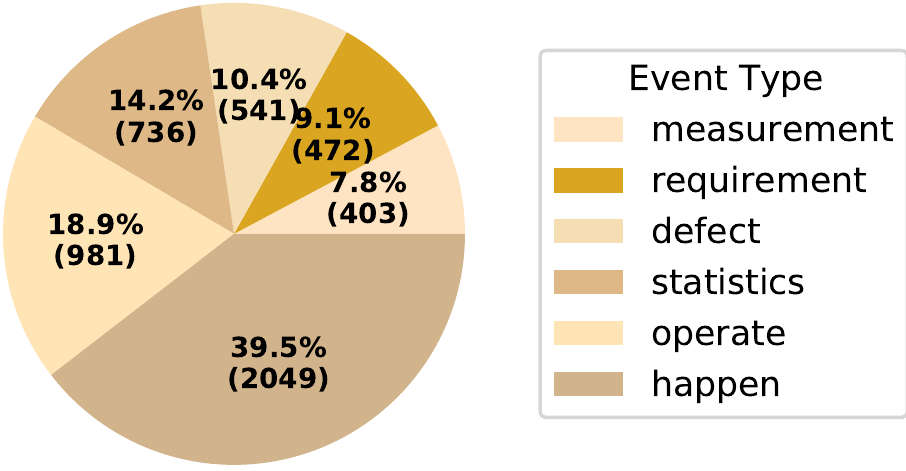}}
    \caption{Frequency and number of event type. (a) The samples are concentrated between 0 and 50. When the sentence length exceeds 100, the samples become very sparse. (b) The sentence distribution of event type. The largest number of event types is the happen and smallest is the measurement event type.}
    \label{Frequency}
\end{figure*}


\begin{figure*}[t]
    \centering
     \includegraphics[width=14cm]{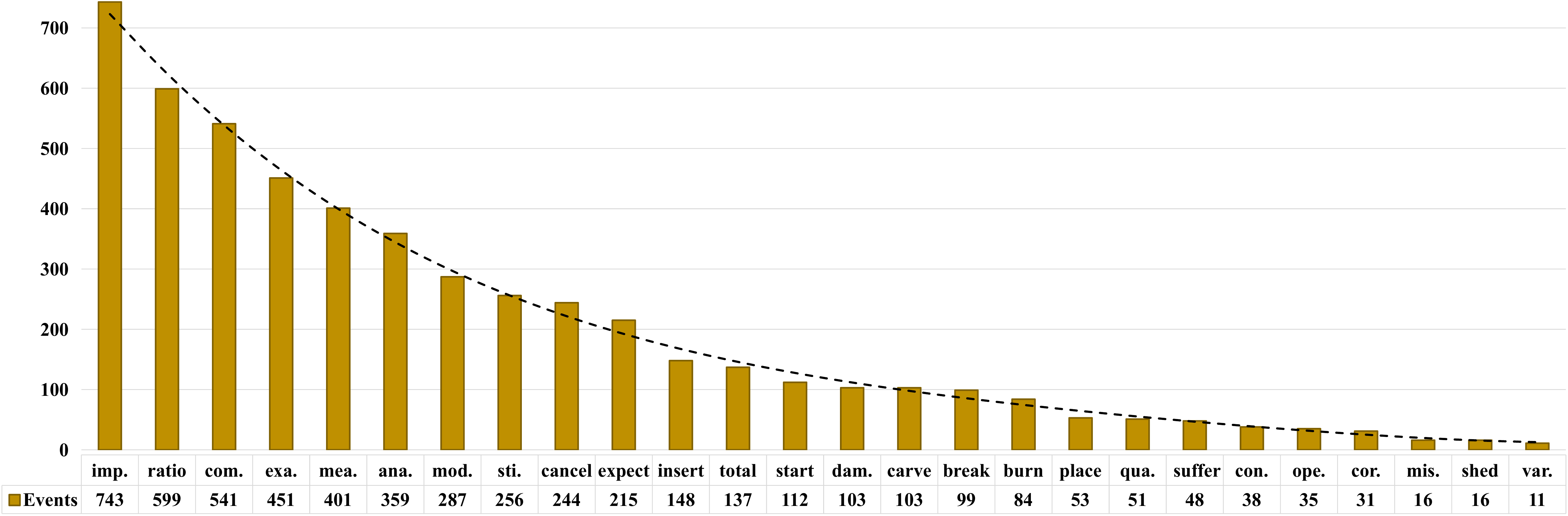}
    \caption{Sub-type event numbers. The distribution of sub-type events is uneven. The implement sub-type that belongs to the happen event type contains more than 700 sentences and the variant sub-type coming from the defect event type has less than 10 sentences.}
    \label{numbers}
\end{figure*}

Table \ref{datasetours} shows the event type and sub-type detail of PoE. It contains 6 categories of event type and 26 categories of event sub-type.
The PoE mainly describes the equipment statistics, defects, requirements and maintenance events in the electric power systems. The data set focuses on nine major equipment defects, including damage, burn, corrosion, long-term placement, dislocation, deformation, fall off and open circuit.
The datasets are divided into training, validation, and test sets at an 8:1:1 ratio according to the sentence number, as shown in Table \ref{trainValid}.
The proportion of event number, event trigger, and no event trigger in the three datasets are basically the same. The training set contains all event types to ensure that there are no invisible event types in the test set.

Furthermore, we analysis the sentence length and event type distributions in Fig. \ref{Frequency} and \ref{numbers}. In Fig. \ref{Frequency} (a), the maximum and shortest sentence length are 224 and 6. The most sentences are in [10,25], accounting for 72\% of the total data set. The average sentence length is 18.88. It can be seen that the text length is short in PoE, increasing the difficulty of event detection. 
In Fig. \ref{Frequency} (b), the event ``operate'' is the largest number of events in the six categories , with a total of 2,049 sentences, accounting for 39.98\%. The events with the least number are ``measure'' events (403 events in total, accounting for 7.86\%). 
In Fig. \ref{numbers}, the type with the largest number of events is ``implementation'' in 26 sub-categories, with a total of 743 events. The least number of events is the ``deformation'' event, 11 events in total. The difference between the type with the largest number of events and the type with the least number of events is very large, indicating that the long tail phenomenon is obvious in the sub-categories, the same as ACE 2005 \cite{DBLP:conf/lrec/DoddingtonMPRSW04} and MAVEN \cite{DBLP:conf/emnlp/WangWHJHLLLLZ20}.

Compared with the common ED datasets, PoE contains the following four differences:
\begin{itemize}[leftmargin=*]
    \item \textbf{Power Field.} Different from ACE 2005 and MAVEN, PoE is an event detection dataset from the power systems instead of the general field. The events in PoE, including electrical equipment maintenance measures and recommendations, are closely related to the business in the electric power field.
    \item \textbf{Short Text.} The texts in PoE are collected from the electrical records, which are usually the short sentences with non-compliant grammar. The little information brings difficulty to semantic understanding. (In PoE, the average text length is $18.8$.)
    \item \textbf{Professional Terms.} PoE is a Chinese sentence-level event detection dataset, containing numerous professional terms in the power systems, such as ``laying path'', ``direct buried'', and ``penetration seal''. It brings limitations to the existing methods in the general field, since the semantics of some terms changes when they are used in the power systems.
    \item \textbf{Colloquial expressions.} The existing data sets are crawling from news, whose expression is more formal, basically in accordance with the grammar rules. The PoE comes from equipment maintenance records of electrical worker. Therefore, it contains numerous colloquial expressions, and missing sentence elements.
\end{itemize}

\begin{table}[bp]\normalsize
\centering
\renewcommand\arraystretch{1.2}
\caption{Hyper-parameter settings in MC-TED. Due to the differences among the three datasets, the optimal hyper-parameters are different.}
\resizebox{0.65\linewidth}{!}{
\begin{tabular}{l|ccc}
\toprule 
\textbf{Hyper-parameters}  & \textbf{ACE 2005}  & \textbf{MAVEN} & \textbf{PoE} \\
\midrule
GCN initial learning rate   &  0.15 & 0.15   & 0.15 \\
Train epoch   &100  &100   &100 \\
Model dimensions &150  &150  &150 \\
Batch size    &10  &32  &10 \\
Model layer &2   &2  &2\\
Dropout rate   &0.55  &0.50  &0.60   \\
Threshold $\rho_{\text {sem }}$ &0.10  &0.12  &0.15\\
$\lambda_{1}$ & 2  &2  &2\\
$\lambda_{2}$ & 1  &1  &1\\
$\lambda_{3}$  & 2  &2  &2\\
\bottomrule
\end{tabular}}
\label{parameters}
\end{table}

\subsection{Implementation Details} \label{details}

All models are implemented based on PyTorch, an open-source deep learning framework. 
The model parameters are initialized as Gaussian distribution and updated and optimized by Adam algorithm~\cite{kingma2015adam}. 
For re-implementation, Table~\ref{parameters} reports our hyper-parameter settings in MC-TED on ACE 2005, MAVEN and PoE. 
The common parameters are set as follows: the learning rate is 0.15, the regularization coefficient is 0.001, the model layer is 2, the word representation dimension is 100, the word type representation dimension is 25, and the relation type representation dimension is 25. The batch size is 10 on ACE 2005 and PoE, and 32 on MAVEN. For our MC-TED, we adopt a two-layer GCN \cite{DBLP:conf/www/WangJSWYCY19} as the graph neural network in our framework. The random dropout ratio is set to 0.55, 0.50, and 0.60 on ACE 2005, MAVEN and PoE. The similarity score thresholds $\rho_{\text {sem }}$ are 0.10, 0.12 and 0.15 on ACE 2005, MAVEN and PoE. The early stop strategy is selected to prevent the model from overfitting. Specifically, if the model does not decrease the loss function of the validation set for 100 consecutive turns, the operation is stopped.

To ensure the fairness, all baseline models use the same dataset partitioning as stated in Section~\ref{datasets}. We use the same value for the common hyper-parameters, including the optimizer, learning rate, batch size and epoch. All graph-based, LSTM-based, and CNN-based models use the same node representation dimension, which is set to 150 dimensions to keep dimensions consistent with the GNN-based models. For CNN-based methods, the filter kernel size is set to $3 \times 3$ with two layers. For LSTM-based methods, the model layer is two. For BERT-based approaches, we use BERT \cite{DBLP:conf/naacl/DevlinCLT19} as sequence encoder, which has 12 layers, 768-dimensional hidden embeddings, and 12 attention heads. All the experiments are conducted on one NVIDIA V100 32GB GPU. Note that all the hyper-parameter settings are tuned on the validation set by the grid search with 5 trials.

\subsection{Baselines}
Many event detection models have been designed for these years, which can be divided into CNN-based, LSTM-based, GNN-based, BERT-based and power systems baselines according to model structure.
We compare our method with the following four types of ED baselines:

\paragraph{\textbf{CNN-based baselines.}} 
(1) CNN is a two-layer CNN model. It uses sequence tagging to implement event detection.
(2) DMCNN \cite{DBLP:conf/acl/ChenXLZ015} leverages a CNN to automatically learn sequence representations with a dynamic multi-pooling mechanism.

\paragraph{\textbf{LSTM-based baselines.}} 
(3) BiLSTM (2-layer) is a two layer BiLSTM.
(4) MLBiNet (2-layer) \cite{DBLP:conf/acl/LouLDZC20} uses a multi-layer bidirectional network to model document-level dependencies. We use CoreNLP to obtain the entity for embedding. 

\paragraph{\textbf{GNN-based baselines.}} 
(5) GCN (2-layer) \cite{DBLP:conf/iclr/KipfW17} is a semi-supervised graph convolutional network without considering node type and relation type. 
(6) GAT (2-layer) \cite{DBLP:conf/iclr/VelickovicCCRLB18} uses attention mechanism for learning neighbor weights. 
(7) MOGANED \cite{DBLP:conf/emnlp/YanJMGC19} uses a multi-order graph attention network to effectively model the multi-order syntactic relations in dependency trees. 
(8) GatedGCN \cite{DBLP:conf/emnlp/LaiNN20} designs a graph based on dependency tree and distance to the candidate trigger of each word. It considers the important scores to build the graph.
(9) EE-GCN \cite{DBLP:conf/emnlp/Cui0L0WS20} is an edge-enhanced graph convolutional network using the syntactic structure, which uses the syntactic structure and the typed dependent label information.

\paragraph{\textbf{BERT-based baselines.}} 
(10) BERT+CRF is a vanilla BERT model. It uses a multi-classification approach to implement event detection.
(12) BERT+BiLSTM uses a BERT based model and a two layer BiLSTM after the BERT.
(13) BERT+CNN adapts a BERT model and a two layer CNN model after the BERT.
(14) DMBERT \cite{DBLP:conf/naacl/WangHLSL19} takes advantage of BERT and adopts the dynamic multi-pooling mechanism to aggregate features.

\paragraph{\textbf{Power systems baselines.}} (15) $\text{HS}^{3} \text{M}$ \cite{DBLP:journals/tsg/ZhouAS18} combines unlabeled data and partly labeled data in a large margin learning objective, which is a classical method for event detection in the power systems. It incorporates partial knowledge for event detection. (16) SS-LOF \cite{liu2019data} is an event detection algorithm in the power systems using reduced data and local outlier factor for detecting the events in power systems. It is employed to determine the region of the event source.

\section{Experiments and Results}\label{sec:Experiments}

\subsection{Experimental Results}
In order to verify the effectiveness of MC-TED, the latest CNN-based, LSTM-based, graph-based, and BERT-based event detection models are selected as the baseline models. We perform event detection on PoE, MAVEN, and ACE 2005 datasets and report the average results with the standard deviation across the 10-folds cross-validation in Table \ref{ACE_MAVEN} and \ref{PoE}. Part of the results are from the original papers \cite{DBLP:conf/acl/ChenXLZ015, DBLP:conf/emnlp/YanJMGC19, DBLP:conf/naacl/WangHLSL19, DBLP:conf/acl/LouLDZC20, DBLP:conf/emnlp/Cui0L0WS20, DBLP:conf/emnlp/WangWHJHLLLLZ20}.

\begin{table*}[t]
\caption{Event detection results on our electrical power dataset PoE and PoE-Multi. The PoE-Multi means the sentences set that contain multiple events in a sentence. Our approach performs best, even when only considering multiple event sentences, which shows MC-TED is the best choice in PoE.}
\centering
\renewcommand\arraystretch{1.2}
\resizebox{\linewidth}{!}{
\begin{tabular}{l|ccc|ccc}
\toprule
 \textbf{Datasets} & \multicolumn{3}{c}{\textbf{PoE (Our Dataset)}}  & \multicolumn{3}{c}{\textbf{PoE-Multi (only multiple event sentences)}} \\ 
\textbf{Metric} & F1 (\%) & Precision (\%) & Recall (\%) & F1 (\%) & Precision (\%) & Recall (\%)  \\ 
\midrule
CNN  &61.03 $\pm$ 1.24 & 59.17 $\pm$ 1.71 & 63.02 $\pm$ 1.10  &55.23 $\pm$ 0.21 & 54.19 $\pm$ 1.01 & 58.63 $\pm$ 1.34  \\ 
DMCNN \cite{DBLP:conf/acl/ChenXLZ015}  &62.34 $\pm$ 1.53 & 59.83 $\pm$ 1.57 & 64.27 $\pm$ 1.93  &55.83 $\pm$ 1.21 & 57.39 $\pm$ 0.94 & 58.30 $\pm$ 0.74  \\  \hline
BiLSTM & 62.50 $\pm$ 1.66 & 56.98 $\pm$ 1.72 & 65.89 $\pm$ 1.56  & 54.83 $\pm$ 1.02 &  53.39 $\pm$ 0.26 &  57.73 $\pm$ 1.07 \\
MLBiNet (2-layer) \cite{DBLP:conf/acl/LouLDZC20} & 62.50 $\pm$ 1.66 & 56.98 $\pm$ 1.72 & 65.89 $\pm$ 1.56 & 60.78 $\pm$ 1.12 & 54.89 $\pm$ 1.20 & 61.83 $\pm$ 1.04\\ \hline
GCN~\cite{DBLP:conf/iclr/KipfW17} &60.36 $\pm$ 1.13 & 61.14 $\pm$ 1.09 & 59.60 $\pm$ 1.54  &57.48 $\pm$ 1.30 & 57.89 $\pm$ 1.98 & 54.38 $\pm$ 1.48  \\ 
GAT~\cite{DBLP:conf/iclr/VelickovicCCRLB18}  & 58.88 $\pm$ 1.47 &57.43 $\pm$ 1.93 & 60.42 $\pm$ 1.27 & 54.89 $\pm$ 2.73  &56.84 $\pm$ 2.83 & 57.38 $\pm$ 1.03 \\ 
MOGANED \cite{DBLP:conf/emnlp/YanJMGC19} & 61.95 $\pm$ 1.78 & 58.94 $\pm$ 1.73 & 64.84 $\pm$ 1.34 & 58.87 $\pm$ 0.84  & 56.48 $\pm$ 0.92  & 59.89 $\pm$ 1.26  \\
GatedGCN \cite{DBLP:conf/emnlp/LaiNN20} &  61.83 $\pm$ 1.92 &  60.73 $\pm$ 1.34  &  65.89 $\pm$ 1.92  & 59.34 $\pm$ 1.43 & 58.72 $\pm$ 2.04 &  60.73 $\pm$ 1.44 \\
EE-GCN \cite{DBLP:conf/emnlp/Cui0L0WS20}   & 62.46 $\pm$ 1.17 & 57.41 $\pm$ 1.42 & 67.80 $\pm$ 1.21 &60.02 $\pm$ 2.32 & 56.37 $\pm$ 1.93& 60.78 $\pm$ 2.34 \\  \hline
BERT+CRF  & 63.48 $\pm$ 1.05 & 61.13 $\pm$ 1.52 & 67.48 $\pm$ 1.81 &60.37 $\pm$ 0.83  & 61.08 $\pm$ 0.89  & 63.92 $\pm$ 0.38  \\ 
BERT+BiLSTM &  63.23 $\pm$ 1.87  & 61.83 $\pm$ 1.30 & 66.83 $\pm$ 1.93 &  61.73 $\pm$ 2.84  &  60.47 $\pm$ 1.93 &  63.78 $\pm$ 0.47 \\ 
BERT+CNN & 63.21 $\pm$  1.92 & 61.83 $\pm$ 0.37 & 66.98 $\pm$ 1.32 & 61.87 $\pm$ 1.37 & 60.29 $\pm$ 1.93 &  63.37 $\pm$ 2.29 \\ 
DMBERT  \cite{DBLP:conf/naacl/WangHLSL19}  & 63.56 $\pm$ 1.72 & 62.21 $\pm$ 1.78 & 67.70 $\pm$ 1.33 & 61.28 $\pm$ 0.73  & 59.48 $\pm$ 1.34 & 64.28 $\pm$ 1.27 \\  \hline

$\text{HS}^{3}\text{M}$  \cite{DBLP:journals/tsg/ZhouAS18}& 60.23 $\pm$ 1.21 & 57.95 $\pm$ 1.25 & 64.93 $\pm$ 1.75 & 56.32 $\pm$ 1.02  & 56.76 $\pm$ 1.52 & 56.69 $\pm$ 1.59 \\ 

SS-LOF \cite{liu2019data} & 62.53 $\pm$ 1.17 & 59.65 $\pm$ 1.83 & 62.90 $\pm$ 1.76 & 58.25 $\pm$ 0.70  & 58.19 $\pm$ 1.54 & 55.29 $\pm$ 1.65 \\\hline

\rowcolor{Gray}  \textbf{MC-TED (Ours)} & \textbf{65.58 $\pm$ 1.24} &	\textbf{63.71  $\pm$ 1.29} &	\textbf{69.93 $\pm$ 1.61} & \textbf{62.82 $\pm$ 1.23} &  \textbf{61.03 $\pm$ 1.72}   & \textbf{65.38 $\pm$ 1.29}   \\
\bottomrule
\end{tabular}
}

\label{PoE}
\end{table*}

\begin{table*}[t]
\caption{Event detection results on general datasets MAVEN and ACE 2005. Our model achieves the best performance on F1, which demonstrates the universality of our method.}
\centering
\renewcommand\arraystretch{1.2}
\resizebox{\linewidth}{!}{
\begin{tabular}{l|ccc|ccc}
\toprule
 \textbf{Datasets} &  \multicolumn{3}{c}{\textbf{MAVEN}} & \multicolumn{3}{c}{\textbf{ACE 2005}} \\ 
\textbf{Metric} & F1 (\%) & Precision (\%) & Recall (\%) & F1 (\%) & Precision (\%) & Recall (\%) \\ 
\midrule
DMCNN \cite{DBLP:conf/acl/ChenXLZ015}   &60.60 $\pm$ 0.20 & 66.30 $\pm$ 0.89 & 55.90 $\pm$ 0.50 &  68.00 $\pm$ 1.95 & 73.70 $\pm$ 2.42 &63.30 $\pm$ 3.30 \\ 
MLBiNet (2-layer) \cite{DBLP:conf/acl/LouLDZC20}  & 63.69 $\pm$ 1.29 & 62.52 $\pm$ 1.61 & 66.91 $\pm$ 2.75 & 74.60 $\pm$ 1.82 & 72.62 $\pm$ 1.35 & 76.74 $\pm$ 1.83 \\ 
MOGANED \cite{DBLP:conf/emnlp/YanJMGC19}  & 63.80 $\pm$ 0.18  & 63.40 $\pm$ 0.88  & 64.10 $\pm$ 0.90  & 72.10 $\pm$ 0.39   & 70.40  $\pm$ 1.38  &73.90 $\pm$ 2.24  \\
GatedGCN \cite{DBLP:conf/emnlp/LaiNN20}  & 65.82 $\pm$ 1.37 & 63.72 $\pm$ 1.73 & 67.78 $\pm$ 1.83 & 73.40 $\pm$ 1.89 &76.70 $\pm$ 1.48 &70.50 $\pm$ 1.72 \\
EE-GCN \cite{DBLP:conf/emnlp/Cui0L0WS20}    &66.92 $\pm$ 1.53 & 64.29 $\pm$ 1.67& 69.82 $\pm$ 1.38 & 74.84 $\pm$ 1.87 & \textbf{75.84 $\pm$ 1.92} &72.93 $\pm$ 1.62 \\ 
BERT+CRF  &67.80 $\pm$ 0.15  & 65.00 $\pm$ 0.84  & 70.90 $\pm$ 0.94 & 74.10 $\pm$ 1.56 & 71.30 $\pm$ 1.77 &77.10 $\pm$ 1.99 \\ 
DMBERT  \cite{DBLP:conf/naacl/WangHLSL19}   & 67.10 $\pm$ 0.41  & 62.70 $\pm$ 1.01 & \textbf{72.30 $\pm$ 1.03} & 74.30 $\pm$ 0.81 & 70.20 $\pm$ 1.71 & \textbf{78.90 $\pm$ 1.64}  \\  \hline
 \rowcolor{Gray} \textbf{MC-TED (Ours)} & \textbf{68.47 $\pm$ 1.04} &  \textbf{66.55 $\pm$ 1.43}   & 70.09 $\pm$ 1.78 & \textbf{75.26  $\pm$ 1.53}  &  74.62 $\pm$ 1.73  &  73.38  $\pm$ 1.82  \\
\bottomrule
\end{tabular}
}

\label{ACE_MAVEN}
\end{table*}

MC-TED consistently outperforms all comparison algorithms on the electrical power datasets (PoE), as well as only having multiple event sentences on the PoE in Table \ref{PoE}. 
Compared with CNN-based models (CNN and DMCNN), MC-TED achieves $4.55\%$ and $3.24\%$ improvements in terms of the F1-score on PoE, and promotes at least $6\%$ points on only multiple event sentences, respectively. 
It should be contributed to semantic dependencies modeling and syntactic topology preserving ability of the proposed model. 
Compared with BERT+CRF, BERT+BiLSTM, BERT+CNN, and DMBERT, MC-TED improves the F1-score by at least $2\%$ on PoE and $1\%$ on only multiple event sentences through modeling multi-channel graphs catching more semantic information in a complex context.
MC-TED models the heterogeneity of the graph compared with other graph-based models. It fully considers the semantic information of multiple node types and relations and performs weighted fusion, achieving better results.

Table \ref{ACE_MAVEN} shows the overall results of each approach per evaluation dataset using uniform parameter settings in Section~\ref{details}. 
MC-TED achieves a significant improvement in most cases on MAVEN and ACE 2005. 
Compared with DMCNN and MLBiNet, our model consistently outperforms in both ACE 2005 and MAVEN on F1, precision and recall. Thus, our method performed significantly better than the CNN-based and LSTM-based methods, which learns word representation from topological and semantic perspectives.
Compared with BERT+CRF and DMBERT, MC-TED has the best performance and uses significantly fewer word representation dimensions through building graph for each sentence.
Compared with MOGANED \cite{DBLP:conf/emnlp/YanJMGC19}, 
GatedGCN \cite{DBLP:conf/emnlp/LaiNN20} and
EE-GCN \cite{DBLP:conf/emnlp/Cui0L0WS20}, three recent graph neural network models, our model respectively achieves $4.67\%$, $2.65\%$ and $1.55\%$ improvements on the F1-score on MAVEN, and improves at least $0.42\%$ points on ACE 2005. It shows that our method is significantly superior to the graph-based methods due to utilizing node and edge types information.

Specifically, compared with EE-GCN, it achieves $3.12\%$, $1.55\%$, and $0.42\%$ improvements of F1-score on PoE, MAVEN, and ACE 2005. It demonstrates that MC-TED fully accounts for the rich semantics reflected by multiple node types and relations. Therefore, the node representation is more comprehensive and differentiated. GCN, GAT, EE-GCN, MOGANED, MLBiNet, GatedGCN, and MC-TED consider the importance of neighbors in learning node representations, and the performance of our model 
is more competitive due to its ability to capture multiple semantic information on multi-channel heterogeneous graphs.

It should be noted that the improvement of MC-TED is more significant on PoE dataset. It improves at least 1.5 points on the three evaluation metrics.
The major reason is that PoE is a Chinese power field dataset, which increases the difficulty of syntactic analysis and type recognition.
Therefore, even though MC-TED can fuse multiple node types and relations types, its improvement is not obvious because ACE 2005 does not provide validity information. Therefore, for Precision and Recall evaluation metrics, our model is not optimal. It has the best performance in F1 evaluation metric.
Taken together, MC-TED achieved peak performance across all datasets, demonstrating the need to model relation-type-aware graph, node-type-aware graph as well as semantic-aware graph.

\subsection{Ablation Study on Three Graphs}
For further evaluation on the components of MC-TED, we conduct the ablation experiments on PoE and report the results in Table~\ref{Ablation}.
 
\textbf{Topological channel.}
Both the relation-type-aware graph and the word-type-aware graph are vitally essential for the type utilized multi-channel GNN.
As shown in Table~\ref{Ablation}, on the one hand, compared with the full MC-TED, removing the relation-type-aware graph (G1 in Table~\ref{Ablation}) results in low performance (e.g., a $2.10\%$ decrease in terms of F1 score). 
On the other hand, considering inside the topological channel, compared with only applying the word-type-aware (G2 in Table~\ref{Ablation}), G1 plus G2 makes an improvement of $1.99\%$, implying the relation-type-aware graph contributes to the topological information capture. 
The conclusion is consistent for the word-type-aware graph. The main reason is that it achieves a more detailed and comprehensive description considering the semantics of word and relation types. 
Furthermore, we replace the two graphs with a single GCN and use a homogeneous graph (i.e., without any word type information and relation type information, G0 in Table~\ref{Ablation}) as its input. 
The results show a $2.94\%$ decrease after the replacement and a $2.68\%$ decrease when only applying the semantic channel, which reflect the importance of the topological channel with type utilizing multi-channel GNN. It may because that the topological channel designs two different heterogeneous graphs utilizing word dependencies and part-of-speech tags. The relation-type-aware graph updates the edge representation in each model layer, making the model focus on learning edge information. All the observation indicate that our model through utilizing the relation type improves the event detection performance.

\begin{table}[t]
\small
\caption{Ablation study of topological channel and semantic channel. The bold values are the best result and the underlined values are the second best result.}
\centering
\renewcommand\arraystretch{1.2}
\resizebox{0.9\linewidth}{!}{
\begin{tabular}{c|ccc}
\toprule
 \textbf{Method} &  F1 (\%) & Precision (\%) & Recall (\%)  \\
\midrule
 \textbf{Relation-type-aware graph (G1)} &  62.94 $\pm$ 1.25   & 59.00 $\pm$ 1.91  & 67.45 $\pm$ 1.65 \\ 
 \textbf{Word-type-aware graph (G2)}   & 61.95 $\pm$ 1.93   &  59.91 $\pm$ 1.26 & 64.14 $\pm$ 1.38  \\ 
 \textbf{Semantic-aware graph (G3)}  & 62.90 $\pm$ 1.33 &  57.42 $\pm$ 1.33 & 69.53 $\pm$ 1.25  \\ \midrule
 \textbf{Homogeneous graph (G0)+G3} & 62.64 $\pm$ 1.15 & 59.48 $\pm$ 1.64 & 66.15 $\pm$ 1.48 \\
\textbf{$\text{G1+G2}$} &  63.94 $\pm$ 1.62   & \textbf{64.77 $\pm$ 1.52}  & 63.13 $\pm$ 1.38  \\ 
\textbf{$\text{G1+G3}$}  & \underline{64.07 $\pm$ 1.12}  &  62.27 $\pm$ 1.79 & \textbf{70.25 $\pm$ 1.61} \\ 
 \textbf{$\text{G2+G3}$}   & 63.48 $\pm$ 1.42   &  60.18 $\pm$ 1.92 & 67.17 $\pm$ 1.43 \\ \midrule
 \textbf{$\text{G1+G2+G3}$}  & \textbf{65.58 $\pm$ 1.24} &	\underline{63.71 $\pm$ 1.29} &	\underline{69.93 $\pm$ 1.61} 	 \\
\bottomrule
\end{tabular}
}
\label{Ablation}
\end{table}



\begin{table}[t]
\small
\caption{Ablation study of topological channel. The bold values are the best result and the underlined values are the second best result. }
\centering
\renewcommand\arraystretch{1.2}
\resizebox{0.95\linewidth}{!}{
\begin{tabular}{l|ccc}
\toprule
 \textbf{Method} &  F1 (\%) & Precision (\%) & Recall (\%)  \\
\midrule
 \textbf{MC-TED (only Topological Channel)} &  \textbf{63.94 $\pm$ 1.62}   & \underline{64.77 $\pm$ 1.52}  & 63.13 $\pm$ 1.38   \\ 
{  }{  } \textbf{w/o relation type module} & 62.82 $\pm$ 1.39 & 62.78 $\pm$ 1.23 & 62.20 $\pm$ 1.89 \\
{  }{  } \textbf{w/o node type module} &  \underline{63.28 $\pm$ 1.90}   & 64.29 $\pm$ 2.14  & 63.01 $\pm$ 1.29  \\ 
{  }{  } \textbf{w/o relation and node types module}  & 61.83 $\pm$ 1.27  &  62.51 $\pm$ 1.43 & \underline{64.10 $\pm$ 1.03} \\ \midrule
 {  }{  } \textbf{w/o relation type learning module} & 63.14 $\pm$ 1.19 & \textbf{65.94 $\pm$ 2.37} & 63.26 $\pm$ 1.43 \\
 {  }{  } \textbf{w/o node type learning module} &  62.39 $\pm$ 1.04   & 63.30 $\pm$ 1.58  & 63.18 $\pm$ 1.13  \\ 
{  }{  } \textbf{w/o relation and node types learning module}  & 62.02 $\pm$ 2.05  &  62.21 $\pm$ 1.61 & \textbf{64.47 $\pm$ 1.75} \\ 
\bottomrule
\end{tabular}
}
\label{Ablation2}
\end{table}

\textbf{Semantic channel.}
The semantic-aware graph helps capture richer semantic information beyond the original syntactic topology. From Table~\ref{Ablation}, it is noticed that the performance drops by $1.64\%$ in terms of F1 score when not including the semantic-aware graph (G3 in Table~\ref{Ablation}). It demonstrates that the semantic channel can establish relationships between words that are far apart and grammatically independent but have similar meanings. The structure of the semantic-aware graph dynamically changes during the training process. Furthermore, after adding the semantic-aware graph, the performance of only applying the relation-type-aware graph or word-type-aware graph increases by $1.13\%$ and $1.53\%$ respectively. Therefore, the semantic channel improves the superiority of MC-TED, independent of the topological channel. It also demonstrates that the semantic channel refines textual representations with semantic similarity, building the semantic information interaction among potential event-related words.

All the observations demonstrate the effectiveness of two channel in our model. With the two channels, MC-TED integrates multiple aspects of information to learn a more comprehensive word representation through fusing multiple sources of information. The semantic channel reflects the similarity of words through adding relation with lower cosine distance.
The topological channel constructs a relation-type-aware graph and a word-type-aware graph to learn more comprehensive node representations.
They consider the word type and relation type to portray nonlinear relationships with stronger representative and inference capabilities fully.

\subsection{Ablation Study on Topological Channel}
We evaluate variants on topological channel of our approach, as shown in Table \ref{Ablation2}. We show the experimental results and observe that:

\textbf{Type module.}
We exclude the relation type module of relation-type-aware graph for the type utilized multi-channel GNN.
As a consequence, excluding the module decrease the F1-scores by $1.12\%$, as well as precision by $1.99\%$, and recall by $0.97\%$.
Thus, the F1-scores decrease in all metrics, thereby showing that the relation type module could significantly improve the performance. 
The relation type module positively affected the learning of word relationship, and relation type knowledge is helpful for event detection with the type utilized multi-channel GNN.
We further exclude the node type module, which improves event detection by utilizing node type knowledge.
The decreases for the F1, precision and recall are $0.66\%$, $0.48\%$, and $0.12\%$, respectively, with significant effects on performance.
When both the relation type and node type are excluded, the performance is decreased dramatically.
These findings indicate that relation type module in G1 and word type module in G2 can effectively improve event detection.

\textbf{Type learning module.}
The topological channel has relation type learning and node type learning module for learning more appropriate word representations according to context. It is designed for weaken word and relation type errors in the electric power texts caused by the terminology vocabulary.  
When the relation type learning module is excluded, the F1-score is decreased by $0.80\%$. Compared with removing the whole relation type module, the performance is increased which demonstrates the validity of adding relation type.
We further exclude both relation and node type learning modules, the F1-score decreases $1.92\%$ points.
These results suggest that all of the modules are useful, and that the relation type module is the most important for event detection because excluding them dramatically degraded the performance.

\begin{table}[t]
\centering
\caption{Evaluation for variants on PoE. The bold values are the best result and the underlined values are the second best result.}
\renewcommand\arraystretch{1.2}
\resizebox{\linewidth}{!}{
\begin{tabular}{lll|ccc}
\toprule
 \textbf{Initial Representation} & \textbf{Encoder} &\textbf{Aggregation}   & F1 (\%) & Precision (\%)   & Recall (\%)  \\
\midrule
\multirow{6}{*}{\textbf{Random}}& \multirow{2}{*}{\textbf{GAT}} & Concatenation   &62.38 $\pm$ 1.66 &61.28  $\pm$ 1.98  & 65.21  $\pm$ 1.46   \\
&  &  Weighted  &60.02  $\pm$ 1.62 &58.24 $\pm$ 1.20 &62.32 $\pm$ 1.74   \\ \cline{2-6}
   & \multirow{2}{*}{\textbf{GCN}} & Concatenation   & 61.19 $\pm$ 1.08  &60.23  $\pm$ 1.60 & 62.79  $\pm$ 1.04 \\
&  &  Weighted   &61.76  $\pm$ 1.01  &60.23 $\pm$ 1.01   &59.37 $\pm$ 1.63  \\ \cline{2-6}
  & \multirow{2}{*}{\textbf{BERT}} & Concatenation   & 62.53 $\pm$ 1.12  &60.93  $\pm$ 1.29 & 66.93  $\pm$ 1.32 \\
&  &  Weighted   &63.83  $\pm$ 1.27  &61.92 $\pm$ 1.72   &67.28 $\pm$ 1.28  \\\hline

\multirow{6}{*}{\textbf{Random+BiLSTM}}& \multirow{2}{*}{\textbf{GAT}} & Concatenation   &60.87 $\pm$ 1.76 &61.39 $\pm$ 1.03   &64.29 $\pm$ 1.84    \\ 
&  &  Weighted    &60.06 $\pm$ 1.44  &60.90 $\pm$ 1.37    &63.05 $\pm$ 1.17   \\\cline{2-6}
 & \multirow{2}{*}{\textbf{GCN}} & Concatenation & 59.64 $\pm$ 1.81   &60.45 $\pm$ 1.10  & 62.82   $\pm$ 1.39   \\
&  &  \textbf{Weighted (Ours)}    &\textbf{65.58 $\pm$ 1.24} &	\textbf{63.71  $\pm$ 1.29}    &	\textbf{69.93 $\pm$ 1.61}  \\\cline{2-6}
 & \multirow{2}{*}{BERT} & Concatenation & 63.82 $\pm$ 1.01   &60.92 $\pm$ 1.73  & 66.38   $\pm$ 1.83   \\
&  &  Weighted    &63.23 $\pm$ 1.87 &	61.83  $\pm$ 1.30    &	66.83 $\pm$ 1.93  \\\hline

\multirow{6}{*}{\textbf{Glove}}& \multirow{2}{*}{\textbf{GAT}} & Concatenation    &60.76 $\pm$ 1.19 &59.74 $\pm$ 1.29   & 62.22 $\pm$ 1.98  \\
&  &  Weighted   &60.20 $\pm$ 1.92  &58.20 $\pm$ 1.76  &62.72 $\pm$ 1.93  \\\cline{2-6}
 & \multirow{2}{*}{\textbf{GCN}} & Concatenation  & 59.32 $\pm$ 1.90  &62.36 $\pm$ 1.81  & 61.01 $\pm$ 1.89 \\
&  &  Weighted   &62.14 $\pm$ 1.25   &57.18 $\pm$ 1.53   &56.67 $\pm$ 1.65  \\\cline{2-6}
 & \multirow{2}{*}{\textbf{BERT}} & Concatenation  & 63.34 $\pm$ 1.74  &62.03 $\pm$ 1.24  & 67.38 $\pm$ 1.74 \\
&  &  Weighted   &62.93 $\pm$ 1.83   &60.35 $\pm$ 1.39   &66.87 $\pm$ 1.39  \\\hline

\multirow{6}{*}{\textbf{Glove+BiLSTM}}& \multirow{2}{*}{\textbf{GAT}} & Concatenation  &63.85 $\pm$ 1.52 &60.13 $\pm$ 1.39   &61.71 $\pm$ 1.21    \\ 
&  &  Weighted   &60.91 $\pm$ 1.63 &53.52 $\pm$ 1.94     &61.62 $\pm$ 1.75   \\\cline{2-6}
 & \multirow{2}{*}{\textbf{GCN}} & Concatenation   & 59.16 $\pm$ 1.51   &60.73 $\pm$ 1.09   & 59.58 $\pm$ 1.02 \\
&  &  Weighted   &62.55 $\pm$ 1.14   &\underline{62.64 $\pm$ 1.12}  & 61.01 $\pm$ 1.60 \\\cline{2-6}
 & \multirow{2}{*}{\textbf{BERT}} & Concatenation   & 63.04 $\pm$ 1.37   &60.65 $\pm$ 1.63   & 65.26 $\pm$ 1.35 \\
&  &  Weighted   &63.84 $\pm$ 1.84   &60.74 $\pm$ 1.63  & 67.83 $\pm$ 1.34 \\\hline

\multirow{6}{*}{\textbf{BERT}}& \multirow{2}{*}{\textbf{GAT}} & Concatenation    &61.68 $\pm$ 1.25 &60.28 $\pm$ 1.29   & 65.12 $\pm$ 1.78   \\
&  &  Weighted   &60.90 $\pm$ 1.84  &58.49 $\pm$ 1.02  &59.94 $\pm$ 1.82  \\\cline{2-6}
& \multirow{2}{*}{\textbf{GCN}} & Concatenation   & 62.96 $\pm$ 1.53  &61.37 $\pm$ 1.35 & 66.40 $\pm$ 1.06  \\
&  &  Weighted   &61.61 $\pm$ 1.10   &60.46 $\pm$ 1.30  &65.50 $\pm$ 1.43   \\\cline{2-6}
& \multirow{2}{*}{\textbf{BERT}} & Concatenation   & 62.84 $\pm$ 1.23  &61.05 $\pm$ 1.35 & 66.92 $\pm$ 1.83  \\
&  &  Weighted   &63.48 $\pm$ 1.05   &61.13 $\pm$ 1.52  &\underline{67.48 $\pm$ 1.81}   \\\hline

\multirow{6}{*}{\textbf{BERT+BiLSTM}} & \multirow{2}{*}{\textbf{GAT}} & Concatenation      &63.64 $\pm$ 1.83 &60.38 $\pm$ 1.23  &61.70 $\pm$ 1.87  \\ 
&  &  Weighted     &61.74 $\pm$ 1.64 &59.53 $\pm$ 1.92  &62.76 $\pm$ 1.85    \\\cline{2-6}
 & \multirow{2}{*}{\textbf{GCN}} & Concatenation   & 59.60 $\pm$ 1.34   &60.63 $\pm$ 1.29   & 63.01 $\pm$ 1.54 \\
&  &  Weighted   &\underline{64.16 $\pm$ 1.81}   &60.25 $\pm$ 1.62  & 64.17 $\pm$ 1.01  \\\cline{2-6}
 & \multirow{2}{*}{\textbf{BERT}} & Concatenation   & 61.63 $\pm$ 1.53   &60.74 $\pm$ 1.26   & 66.26 $\pm$ 1.25 \\
&  &  Weighted   &63.06 $\pm$ 1.03   &61.24 $\pm$ 1.53  & 66.42 $\pm$ 1.32  \\
\bottomrule
\end{tabular}
}
\label{disscuss}
\end{table}

\subsection{Discussion on Generalized Variants}
To evaluate the effectiveness and generality of components, we experiment various variants on our dataset, as shown in Table \ref{disscuss}.
Specifically, for word initial representation, we employ six models: Random, Random+BiLSTM, Glove, Glove+BiLSTM, BERT and BERT+BiLSTM. It consists of two branches, that is whether to use BiLSTM.
For the encoder, we apply GAT, GCN and BERT to learn word embedding. 
For the aggregation method, we adopt concatenation and average methods to merge the representation of three graphs.
We analyze the experimental results and observe that:

\textbf{Initial representation.}
In the most circumstance, using BiLSTM can boost word representation through learning context information.
Furthermore, based on different static word representation methods, the random method acquires the best performance in most instances, which utilizes significantly fewer dimensions of words than BERT.

\textbf{Encoder.}
GCN encoder is better than GAT encoder in our scheme. This is mainly because GCN structure alleviates the effect of word representation smoothing, which is more suitable for our dataset with short sentences. 
For BERT-based encoder, performance will be reduced compared to Random+BiLSTM+GCN method. The main reason is that event detection in short texts is more dependent on syntactic dependency correlation. In our baseline methods, the BERT-based method have better performance than all LSTM-based models and GNN-based models. However, in the BERT-based models, we try to combine the graph method and the model has a bad effect. It may be because the whole model is strongly dependent on BERT's representation, so it is difficult to further improve. Furthermore, the training time of BERT-based model was significantly longer than that of graph-based model. In order to meet the practical application requirements, we use graph-based model as the encoder.

\textbf{Aggregation method.}
Our weighted aggregation method is to set appropriate weights among relation-type-aware graph, word-type-aware graph and semantic-aware graph for obtaining the final representation of each word. Our concatenation aggregation method merge the representation of three graphs, which makes the word dimension is three times larger. In our scheme, the weighted method usually obtains better results than concatenation method.

\begin{figure*}[t]
    \centering
        \subfigure[Impact of label rate on F1.]{
     \includegraphics[width=4.4cm]{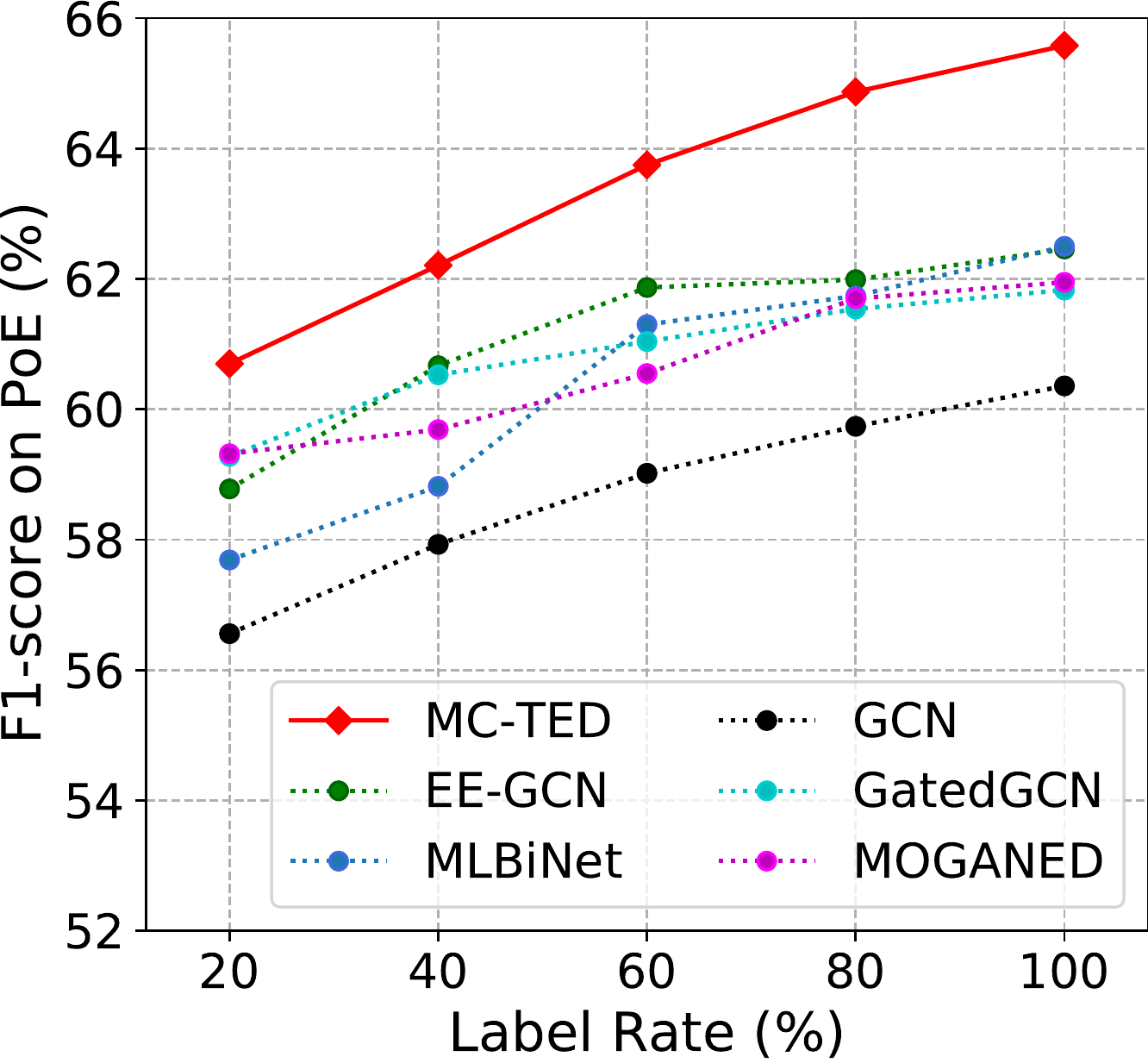}}
    \centering
        \subfigure[Impact of label rate on Precision.]{
     \includegraphics[width=4.4cm]{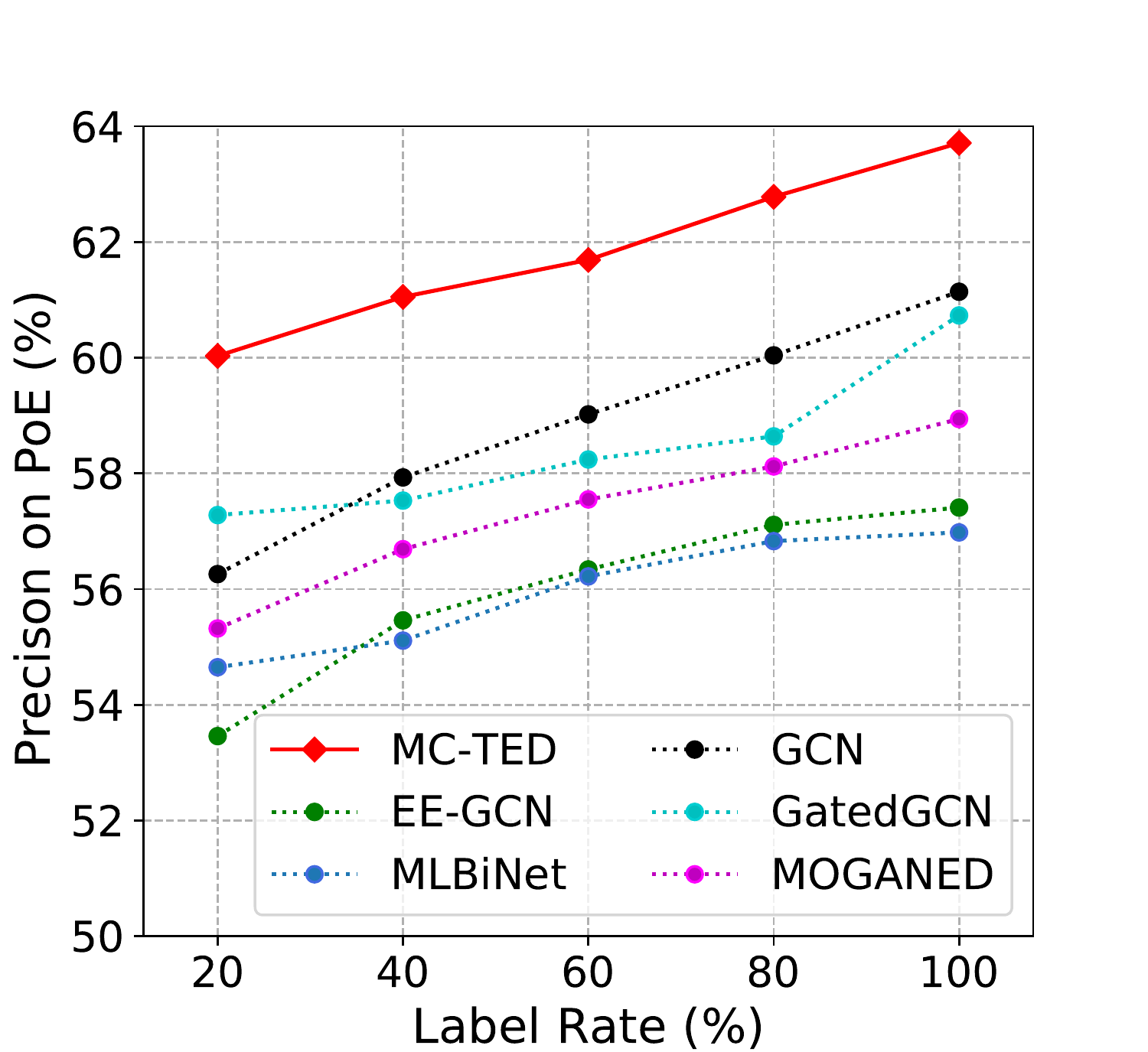}}
    \centering
        \subfigure[Impact of label rate on Recall.]{
     \includegraphics[width=4.4cm]{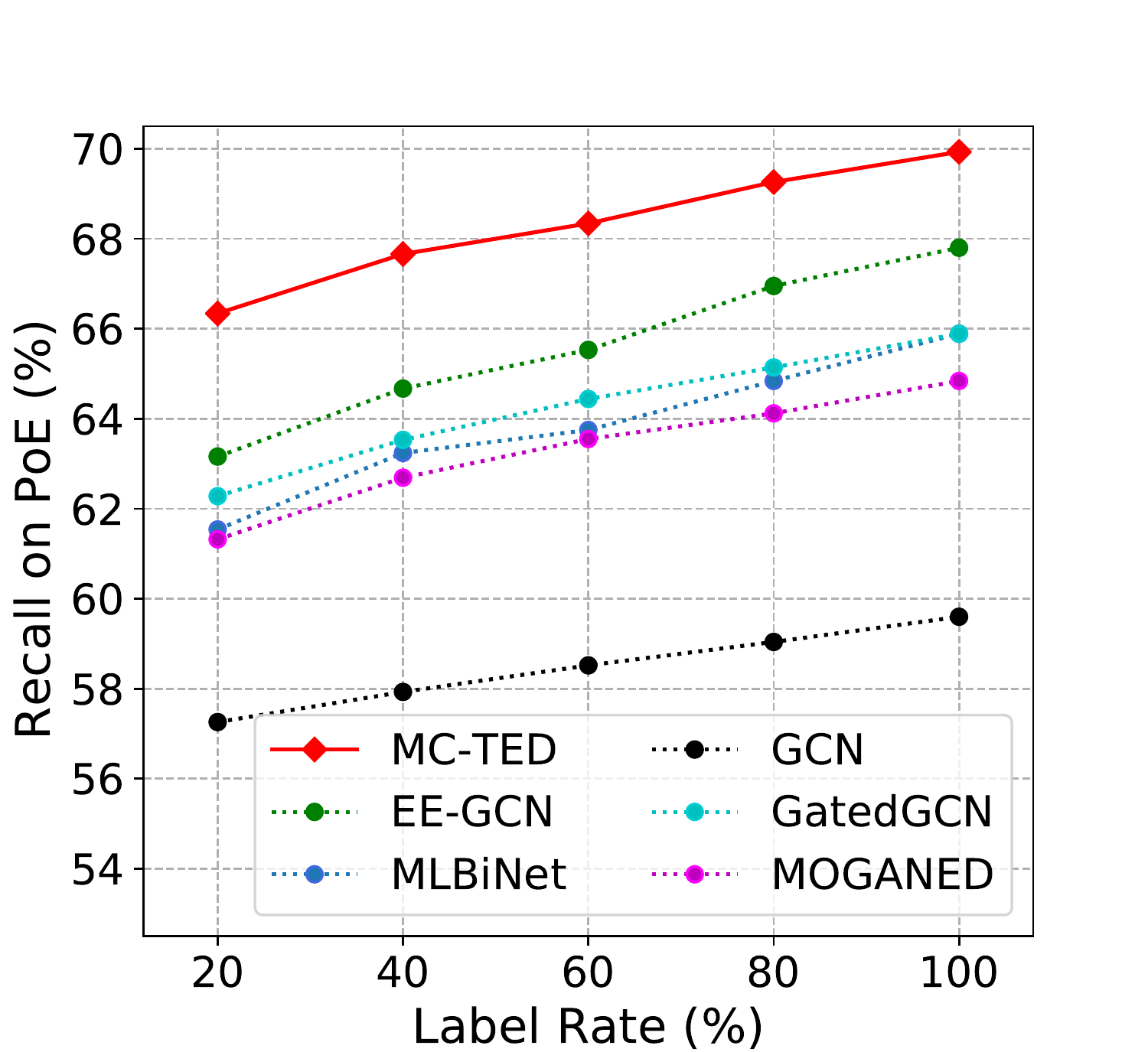}}
    \caption{Parameter study on label rate. The label rate refers to the proportion of labeled data. MC-TED preforms best at all label ratios, which demonstrates our method also works with small samples.}
    \label{rate}
\end{figure*}

\begin{figure*}[t]
    \centering
    \subfigure[Impact of model layer on F1.]{
 \includegraphics[width=4.4cm]{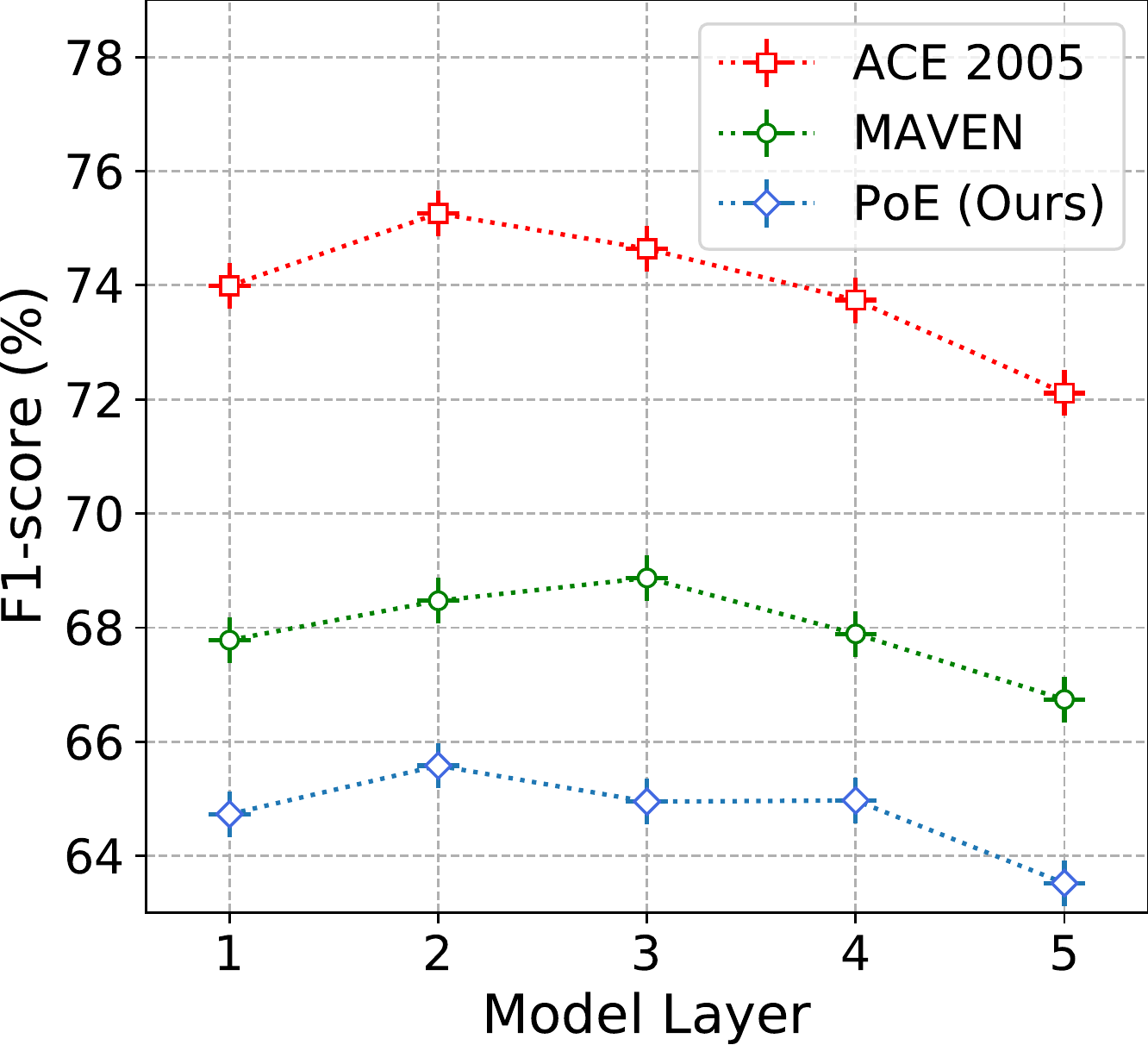}}
    \centering
    \subfigure[Impact of model layer on Precision.]{
 \includegraphics[width=4.4cm]{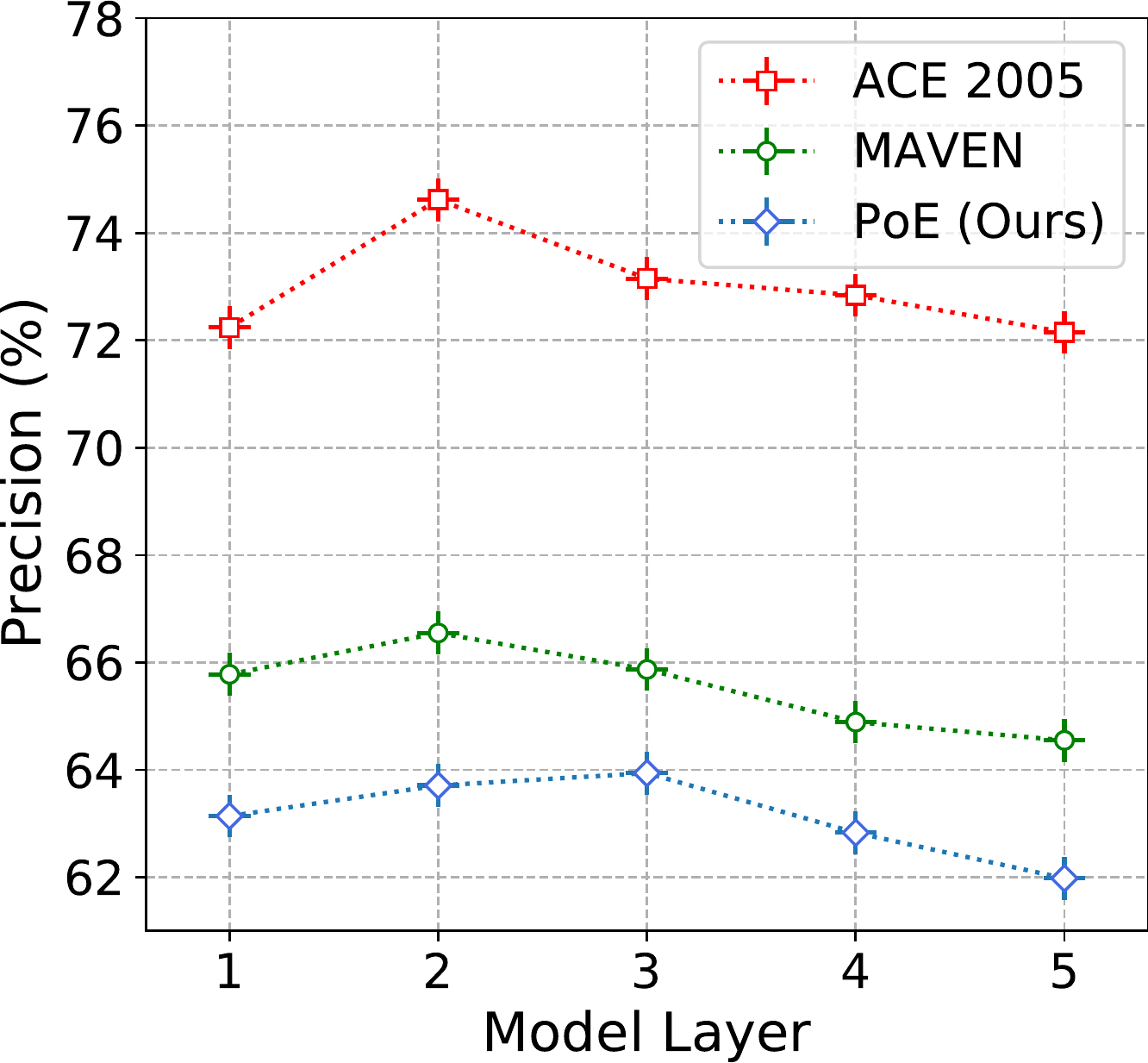}}
    \centering
    \subfigure[Impact of model layer on Recall.]{
 \includegraphics[width=4.4cm]{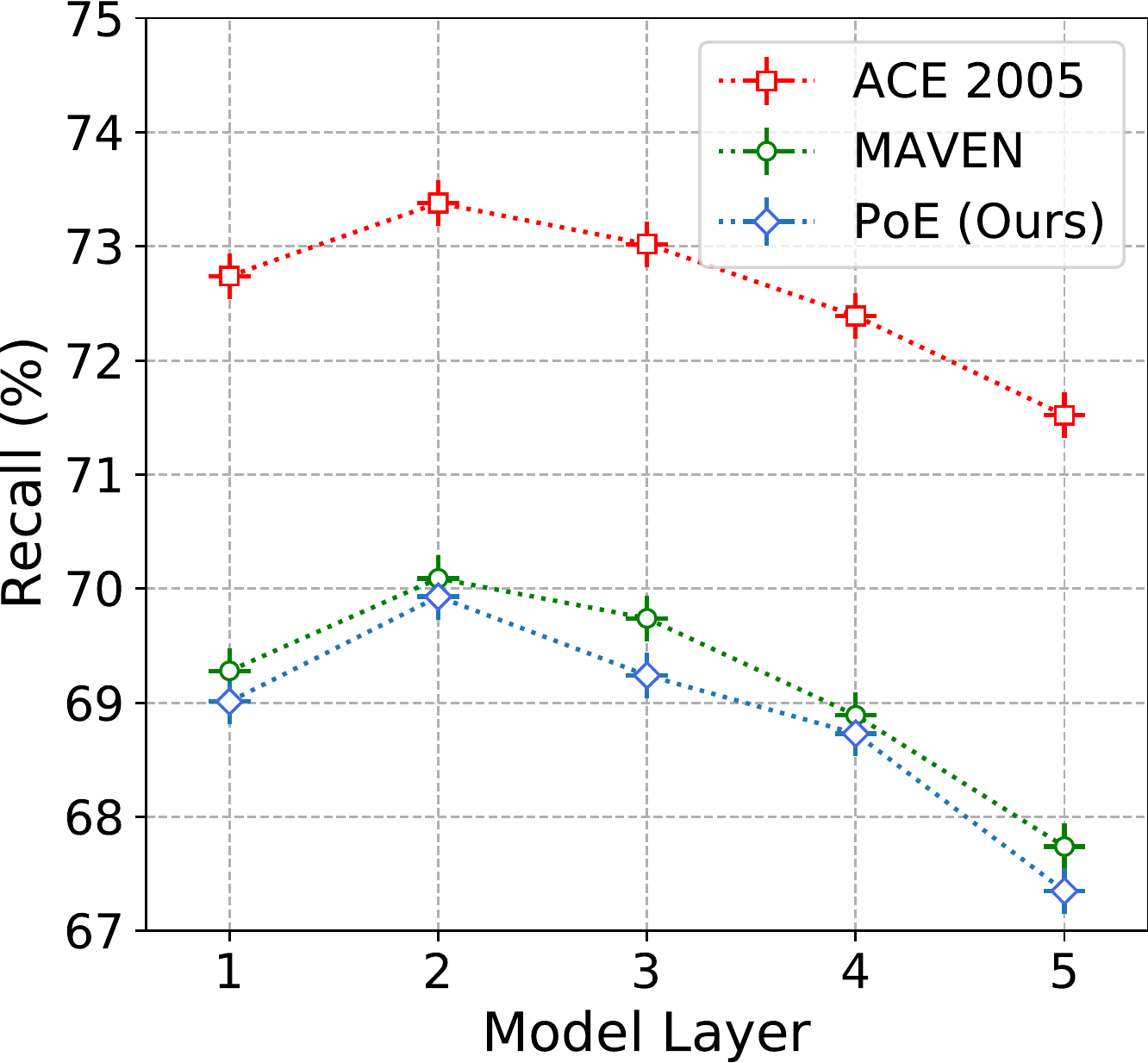}}
    \caption{Parameter study on model layer. We tested the performance of GCN layer between 1 and 5. When the number of model layers is 2, the performance reaches the peak.}
    \label{layer}
\end{figure*}


\begin{figure*}[t]
  \centering
 \subfigure[Impact of representation dimension on F1.]{
 \includegraphics[width=4.4cm]{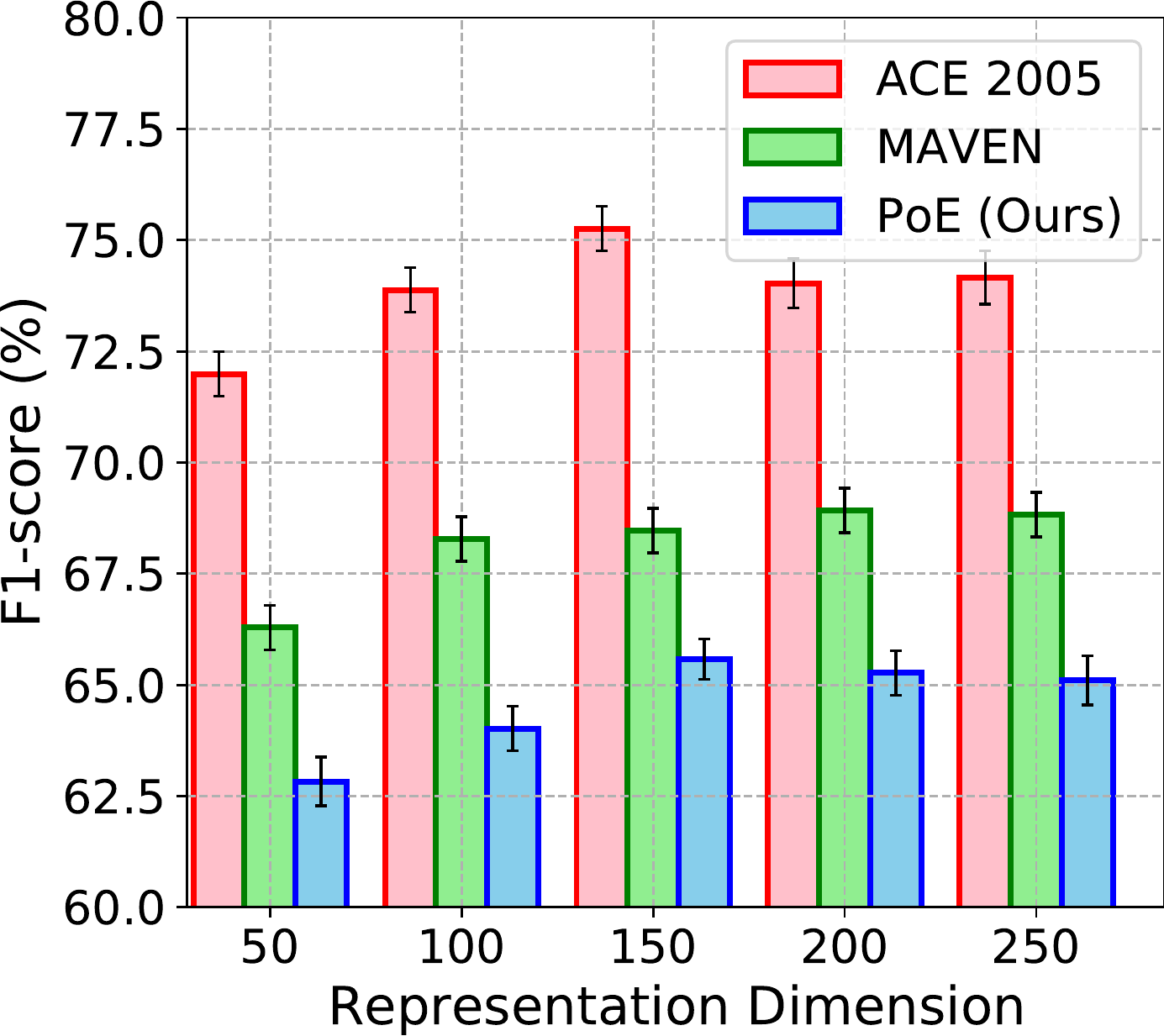}}
    \centering
 \subfigure[Impact of representation dimension on Precision.]{
 \includegraphics[width=4.4cm]{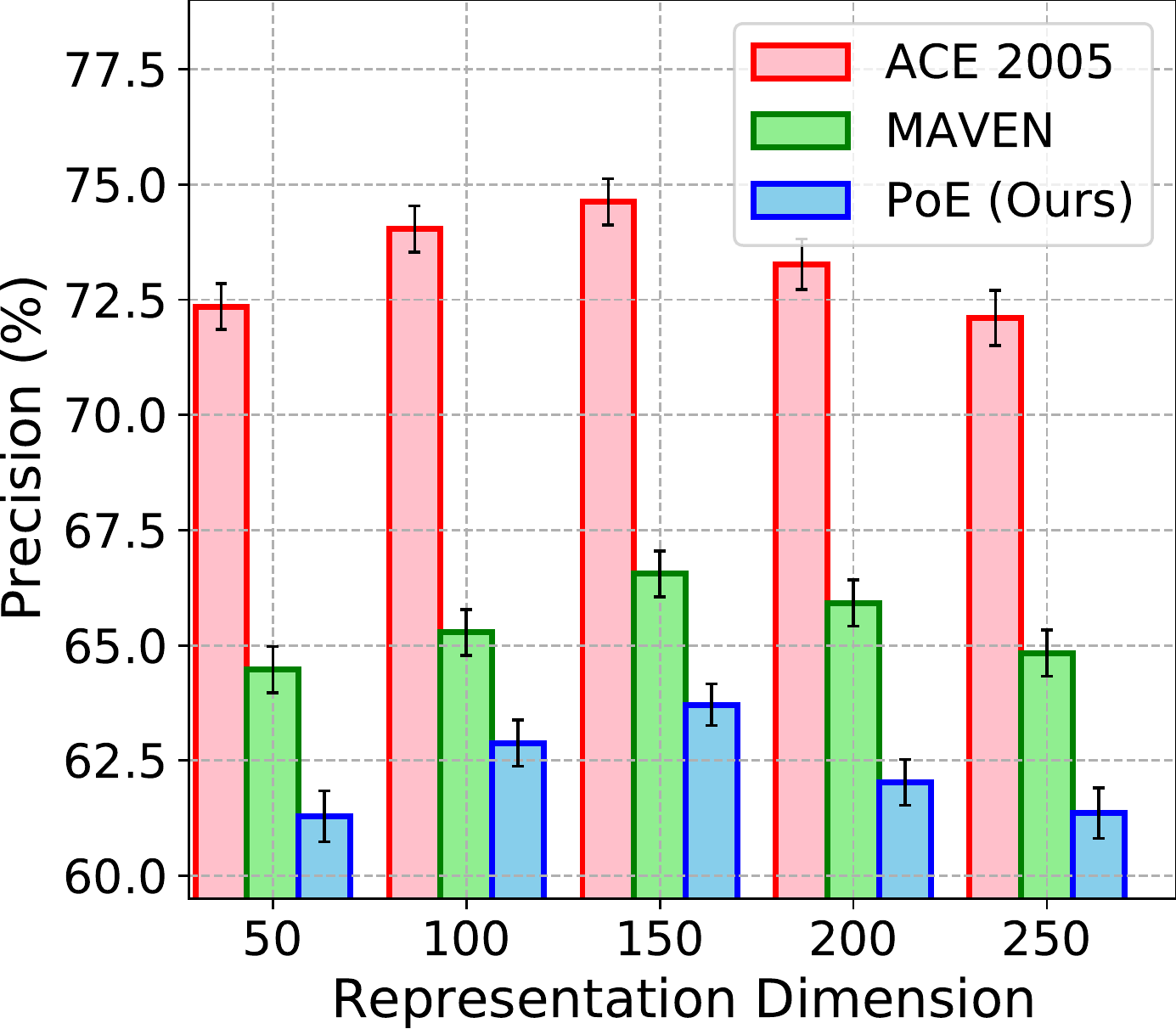}}
 \centering
 \subfigure[Impact of representation dimension on Recall.]{
 \includegraphics[width=4.4cm]{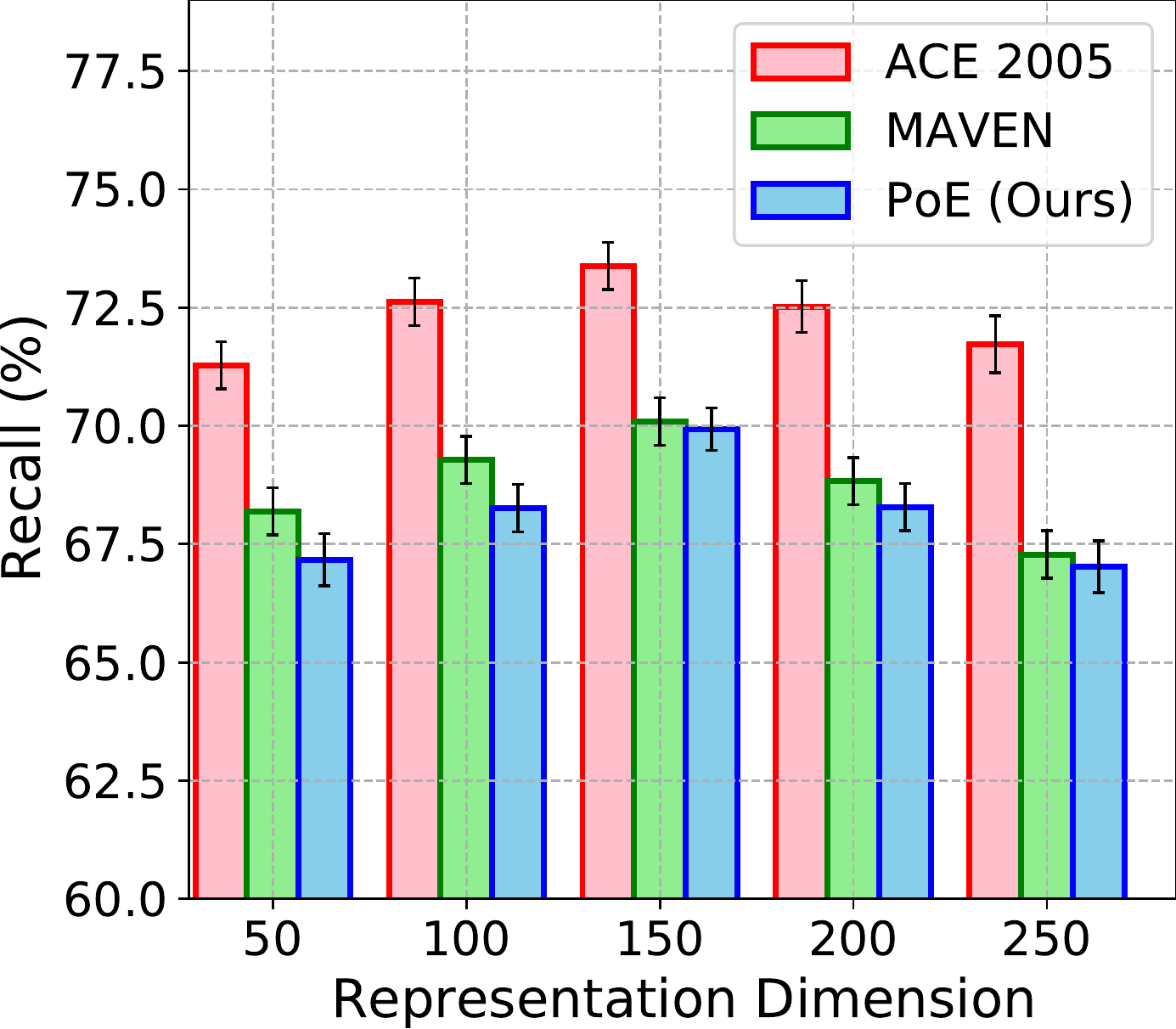}}
    \caption{Parameter study on representation dimension. PoE not only have the problems of long tail and multi-event difficulties existing in ACE and MAVEN, but also have the characteristics of short text, many terms and colloquial language, which increase the difficulty of detection. Therefore, the performance of PoE is lower than that of the other two data sets. In addition, our model performs best on representation dimension of 150.}
    \label{dimension}
\end{figure*}


\begin{figure*}[t]
    \centering
 \includegraphics[width=\linewidth]{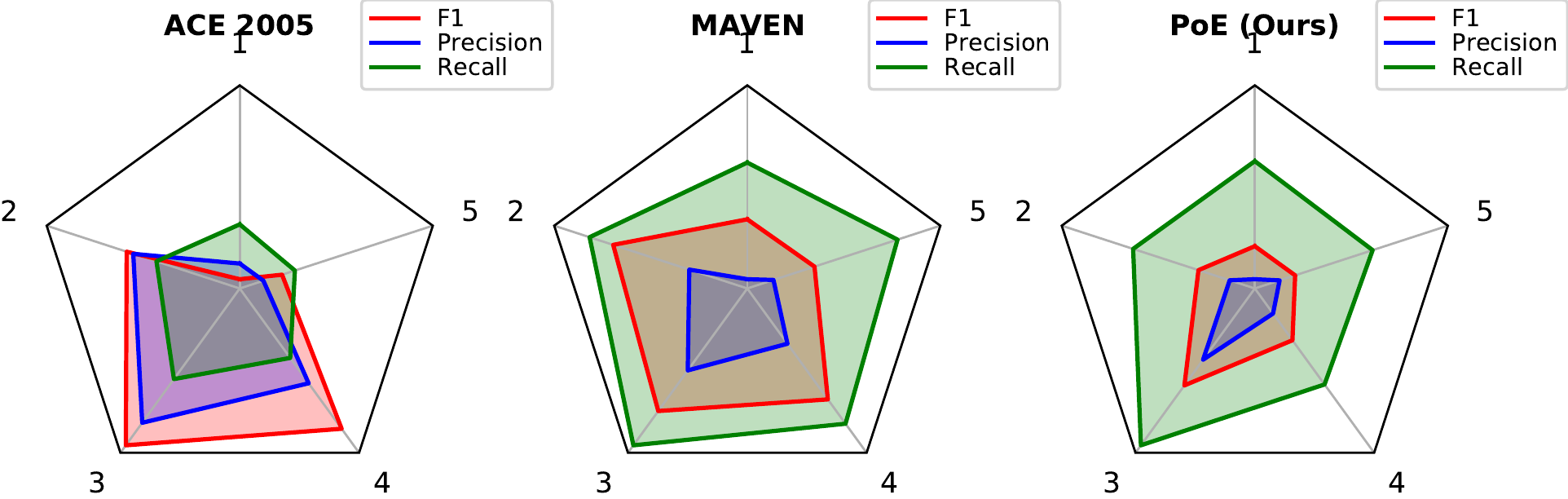}
    \caption{Parameter study on threshold $\rho_{\text {sem}}$. In all datasets, our model performs best on the threshold of 0.3. F1 values vary significantly in our data set and ACE 2005, which demonstrates the validity of semantic channel.}
    \label{threshold}
\end{figure*}



\subsection{Parameter Study}
In order to further evaluate the sensitivity of MC-TED to parameters, we conduct parameter experiments on PoE. 
Fig. \ref{rate}, \ref{layer}, \ref{dimension} and \ref{threshold} show how the effect of label rate, the GCN layer, the dimension of node representation, and the cosine distance of node representation.

\textbf{Label rate.}
As shown in Fig. \ref{rate}, we compare our model with EE-GCN and MLBiNet using two layers. MC-TED is significantly superior to the advanced graph-based methods on most label rate. It shows that our model utilize the node type and relation type is efficient regardless of the number of labeled samples. 
MC-TED performs best when we use the full dataset.


\textbf{Model layer.}
The number of layers of the heterogeneous graph neural network is relatively shallow. As the number of layers deepens, the effect of heterogeneous graph neural network will decrease significantly (semantic confusion phenomenon). Fig. \ref{layer} shows the performance of MC-TED with 1-5 layers GCN. It can be seen that as the number of model layers deepens, the performance of our model will decrease significantly (semantic confusion phenomenon). The 2-layer, MC-TED can learn the node representation with distinguishing degree by capturing the heterogeneous structure and retain rich semantics on the graph, while the node representation learned by the 5-layer of our model has become indistinguishable.
The number of layers of MC-TED is relatively shallow (usually, 1-2 layers are better). 
The MC-TED model performs best with 2 GCN layers.

\textbf{Model dimensions.}
The representation dimension of nodes will directly affect the performance of MC-TED. As shown in Fig. \ref{dimension}, with the increase of node representation dimension, the performance of our model increased slowly at first, then remained unchanged, and finally decreased slowly. This is because MC-TED needs enough dimensions to encode semantic and structural information, but too large dimensions may introduce some redundancy and lead to the over-fitting phenomenon.
MC-TED performs best when the node representation dimension is 150. 
Therefore, we finally sets the node representation dimension as 150.

\textbf{Cosine distance.}
In the semantic-aware channel, we construct the edge between nodes according to cosine distance. We let nodes whose distance is less than the threshold have no edges, and MC-TED updates edge in training. As shown in Fig. \ref{threshold}, with the increase of the threshold, the performance of the MC-TED model increased slowly at first, and then decreased slowly. 
It achieves optimal performance when the threshold value is 0.3.

\begin{table}
\caption{Analysis of type learning mechanism. The bold values are the best result.}
\centering
\renewcommand\arraystretch{1.2}
\resizebox{0.85\linewidth}{!}{
\begin{tabular}{l|ccc}
\toprule
 \textbf{Method} &  F1 (\%) & Precison (\%) & Recall (\%) \\
\midrule
 \textbf{MC-TED}  & \textbf{65.58 $\pm$ 1.24} &	63.71 $\pm$ 1.29 &	\textbf{69.93 $\pm$ 1.61}  \\ 
 \textbf{w/o relation type learning module}   & 63.75 $\pm$ 1.47   &  \textbf{64.92 $\pm$ 1.92} & 62.63 $\pm$ 1.03  \\ 
 \textbf{w/o word type learning module}  & 63.29 $\pm$ 1.58 &   60.65 $\pm$ 1.37 & 66.16 $\pm$ 1.65 \\ 
\bottomrule
\end{tabular}
}
\label{Ablation3}
\end{table}

\begin{figure*}[t]
    \centering
    \includegraphics[width=\linewidth]{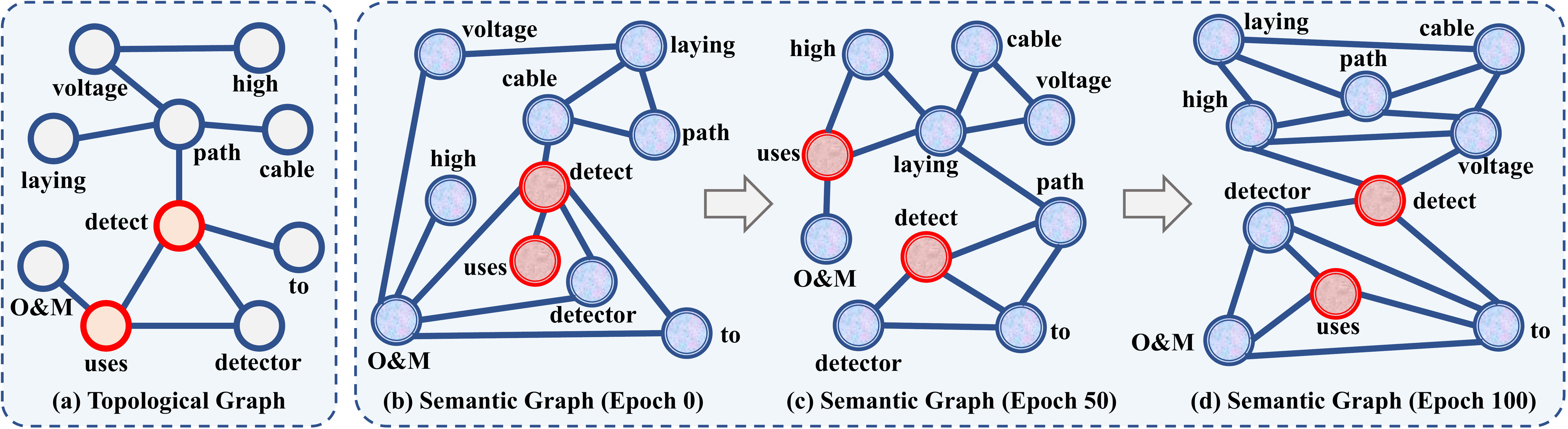}
    \caption{Analysis of multiple channels: topological graph and learning process of the semantic graph. In the semantic channel, the connection of the graph will change with the increase of model training epoches. The words in one event will be connected, and the connection of words in different events will be disconnected.}
    \label{Analysis}
\end{figure*}

\subsection{Analysis of Multiple Channels}
We further provide more detailed analysis on topological channel and semantic channel. Specifically, we (1) investigate the type learning mechanism in topological channel for the type utilized multi-channel GNN by ablation study, (2) visualize the structure change of the semantic-aware graph to study the effect of network reconstruction, and (3) compare the two graphs to seek the relation between the two channels. We have the following three observations.

\textbf{Type learning in topological channel is essential.}
As an important part of reducing the error of wrong analysis of both relation types and word types, the type learning mechanism is proven to be necessary in Table~\ref{Ablation3}. Without the relation type learning or word type learning, the results of event detection have a decline by $1.83\%$ and $2.29\%$ in terms of F1 score. Thus, the type learning mechanism can effectively reduce the cumulative error caused by the analysis tool, which is exacerbated by the professional domain-specific terms in the power systems.

\textbf{Dynamic structure learning in semantic channel is obviously effective.}
The semantic channel reconstructs the semantic-aware graph and learn a new structure based on the current representations in each training step. We visualize the structure change of semantic-aware graph during a training process in Fig.~\ref{Analysis} (b) (c) (d), taking 50 epochs as an interval. The effectiveness of structure learning is obvious that the semantic-relevance words in an event are becoming a cluster during the training process and the corresponding trigger is eventually at the cluster center. The tight connections between the trigger and core arguments makes the trigger identification more accurate, and cluster characteristics helps to classify the event type.

\textbf{Semantic channel and topological channel complement each other.} 
Fig.~\ref{Analysis} (a) (b) show examples of the topological graph and semantic graph. It is clearly noticed that the semanticial channel learns a graph which supplements the non-semantic shortcomings of the topological one in some cases. For instance, the words ``voltage'' and ``cable'' are connected with semantic relevance, while ignored by the topological graph. The above conclusion is also the same for the topological channel with the type utilized multi-channel GNN.



\begin{figure}[t]
    \centering
    \includegraphics[width=\linewidth]{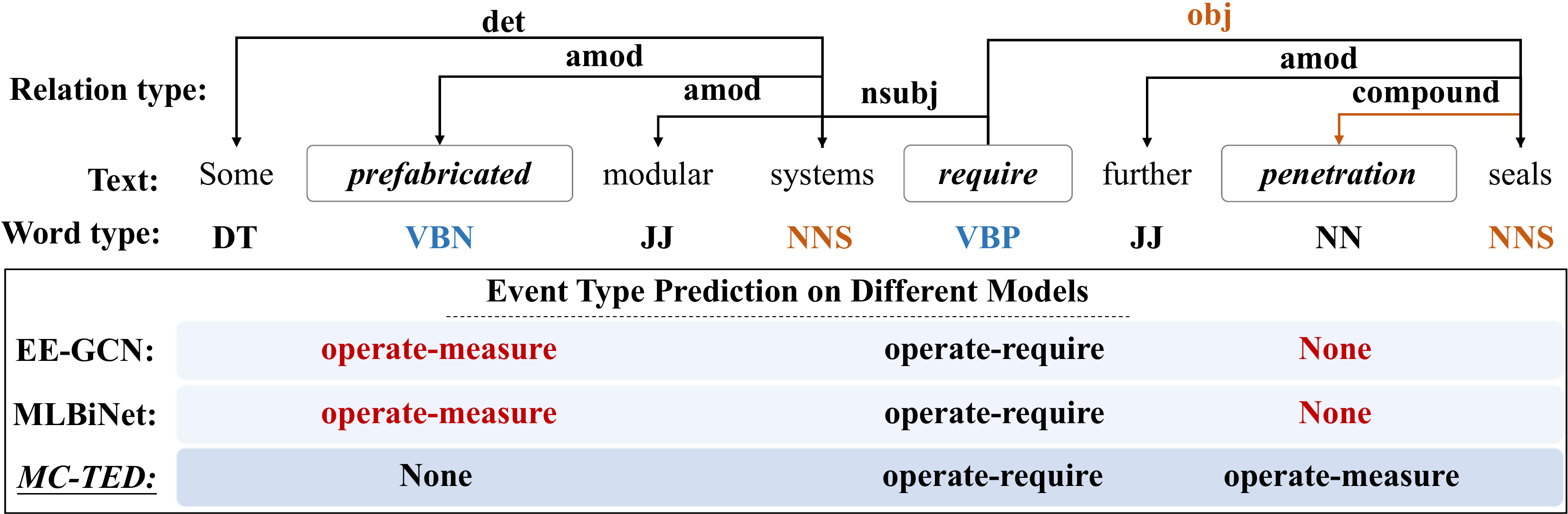}
    \caption{An example of event type prediction. The red color represents wrong prediction. Our method identifies the candidate trigger penetration is a real trigger and the event type is operate-measure, and judges the candidate trigger prefabricated is not a real trigger. It demonstrates that our model can  accurately predict some easily confused events.}
    \label{example2}
\end{figure}

\subsection{Case Study}\label{case_study}

As shown in Fig. \ref{example2}, we give the event detection results of EE-GCN, MLBiNet, and MC-TED. 
Through the CoreNLP tools, an event ``penetration seals'' is an ``compound'' and misrepresented as an element of the clause. Therefore, it is difficult to detect the trigger word ``penetration'' based on the syntactic dependency graph.
MC-TED utilizes the type information and makes the node type and relation type learnable.
MC-TED can accurately classify the events corresponding to the trigger ``penetration'' by introducing node and relation type. It shows that MC-TED can optimize the representation of nodes and edges with a relation-type-aware graph and node-type-aware graph.
Furthermore, the topological graph is constructed based on CoreNLP syntactic analysis tool, there are some node type and relation type are easily to be classified to trigger, such as word ``prefabricated'' with ``VBN'' word type. 
It demonstrates that MC-TED introduces nodes type and edges type learnable mechanism, weakening misleading effect of node type.


\section{Conclusion}
In this paper, we propose MC-TED, a multi-channel GNN which utilizes type information for event detection in power systems. To solve the problems incurred by short texts and professional terminologies in power systems, MC-TED learns representations from topological and semantic channels to capture richer information.
The topological channel generates a relation-type-aware graph and a word-type-aware graph, learning relation-type- and word-type-constrained features. The relation type and word type can both be learned in these two graphs to reduce errors generated by syntactic analysis tools. Furthermore, we build a Chinese event detection dataset in power systems, called PoE, which is derived from electrical power records. 
Extensive experimental results demonstrate that MC-TED achieves state-of-the-art performance on ACE, MAVEN, and PoE datasets.

\begin{acks}
We thank the anonymous reviewers for their insightful comments and suggestions. 
Jianxin Li is the corresponding author.
The authors of this paper were supported by the NSFC through grant 6187202, 62106059 and the Academic Excellence Foundation of Beihang University for PhD Students. 

\end{acks}

\footnotesize
\bibliographystyle{ACM-Reference-Format}
\bibliography{EventDetection}


\begin{thebibliography}{66}


\ifx \showCODEN    \undefined \def \showCODEN     #1{\unskip}     \fi
\ifx \showDOI      \undefined \def \showDOI       #1{#1}\fi
\ifx \showISBNx    \undefined \def \showISBNx     #1{\unskip}     \fi
\ifx \showISBNxiii \undefined \def \showISBNxiii  #1{\unskip}     \fi
\ifx \showISSN     \undefined \def \showISSN      #1{\unskip}     \fi
\ifx \showLCCN     \undefined \def \showLCCN      #1{\unskip}     \fi
\ifx \shownote     \undefined \def \shownote      #1{#1}          \fi
\ifx \showarticletitle \undefined \def \showarticletitle #1{#1}   \fi
\ifx \showURL      \undefined \def \showURL       {\relax}        \fi
\providecommand\bibfield[2]{#2}
\providecommand\bibinfo[2]{#2}
\providecommand\natexlab[1]{#1}
\providecommand\showeprint[2][]{arXiv:#2}

\bibitem[\protect\citeauthoryear{Arrillage, Arnold, and Harker}{Arrillage
  et~al\mbox{.}}{1983}]%
        {arrillage1983computer}
\bibfield{author}{\bibinfo{person}{J Arrillage}, \bibinfo{person}{Charles~P
  Arnold}, {and} \bibinfo{person}{BJ Harker}.} \bibinfo{year}{1983}\natexlab{}.
\newblock \showarticletitle{Computer modelling of electrical power systems}.
\newblock  (\bibinfo{year}{1983}).
\newblock


\bibitem[\protect\citeauthoryear{Cao, Peng, Wu, Dou, Li, and Yu}{Cao
  et~al\mbox{.}}{2021}]%
        {DBLP:conf/www/CaoPWDLY21}
\bibfield{author}{\bibinfo{person}{Yuwei Cao}, \bibinfo{person}{Hao Peng},
  \bibinfo{person}{Jia Wu}, \bibinfo{person}{Yingtong Dou},
  \bibinfo{person}{Jianxin Li}, {and} \bibinfo{person}{Philip~S. Yu}.}
  \bibinfo{year}{2021}\natexlab{}.
\newblock \showarticletitle{Knowledge-Preserving Incremental Social Event
  Detection via Heterogeneous GNNs}. In \bibinfo{booktitle}{\emph{{WWW} '21:
  The Web Conference 2021, Virtual Event / Ljubljana, Slovenia, April 19-23,
  2021}}. \bibinfo{pages}{3383--3395}.
\newblock
\urldef\tempurl%
\url{https://doi.org/10.1145/3442381.3449834}
\showDOI{\tempurl}


\bibitem[\protect\citeauthoryear{Chen, Xu, Liu, Zeng, and Zhao}{Chen
  et~al\mbox{.}}{2015}]%
        {DBLP:conf/acl/ChenXLZ015}
\bibfield{author}{\bibinfo{person}{Yubo Chen}, \bibinfo{person}{Liheng Xu},
  \bibinfo{person}{Kang Liu}, \bibinfo{person}{Daojian Zeng}, {and}
  \bibinfo{person}{Jun Zhao}.} \bibinfo{year}{2015}\natexlab{}.
\newblock \showarticletitle{Event extraction via dynamic multi-pooling
  convolutional neural networks}. In \bibinfo{booktitle}{\emph{Proceedings of
  the 53rd Annual Meeting of the Association for Computational Linguistics and
  the 7th International Joint Conference on Natural Language Processing (Volume
  1: Long Papers)}}. \bibinfo{pages}{167--176}.
\newblock


\bibitem[\protect\citeauthoryear{Cong, Cui, Yu, Liu, Wang, and Wang}{Cong
  et~al\mbox{.}}{2021}]%
        {DBLP:conf/acl/CongCYLWW21}
\bibfield{author}{\bibinfo{person}{Xin Cong}, \bibinfo{person}{Shiyao Cui},
  \bibinfo{person}{Bowen Yu}, \bibinfo{person}{Tingwen Liu},
  \bibinfo{person}{Yubin Wang}, {and} \bibinfo{person}{Bin Wang}.}
  \bibinfo{year}{2021}\natexlab{}.
\newblock \showarticletitle{Few-Shot Event Detection with Prototypical
  Amortized Conditional Random Field}. In \bibinfo{booktitle}{\emph{Findings of
  the Association for Computational Linguistics: {ACL/IJCNLP} 2021, Online
  Event, August 1-6, 2021}} \emph{(\bibinfo{series}{Findings of {ACL}},
  Vol.~\bibinfo{volume}{{ACL/IJCNLP} 2021})}. \bibinfo{pages}{28--40}.
\newblock
\urldef\tempurl%
\url{https://doi.org/10.18653/v1/2021.findings-acl.3}
\showDOI{\tempurl}


\bibitem[\protect\citeauthoryear{Cui, Yu, Liu, Zhang, Wang, and Shi}{Cui
  et~al\mbox{.}}{2020}]%
        {DBLP:conf/emnlp/Cui0L0WS20}
\bibfield{author}{\bibinfo{person}{Shiyao Cui}, \bibinfo{person}{Bowen Yu},
  \bibinfo{person}{Tingwen Liu}, \bibinfo{person}{Zhenyu Zhang},
  \bibinfo{person}{Xuebin Wang}, {and} \bibinfo{person}{Jinqiao Shi}.}
  \bibinfo{year}{2020}\natexlab{}.
\newblock \showarticletitle{Edge-Enhanced Graph Convolution Networks for Event
  Detection with Syntactic Relation}. In \bibinfo{booktitle}{\emph{Findings of
  the Association for Computational Linguistics: {EMNLP} 2020, Online Event,
  16-20 November 2020}} \emph{(\bibinfo{series}{Findings of {ACL}},
  Vol.~\bibinfo{volume}{{EMNLP} 2020})}. \bibinfo{pages}{2329--2339}.
\newblock
\urldef\tempurl%
\url{https://doi.org/10.18653/v1/2020.findings-emnlp.211}
\showDOI{\tempurl}


\bibitem[\protect\citeauthoryear{Devlin, Chang, Lee, and Toutanova}{Devlin
  et~al\mbox{.}}{2019}]%
        {DBLP:conf/naacl/DevlinCLT19}
\bibfield{author}{\bibinfo{person}{Jacob Devlin}, \bibinfo{person}{Ming{-}Wei
  Chang}, \bibinfo{person}{Kenton Lee}, {and} \bibinfo{person}{Kristina
  Toutanova}.} \bibinfo{year}{2019}\natexlab{}.
\newblock \showarticletitle{{BERT:} Pre-training of Deep Bidirectional
  Transformers for Language Understanding}. In
  \bibinfo{booktitle}{\emph{Proceedings of the 2019 Conference of the North
  American Chapter of the Association for Computational Linguistics: Human
  Language Technologies, {NAACL-HLT} 2019, Minneapolis, MN, USA, June 2-7,
  2019, Volume 1 (Long and Short Papers)}}. \bibinfo{pages}{4171--4186}.
\newblock
\urldef\tempurl%
\url{https://doi.org/10.18653/v1/n19-1423}
\showDOI{\tempurl}


\bibitem[\protect\citeauthoryear{Doddington, Mitchell, Przybocki, Ramshaw,
  Strassel, and Weischedel}{Doddington et~al\mbox{.}}{2004}]%
        {DBLP:conf/lrec/DoddingtonMPRSW04}
\bibfield{author}{\bibinfo{person}{George~R. Doddington},
  \bibinfo{person}{Alexis Mitchell}, \bibinfo{person}{Mark~A. Przybocki},
  \bibinfo{person}{Lance~A. Ramshaw}, \bibinfo{person}{Stephanie~M. Strassel},
  {and} \bibinfo{person}{Ralph~M. Weischedel}.}
  \bibinfo{year}{2004}\natexlab{}.
\newblock \showarticletitle{The Automatic Content Extraction {(ACE)} Program -
  Tasks, Data, and Evaluation}. In \bibinfo{booktitle}{\emph{Proceedings of the
  Fourth International Conference on Language Resources and Evaluation, {LREC}
  2004, May 26-28, 2004, Lisbon, Portugal}}.
\newblock
\urldef\tempurl%
\url{http://www.lrec-conf.org/proceedings/lrec2004/summaries/5.htm}
\showURL{%
\tempurl}


\bibitem[\protect\citeauthoryear{Dong, Chawla, and Swami}{Dong
  et~al\mbox{.}}{2017}]%
        {DBLP:conf/kdd/DongCS17}
\bibfield{author}{\bibinfo{person}{Yuxiao Dong}, \bibinfo{person}{Nitesh~V.
  Chawla}, {and} \bibinfo{person}{Ananthram Swami}.}
  \bibinfo{year}{2017}\natexlab{}.
\newblock \showarticletitle{metapath2vec: Scalable Representation Learning for
  Heterogeneous Networks}. In \bibinfo{booktitle}{\emph{Proceedings of the 23rd
  {ACM} {SIGKDD} International Conference on Knowledge Discovery and Data
  Mining, Halifax, NS, Canada, August 13 - 17, 2017}}.
  \bibinfo{pages}{135--144}.
\newblock
\urldef\tempurl%
\url{https://doi.org/10.1145/3097983.3098036}
\showDOI{\tempurl}


\bibitem[\protect\citeauthoryear{Dutta, Ma, Saha, Lu, Tetreault, and
  Jaimes}{Dutta et~al\mbox{.}}{2021}]%
        {DBLP:journals/corr/abs-2104-15104}
\bibfield{author}{\bibinfo{person}{Sanghamitra Dutta}, \bibinfo{person}{Liang
  Ma}, \bibinfo{person}{Tanay~Kumar Saha}, \bibinfo{person}{Di Lu},
  \bibinfo{person}{Joel~R. Tetreault}, {and} \bibinfo{person}{Alex Jaimes}.}
  \bibinfo{year}{2021}\natexlab{}.
\newblock \showarticletitle{{GTN-ED:} Event Detection Using Graph Transformer
  Networks}.
\newblock \bibinfo{journal}{\emph{CoRR}}  \bibinfo{volume}{abs/2104.15104}
  (\bibinfo{year}{2021}).
\newblock
\showeprint[arXiv]{2104.15104}
\urldef\tempurl%
\url{https://arxiv.org/abs/2104.15104}
\showURL{%
\tempurl}


\bibitem[\protect\citeauthoryear{El-Hawary}{El-Hawary}{1995}]%
        {el1995electrical}
\bibfield{author}{\bibinfo{person}{Mohamed~E El-Hawary}.}
  \bibinfo{year}{1995}\natexlab{}.
\newblock \bibinfo{booktitle}{\emph{Electrical power systems: design and
  analysis}}. Vol.~\bibinfo{volume}{2}.
\newblock


\bibitem[\protect\citeauthoryear{Fan, Zhu, Han, Shi, Hu, Ma, and Li}{Fan
  et~al\mbox{.}}{2019}]%
        {DBLP:conf/kdd/FanZHSHML19}
\bibfield{author}{\bibinfo{person}{Shaohua Fan}, \bibinfo{person}{Junxiong
  Zhu}, \bibinfo{person}{Xiaotian Han}, \bibinfo{person}{Chuan Shi},
  \bibinfo{person}{Linmei Hu}, \bibinfo{person}{Biyu Ma}, {and}
  \bibinfo{person}{Yongliang Li}.} \bibinfo{year}{2019}\natexlab{}.
\newblock \showarticletitle{Metapath-guided Heterogeneous Graph Neural Network
  for Intent Recommendation}. In \bibinfo{booktitle}{\emph{Proceedings of the
  25th {ACM} {SIGKDD} International Conference on Knowledge Discovery {\&} Data
  Mining, {KDD} 2019, Anchorage, AK, USA, August 4-8, 2019}}.
  \bibinfo{pages}{2478--2486}.
\newblock
\urldef\tempurl%
\url{https://doi.org/10.1145/3292500.3330673}
\showDOI{\tempurl}


\bibitem[\protect\citeauthoryear{Feng, Huang, Tang, Ji, Qin, and Liu}{Feng
  et~al\mbox{.}}{2016}]%
        {DBLP:conf/acl/FengHTJQL16}
\bibfield{author}{\bibinfo{person}{Xiaocheng Feng}, \bibinfo{person}{Lifu
  Huang}, \bibinfo{person}{Duyu Tang}, \bibinfo{person}{Heng Ji},
  \bibinfo{person}{Bing Qin}, {and} \bibinfo{person}{Ting Liu}.}
  \bibinfo{year}{2016}\natexlab{}.
\newblock \showarticletitle{A Language-Independent Neural Network for Event
  Detection}. In \bibinfo{booktitle}{\emph{Proceedings of the 54th Annual
  Meeting of the Association for Computational Linguistics, {ACL} 2016, August
  7-12, 2016, Berlin, Germany, Volume 2: Short Papers}}.
\newblock
\urldef\tempurl%
\url{https://doi.org/10.18653/v1/p16-2011}
\showDOI{\tempurl}


\bibitem[\protect\citeauthoryear{Hu, Yang, Shi, Ji, and Li}{Hu
  et~al\mbox{.}}{2019}]%
        {DBLP:conf/emnlp/HuYSJL19}
\bibfield{author}{\bibinfo{person}{Linmei Hu}, \bibinfo{person}{Tianchi Yang},
  \bibinfo{person}{Chuan Shi}, \bibinfo{person}{Houye Ji}, {and}
  \bibinfo{person}{Xiaoli Li}.} \bibinfo{year}{2019}\natexlab{}.
\newblock \showarticletitle{Heterogeneous Graph Attention Networks for
  Semi-supervised Short Text Classification}. In
  \bibinfo{booktitle}{\emph{Proceedings of the 2019 Conference on Empirical
  Methods in Natural Language Processing and the 9th International Joint
  Conference on Natural Language Processing, {EMNLP-IJCNLP} 2019, Hong Kong,
  China, November 3-7, 2019}}. \bibinfo{pages}{4820--4829}.
\newblock
\urldef\tempurl%
\url{https://doi.org/10.18653/v1/D19-1488}
\showDOI{\tempurl}


\bibitem[\protect\citeauthoryear{Hu, Dong, Wang, and Sun}{Hu
  et~al\mbox{.}}{2020}]%
        {DBLP:conf/www/HuDWS20}
\bibfield{author}{\bibinfo{person}{Ziniu Hu}, \bibinfo{person}{Yuxiao Dong},
  \bibinfo{person}{Kuansan Wang}, {and} \bibinfo{person}{Yizhou Sun}.}
  \bibinfo{year}{2020}\natexlab{}.
\newblock \showarticletitle{Heterogeneous Graph Transformer}. In
  \bibinfo{booktitle}{\emph{{WWW} '20: The Web Conference 2020, Taipei, Taiwan,
  April 20-24, 2020}}. \bibinfo{pages}{2704--2710}.
\newblock
\urldef\tempurl%
\url{https://doi.org/10.1145/3366423.3380027}
\showDOI{\tempurl}


\bibitem[\protect\citeauthoryear{Huang, Ji, Cho, Dagan, Riedel, and Voss}{Huang
  et~al\mbox{.}}{2018}]%
        {DBLP:conf/acl/DaganJVHCR18}
\bibfield{author}{\bibinfo{person}{Lifu Huang}, \bibinfo{person}{Heng Ji},
  \bibinfo{person}{Kyunghyun Cho}, \bibinfo{person}{Ido Dagan},
  \bibinfo{person}{Sebastian Riedel}, {and} \bibinfo{person}{Clare~R. Voss}.}
  \bibinfo{year}{2018}\natexlab{}.
\newblock \showarticletitle{Zero-Shot Transfer Learning for Event Extraction}.
  In \bibinfo{booktitle}{\emph{Proceedings of the 56th Annual Meeting of the
  Association for Computational Linguistics, {ACL} 2018, Melbourne, Australia,
  July 15-20, 2018, Volume 1: Long Papers}}. \bibinfo{pages}{2160--2170}.
\newblock
\urldef\tempurl%
\url{https://doi.org/10.18653/v1/P18-1201}
\showDOI{\tempurl}


\bibitem[\protect\citeauthoryear{Huang, Li, Ye, and Ng}{Huang
  et~al\mbox{.}}{2020}]%
        {DBLP:conf/ijcai/HuangLYN20}
\bibfield{author}{\bibinfo{person}{Zhichao Huang}, \bibinfo{person}{Xutao Li},
  \bibinfo{person}{Yunming Ye}, {and} \bibinfo{person}{Michael~K. Ng}.}
  \bibinfo{year}{2020}\natexlab{}.
\newblock \showarticletitle{{MR-GCN:} Multi-Relational Graph Convolutional
  Networks based on Generalized Tensor Product}. In
  \bibinfo{booktitle}{\emph{Proceedings of the Twenty-Ninth International Joint
  Conference on Artificial Intelligence, {IJCAI} 2020}}.
  \bibinfo{pages}{1258--1264}.
\newblock
\urldef\tempurl%
\url{https://doi.org/10.24963/ijcai.2020/175}
\showDOI{\tempurl}


\bibitem[\protect\citeauthoryear{Hussain, Alammari, Iqbal, and Shikfa}{Hussain
  et~al\mbox{.}}{2020}]%
        {DBLP:conf/iciot3/HussainAIS20}
\bibfield{author}{\bibinfo{person}{Shahbaz Hussain}, \bibinfo{person}{Rashid
  Alammari}, \bibinfo{person}{Atif Iqbal}, {and} \bibinfo{person}{Abdullatif
  Shikfa}.} \bibinfo{year}{2020}\natexlab{}.
\newblock \showarticletitle{Application of artificial intelligence in
  electrical power systems}. In \bibinfo{booktitle}{\emph{{IEEE} International
  Conference on Informatics, IoT, and Enabling Technologies, ICIoT 2020, Doha,
  Qatar, February 2-5, 2020}}. \bibinfo{pages}{13--17}.
\newblock
\urldef\tempurl%
\url{https://doi.org/10.1109/ICIoT48696.2020.9089447}
\showDOI{\tempurl}


\bibitem[\protect\citeauthoryear{Jiang}{Jiang}{2021}]%
        {DBLP:journals/tsg/Jiang21}
\bibfield{author}{\bibinfo{person}{Yazhou Jiang}.}
  \bibinfo{year}{2021}\natexlab{}.
\newblock \showarticletitle{Data-Driven Probabilistic Fault Location of
  Electric Power Distribution Systems Incorporating Data Uncertainties}.
\newblock \bibinfo{journal}{\emph{{IEEE} Trans. Smart Grid}}
  \bibinfo{volume}{12}, \bibinfo{number}{5} (\bibinfo{year}{2021}),
  \bibinfo{pages}{4522--4534}.
\newblock
\urldef\tempurl%
\url{https://doi.org/10.1109/TSG.2021.3070550}
\showDOI{\tempurl}


\bibitem[\protect\citeauthoryear{Khan and Blumenstock}{Khan and
  Blumenstock}{2019}]%
        {DBLP:conf/aaai/KhanB19}
\bibfield{author}{\bibinfo{person}{Muhammad~Raza Khan} {and}
  \bibinfo{person}{Joshua~E. Blumenstock}.} \bibinfo{year}{2019}\natexlab{}.
\newblock \showarticletitle{Multi-GCN: Graph Convolutional Networks for
  Multi-View Networks, with Applications to Global Poverty}. In
  \bibinfo{booktitle}{\emph{The Thirty-Third {AAAI} Conference on Artificial
  Intelligence, {AAAI} 2019, The Thirty-First Innovative Applications of
  Artificial Intelligence Conference, {IAAI} 2019, The Ninth {AAAI} Symposium
  on Educational Advances in Artificial Intelligence, {EAAI} 2019, Honolulu,
  Hawaii, USA, January 27 - February 1, 2019}}. \bibinfo{pages}{606--613}.
\newblock
\urldef\tempurl%
\url{https://doi.org/10.1609/aaai.v33i01.3301606}
\showDOI{\tempurl}


\bibitem[\protect\citeauthoryear{Khodayar, Liu, Wang, and Khodayar}{Khodayar
  et~al\mbox{.}}{2020}]%
        {khodayar2020deep}
\bibfield{author}{\bibinfo{person}{Mahdi Khodayar}, \bibinfo{person}{Guangyi
  Liu}, \bibinfo{person}{Jianhui Wang}, {and} \bibinfo{person}{Mohammad~E
  Khodayar}.} \bibinfo{year}{2020}\natexlab{}.
\newblock \showarticletitle{Deep learning in power systems research: A review}.
\newblock \bibinfo{journal}{\emph{CSEE Journal of Power and Energy Systems}}
  (\bibinfo{year}{2020}).
\newblock


\bibitem[\protect\citeauthoryear{Khodayar, Mohammadi, Khodayar, Wang, and
  Liu}{Khodayar et~al\mbox{.}}{2019}]%
        {khodayar2019convolutional}
\bibfield{author}{\bibinfo{person}{Mahdi Khodayar}, \bibinfo{person}{Saeed
  Mohammadi}, \bibinfo{person}{Mohammad~E Khodayar}, \bibinfo{person}{Jianhui
  Wang}, {and} \bibinfo{person}{Guangyi Liu}.} \bibinfo{year}{2019}\natexlab{}.
\newblock \showarticletitle{Convolutional graph autoencoder: A generative deep
  neural network for probabilistic spatio-temporal solar irradiance
  forecasting}.
\newblock \bibinfo{journal}{\emph{IEEE Transactions on Sustainable Energy}}
  \bibinfo{volume}{11}, \bibinfo{number}{2} (\bibinfo{year}{2019}),
  \bibinfo{pages}{571--583}.
\newblock


\bibitem[\protect\citeauthoryear{Khodayar and Wang}{Khodayar and Wang}{2018}]%
        {khodayar2018spatio}
\bibfield{author}{\bibinfo{person}{Mahdi Khodayar} {and}
  \bibinfo{person}{Jianhui Wang}.} \bibinfo{year}{2018}\natexlab{}.
\newblock \showarticletitle{Spatio-temporal graph deep neural network for
  short-term wind speed forecasting}.
\newblock \bibinfo{journal}{\emph{IEEE Transactions on Sustainable Energy}}
  \bibinfo{volume}{10}, \bibinfo{number}{2} (\bibinfo{year}{2018}),
  \bibinfo{pages}{670--681}.
\newblock


\bibitem[\protect\citeauthoryear{Kingma and Ba}{Kingma and Ba}{2015}]%
        {kingma2015adam}
\bibfield{author}{\bibinfo{person}{Diederik~P. Kingma} {and}
  \bibinfo{person}{Jimmy Ba}.} \bibinfo{year}{2015}\natexlab{}.
\newblock \showarticletitle{Adam: {A} Method for Stochastic Optimization}. In
  \bibinfo{booktitle}{\emph{3rd International Conference on Learning
  Representations, {ICLR} 2015, San Diego, CA, USA, May 7-9, 2015, Conference
  Track Proceedings}}.
\newblock
\urldef\tempurl%
\url{http://arxiv.org/abs/1412.6980}
\showURL{%
\tempurl}


\bibitem[\protect\citeauthoryear{Kipf and Welling}{Kipf and Welling}{2017}]%
        {DBLP:conf/iclr/KipfW17}
\bibfield{author}{\bibinfo{person}{Thomas~N. Kipf} {and} \bibinfo{person}{Max
  Welling}.} \bibinfo{year}{2017}\natexlab{}.
\newblock \showarticletitle{Semi-Supervised Classification with Graph
  Convolutional Networks}. In \bibinfo{booktitle}{\emph{5th International
  Conference on Learning Representations, {ICLR} 2017, Toulon, France, April
  24-26, 2017, Conference Track Proceedings}}.
\newblock
\urldef\tempurl%
\url{https://openreview.net/forum?id=SJU4ayYgl}
\showURL{%
\tempurl}


\bibitem[\protect\citeauthoryear{Lai, Nguyen, Nguyen, and Dernoncourt}{Lai
  et~al\mbox{.}}{2021}]%
        {DBLP:conf/sigir/LaiNND21}
\bibfield{author}{\bibinfo{person}{Viet~Dac Lai}, \bibinfo{person}{Minh~Van
  Nguyen}, \bibinfo{person}{Thien~Huu Nguyen}, {and} \bibinfo{person}{Franck
  Dernoncourt}.} \bibinfo{year}{2021}\natexlab{}.
\newblock \showarticletitle{Graph Learning Regularization and Transfer Learning
  for Few-Shot Event Detection}. In \bibinfo{booktitle}{\emph{{SIGIR} '21: The
  44th International {ACM} {SIGIR} Conference on Research and Development in
  Information Retrieval, Virtual Event, Canada, July 11-15, 2021}}.
  \bibinfo{pages}{2172--2176}.
\newblock
\urldef\tempurl%
\url{https://doi.org/10.1145/3404835.3463054}
\showDOI{\tempurl}


\bibitem[\protect\citeauthoryear{Lai, Nguyen, and Nguyen}{Lai
  et~al\mbox{.}}{2020a}]%
        {DBLP:conf/emnlp/LaiNN20}
\bibfield{author}{\bibinfo{person}{Viet~Dac Lai}, \bibinfo{person}{Tuan~Ngo
  Nguyen}, {and} \bibinfo{person}{Thien~Huu Nguyen}.}
  \bibinfo{year}{2020}\natexlab{a}.
\newblock \showarticletitle{Event Detection: Gate Diversity and Syntactic
  Importance Scores for Graph Convolution Neural Networks}. In
  \bibinfo{booktitle}{\emph{Proceedings of the 2020 Conference on Empirical
  Methods in Natural Language Processing, {EMNLP} 2020, Online, November 16-20,
  2020}}. \bibinfo{pages}{5405--5411}.
\newblock
\urldef\tempurl%
\url{https://doi.org/10.18653/v1/2020.emnlp-main.435}
\showDOI{\tempurl}


\bibitem[\protect\citeauthoryear{Lai, Nguyen, and Nguyen}{Lai
  et~al\mbox{.}}{2020b}]%
        {DBLP:journals/corr/abs-2010-14123}
\bibfield{author}{\bibinfo{person}{Viet~Dac Lai}, \bibinfo{person}{Tuan~Ngo
  Nguyen}, {and} \bibinfo{person}{Thien~Huu Nguyen}.}
  \bibinfo{year}{2020}\natexlab{b}.
\newblock \showarticletitle{Event Detection: Gate Diversity and Syntactic
  Importance Scoresfor Graph Convolution Neural Networks}.
\newblock \bibinfo{journal}{\emph{CoRR}}  \bibinfo{volume}{abs/2010.14123}
  (\bibinfo{year}{2020}).
\newblock
\showeprint[arXiv]{2010.14123}
\urldef\tempurl%
\url{https://arxiv.org/abs/2010.14123}
\showURL{%
\tempurl}


\bibitem[\protect\citeauthoryear{Li, Zhu, Zhou, Sun, Jiang, Zhang, and Hu}{Li
  et~al\mbox{.}}{2022d}]%
        {li2022aiqoser}
\bibfield{author}{\bibinfo{person}{Jianxin Li}, \bibinfo{person}{Tianchen Zhu},
  \bibinfo{person}{Haoyi Zhou}, \bibinfo{person}{Qingyun Sun},
  \bibinfo{person}{Chunyang Jiang}, \bibinfo{person}{Shuai Zhang}, {and}
  \bibinfo{person}{Chunming Hu}.} \bibinfo{year}{2022}\natexlab{d}.
\newblock \showarticletitle{AIQoSer: Building the efficient Inference-QoS for
  AI Services}. In \bibinfo{booktitle}{\emph{2022 IEEE/ACM 30th International
  Symposium on Quality of Service (IWQoS)}}. IEEE, \bibinfo{pages}{1--10}.
\newblock


\bibitem[\protect\citeauthoryear{Li, Li, Sheng, Cui, Wu, Hei, Peng, Guo, Wang,
  Beheshti, and Yu}{Li et~al\mbox{.}}{2022a}]%
        {9927311}
\bibfield{author}{\bibinfo{person}{Qian Li}, \bibinfo{person}{Jianxin Li},
  \bibinfo{person}{Jiawei Sheng}, \bibinfo{person}{Shiyao Cui},
  \bibinfo{person}{Jia Wu}, \bibinfo{person}{Yiming Hei}, \bibinfo{person}{Hao
  Peng}, \bibinfo{person}{Shu Guo}, \bibinfo{person}{Lihong Wang},
  \bibinfo{person}{Amin Beheshti}, {and} \bibinfo{person}{Philip~S. Yu}.}
  \bibinfo{year}{2022}\natexlab{a}.
\newblock \showarticletitle{A Survey on Deep Learning Event Extraction:
  Approaches and Applications}.
\newblock \bibinfo{journal}{\emph{IEEE Transactions on Neural Networks and
  Learning Systems}} (\bibinfo{year}{2022}), \bibinfo{pages}{1--21}.
\newblock
\urldef\tempurl%
\url{https://doi.org/10.1109/TNNLS.2022.3213168}
\showDOI{\tempurl}


\bibitem[\protect\citeauthoryear{Li, Peng, Li, Wu, Ning, Wang, Yu, and Wang}{Li
  et~al\mbox{.}}{2022b}]%
        {DBLP:journals/taslp/LiPLWNWYW22}
\bibfield{author}{\bibinfo{person}{Qian Li}, \bibinfo{person}{Hao Peng},
  \bibinfo{person}{Jianxin Li}, \bibinfo{person}{Jia Wu},
  \bibinfo{person}{Yuanxing Ning}, \bibinfo{person}{Lihong Wang},
  \bibinfo{person}{Philip~S. Yu}, {and} \bibinfo{person}{Zheng Wang}.}
  \bibinfo{year}{2022}\natexlab{b}.
\newblock \showarticletitle{Reinforcement Learning-Based Dialogue Guided Event
  Extraction to Exploit Argument Relations}.
\newblock \bibinfo{journal}{\emph{{IEEE} {ACM} Trans. Audio Speech Lang.
  Process.}}  \bibinfo{volume}{30} (\bibinfo{year}{2022}),
  \bibinfo{pages}{520--533}.
\newblock
\urldef\tempurl%
\url{https://doi.org/10.1109/TASLP.2021.3138670}
\showDOI{\tempurl}


\bibitem[\protect\citeauthoryear{Li, Peng, Li, Xia, Yang, Sun, Yu, and He}{Li
  et~al\mbox{.}}{2022c}]%
        {10.1145/3495162}
\bibfield{author}{\bibinfo{person}{Qian Li}, \bibinfo{person}{Hao Peng},
  \bibinfo{person}{Jianxin Li}, \bibinfo{person}{Congying Xia},
  \bibinfo{person}{Renyu Yang}, \bibinfo{person}{Lichao Sun},
  \bibinfo{person}{Philip~S. Yu}, {and} \bibinfo{person}{Lifang He}.}
  \bibinfo{year}{2022}\natexlab{c}.
\newblock \showarticletitle{A Survey on Text Classification: From Traditional
  to Deep Learning}.
\newblock \bibinfo{journal}{\emph{ACM Trans. Intell. Syst. Technol.}}
  \bibinfo{volume}{13}, \bibinfo{number}{2}, Article \bibinfo{articleno}{31}
  (\bibinfo{date}{apr} \bibinfo{year}{2022}), \bibinfo{numpages}{41}~pages.
\newblock
\showISSN{2157-6904}
\urldef\tempurl%
\url{https://doi.org/10.1145/3495162}
\showDOI{\tempurl}


\bibitem[\protect\citeauthoryear{Liao, Zheng, and Cao}{Liao
  et~al\mbox{.}}{2021}]%
        {DBLP:journals/bigdatama/LiaoZC21}
\bibfield{author}{\bibinfo{person}{Xueting Liao}, \bibinfo{person}{Danyang
  Zheng}, {and} \bibinfo{person}{Xiaojun Cao}.}
  \bibinfo{year}{2021}\natexlab{}.
\newblock \showarticletitle{Coronavirus pandemic analysis through tripartite
  graph clustering in online social networks}.
\newblock \bibinfo{journal}{\emph{Big Data Min. Anal.}} \bibinfo{volume}{4},
  \bibinfo{number}{4} (\bibinfo{year}{2021}), \bibinfo{pages}{242--251}.
\newblock
\urldef\tempurl%
\url{https://doi.org/10.26599/BDMA.2021.9020010}
\showDOI{\tempurl}


\bibitem[\protect\citeauthoryear{Lin, Lu, Han, and Sun}{Lin
  et~al\mbox{.}}{2018}]%
        {DBLP:conf/acl/SunHLL18a}
\bibfield{author}{\bibinfo{person}{Hongyu Lin}, \bibinfo{person}{Yaojie Lu},
  \bibinfo{person}{Xianpei Han}, {and} \bibinfo{person}{Le Sun}.}
  \bibinfo{year}{2018}\natexlab{}.
\newblock \showarticletitle{Nugget Proposal Networks for Chinese Event
  Detection}. In \bibinfo{booktitle}{\emph{Proceedings of the 56th Annual
  Meeting of the Association for Computational Linguistics, {ACL} 2018,
  Melbourne, Australia, July 15-20, 2018, Volume 1: Long Papers}}.
  \bibinfo{pages}{1565--1574}.
\newblock
\urldef\tempurl%
\url{https://doi.org/10.18653/v1/P18-1145}
\showDOI{\tempurl}


\bibitem[\protect\citeauthoryear{Liu, Chen, Liu, and Zhao}{Liu
  et~al\mbox{.}}{2018}]%
        {DBLP:conf/aaai/Liu00018}
\bibfield{author}{\bibinfo{person}{Jian Liu}, \bibinfo{person}{Yubo Chen},
  \bibinfo{person}{Kang Liu}, {and} \bibinfo{person}{Jun Zhao}.}
  \bibinfo{year}{2018}\natexlab{}.
\newblock \showarticletitle{Event Detection via Gated Multilingual Attention
  Mechanism}. In \bibinfo{booktitle}{\emph{Proceedings of the Thirty-Second
  {AAAI} Conference on Artificial Intelligence, (AAAI-18), the 30th innovative
  Applications of Artificial Intelligence (IAAI-18), and the 8th {AAAI}
  Symposium on Educational Advances in Artificial Intelligence (EAAI-18), New
  Orleans, Louisiana, USA, February 2-7, 2018}}. \bibinfo{pages}{4865--4872}.
\newblock
\urldef\tempurl%
\url{https://www.aaai.org/ocs/index.php/AAAI/AAAI18/paper/view/16371}
\showURL{%
\tempurl}


\bibitem[\protect\citeauthoryear{Liu, Zhao, Lin, Liu, Ding, Yang, and Yi}{Liu
  et~al\mbox{.}}{2020b}]%
        {liu2019data}
\bibfield{author}{\bibinfo{person}{Shengyuan Liu}, \bibinfo{person}{Yuxuan
  Zhao}, \bibinfo{person}{Zhenzhi Lin}, \bibinfo{person}{Yilu Liu},
  \bibinfo{person}{Yi Ding}, \bibinfo{person}{Li Yang}, {and}
  \bibinfo{person}{Shimin Yi}.} \bibinfo{year}{2020}\natexlab{b}.
\newblock \showarticletitle{Data-Driven Event Detection of Power Systems Based
  on Unequal-Interval Reduction of {PMU} Data and Local Outlier Factor}.
\newblock \bibinfo{journal}{\emph{{IEEE} Trans. Smart Grid}}
  \bibinfo{volume}{11}, \bibinfo{number}{2} (\bibinfo{year}{2020}),
  \bibinfo{pages}{1630--1643}.
\newblock
\urldef\tempurl%
\url{https://doi.org/10.1109/TSG.2019.2941565}
\showDOI{\tempurl}


\bibitem[\protect\citeauthoryear{Liu, Peng, Li, Song, and Li}{Liu
  et~al\mbox{.}}{2020a}]%
        {DBLP:journals/fcsc/LiuPLSL20}
\bibfield{author}{\bibinfo{person}{Yaopeng Liu}, \bibinfo{person}{Hao Peng},
  \bibinfo{person}{Jianxin Li}, \bibinfo{person}{Yangqiu Song}, {and}
  \bibinfo{person}{Xiong Li}.} \bibinfo{year}{2020}\natexlab{a}.
\newblock \showarticletitle{Event detection and evolution in multi-lingual
  social streams}.
\newblock \bibinfo{journal}{\emph{Frontiers Comput. Sci.}}
  \bibinfo{volume}{14}, \bibinfo{number}{5} (\bibinfo{year}{2020}),
  \bibinfo{pages}{145612}.
\newblock
\urldef\tempurl%
\url{https://doi.org/10.1007/s11704-019-8201-6}
\showDOI{\tempurl}


\bibitem[\protect\citeauthoryear{Lou, Liao, Deng, Zhang, and Chen}{Lou
  et~al\mbox{.}}{2021}]%
        {DBLP:conf/acl/LouLDZC20}
\bibfield{author}{\bibinfo{person}{Dongfang Lou}, \bibinfo{person}{Zhilin
  Liao}, \bibinfo{person}{Shumin Deng}, \bibinfo{person}{Ningyu Zhang}, {and}
  \bibinfo{person}{Huajun Chen}.} \bibinfo{year}{2021}\natexlab{}.
\newblock \showarticletitle{MLBiNet: {A} Cross-Sentence Collective Event
  Detection Network}. In \bibinfo{booktitle}{\emph{Proceedings of the 59th
  Annual Meeting of the Association for Computational Linguistics and the 11th
  International Joint Conference on Natural Language Processing, {ACL/IJCNLP}
  2021, (Volume 1: Long Papers), Virtual Event, August 1-6, 2021}}.
  \bibinfo{pages}{4829--4839}.
\newblock
\urldef\tempurl%
\url{https://doi.org/10.18653/v1/2021.acl-long.373}
\showDOI{\tempurl}


\bibitem[\protect\citeauthoryear{Ma, Wang, Aggarwal, Yin, and Tang}{Ma
  et~al\mbox{.}}{2019}]%
        {DBLP:conf/sdm/MaWAYT19}
\bibfield{author}{\bibinfo{person}{Yao Ma}, \bibinfo{person}{Suhang Wang},
  \bibinfo{person}{Charu~C. Aggarwal}, \bibinfo{person}{Dawei Yin}, {and}
  \bibinfo{person}{Jiliang Tang}.} \bibinfo{year}{2019}\natexlab{}.
\newblock \showarticletitle{Multi-dimensional Graph Convolutional Networks}. In
  \bibinfo{booktitle}{\emph{Proceedings of the 2019 {SIAM} International
  Conference on Data Mining, {SDM} 2019, Calgary, Alberta, Canada, May 2-4,
  2019}}. \bibinfo{pages}{657--665}.
\newblock
\urldef\tempurl%
\url{https://doi.org/10.1137/1.9781611975673.74}
\showDOI{\tempurl}


\bibitem[\protect\citeauthoryear{Mao, Li, Peng, Li, He, Guo, He, and Wang}{Mao
  et~al\mbox{.}}{2021}]%
        {DBLP:journals/fgcs/MaoLPLHGHW21}
\bibfield{author}{\bibinfo{person}{Qianren Mao}, \bibinfo{person}{Xi Li},
  \bibinfo{person}{Hao Peng}, \bibinfo{person}{Jianxin Li},
  \bibinfo{person}{Dongxiao He}, \bibinfo{person}{Shu Guo},
  \bibinfo{person}{Min He}, {and} \bibinfo{person}{Lihong Wang}.}
  \bibinfo{year}{2021}\natexlab{}.
\newblock \showarticletitle{Event prediction based on evolutionary event
  ontology knowledge}.
\newblock \bibinfo{journal}{\emph{Future Gener. Comput. Syst.}}
  \bibinfo{volume}{115} (\bibinfo{year}{2021}), \bibinfo{pages}{76--89}.
\newblock
\urldef\tempurl%
\url{https://doi.org/10.1016/j.future.2020.07.041}
\showDOI{\tempurl}


\bibitem[\protect\citeauthoryear{Nguyen and Grishman}{Nguyen and
  Grishman}{2018}]%
        {DBLP:conf/aaai/NguyenG18}
\bibfield{author}{\bibinfo{person}{Thien~Huu Nguyen} {and}
  \bibinfo{person}{Ralph Grishman}.} \bibinfo{year}{2018}\natexlab{}.
\newblock \showarticletitle{Graph Convolutional Networks With Argument-Aware
  Pooling for Event Detection}. In \bibinfo{booktitle}{\emph{Proceedings of the
  Thirty-Second {AAAI} Conference on Artificial Intelligence, (AAAI-18), the
  30th innovative Applications of Artificial Intelligence (IAAI-18), and the
  8th {AAAI} Symposium on Educational Advances in Artificial Intelligence
  (EAAI-18), New Orleans, Louisiana, USA, February 2-7, 2018}}.
  \bibinfo{pages}{5900--5907}.
\newblock
\urldef\tempurl%
\url{https://www.aaai.org/ocs/index.php/AAAI/AAAI18/paper/view/16329}
\showURL{%
\tempurl}


\bibitem[\protect\citeauthoryear{Ozcanli, Yaprakdal, and Baysal}{Ozcanli
  et~al\mbox{.}}{2020}]%
        {ozcanli2020deep}
\bibfield{author}{\bibinfo{person}{Asiye~K Ozcanli}, \bibinfo{person}{Fatma
  Yaprakdal}, {and} \bibinfo{person}{Mustafa Baysal}.}
  \bibinfo{year}{2020}\natexlab{}.
\newblock \showarticletitle{Deep learning methods and applications for
  electrical power systems: A comprehensive review}.
\newblock \bibinfo{journal}{\emph{International Journal of Energy Research}}
  \bibinfo{volume}{44}, \bibinfo{number}{9} (\bibinfo{year}{2020}),
  \bibinfo{pages}{7136--7157}.
\newblock


\bibitem[\protect\citeauthoryear{Peng, Li, Gong, Song, Ning, Lai, and Yu}{Peng
  et~al\mbox{.}}{2019}]%
        {DBLP:conf/ijcai/PengLGSNLY19}
\bibfield{author}{\bibinfo{person}{Hao Peng}, \bibinfo{person}{Jianxin Li},
  \bibinfo{person}{Qiran Gong}, \bibinfo{person}{Yangqiu Song},
  \bibinfo{person}{Yuanxing Ning}, \bibinfo{person}{Kunfeng Lai}, {and}
  \bibinfo{person}{Philip~S. Yu}.} \bibinfo{year}{2019}\natexlab{}.
\newblock \showarticletitle{Fine-grained Event Categorization with
  Heterogeneous Graph Convolutional Networks}. In
  \bibinfo{booktitle}{\emph{Proceedings of the Twenty-Eighth International
  Joint Conference on Artificial Intelligence, {IJCAI} 2019, Macao, China,
  August 10-16, 2019}}. \bibinfo{pages}{3238--3245}.
\newblock
\urldef\tempurl%
\url{https://doi.org/10.24963/ijcai.2019/449}
\showDOI{\tempurl}


\bibitem[\protect\citeauthoryear{Peng, Li, Song, Yang, Ranjan, Yu, and He}{Peng
  et~al\mbox{.}}{2021a}]%
        {DBLP:journals/tkdd/PengLSYRYH21}
\bibfield{author}{\bibinfo{person}{Hao Peng}, \bibinfo{person}{Jianxin Li},
  \bibinfo{person}{Yangqiu Song}, \bibinfo{person}{Renyu Yang},
  \bibinfo{person}{Rajiv Ranjan}, \bibinfo{person}{Philip~S. Yu}, {and}
  \bibinfo{person}{Lifang He}.} \bibinfo{year}{2021}\natexlab{a}.
\newblock \showarticletitle{Streaming Social Event Detection and Evolution
  Discovery in Heterogeneous Information Networks}.
\newblock \bibinfo{journal}{\emph{{ACM} Trans. Knowl. Discov. Data}}
  \bibinfo{volume}{15}, \bibinfo{number}{5} (\bibinfo{year}{2021}),
  \bibinfo{pages}{89:1--89:33}.
\newblock
\urldef\tempurl%
\url{https://doi.org/10.1145/3447585}
\showDOI{\tempurl}


\bibitem[\protect\citeauthoryear{Peng, Li, Wang, Wang, Gong, Yang, Li, Yu, and
  He}{Peng et~al\mbox{.}}{2021b}]%
        {DBLP:journals/tkde/PengLWWGYLYH21}
\bibfield{author}{\bibinfo{person}{Hao Peng}, \bibinfo{person}{Jianxin Li},
  \bibinfo{person}{Senzhang Wang}, \bibinfo{person}{Lihong Wang},
  \bibinfo{person}{Qiran Gong}, \bibinfo{person}{Renyu Yang},
  \bibinfo{person}{Bo Li}, \bibinfo{person}{Philip~S. Yu}, {and}
  \bibinfo{person}{Lifang He}.} \bibinfo{year}{2021}\natexlab{b}.
\newblock \showarticletitle{Hierarchical Taxonomy-Aware and Attentional Graph
  Capsule RCNNs for Large-Scale Multi-Label Text Classification}.
\newblock \bibinfo{journal}{\emph{{IEEE} Trans. Knowl. Data Eng.}}
  \bibinfo{volume}{33}, \bibinfo{number}{6} (\bibinfo{year}{2021}),
  \bibinfo{pages}{2505--2519}.
\newblock
\urldef\tempurl%
\url{https://doi.org/10.1109/TKDE.2019.2959991}
\showDOI{\tempurl}


\bibitem[\protect\citeauthoryear{Peng, Zhang, Li, Cao, Pan, and Yu}{Peng
  et~al\mbox{.}}{2022}]%
        {peng2022reinforced}
\bibfield{author}{\bibinfo{person}{Hao Peng}, \bibinfo{person}{Ruitong Zhang},
  \bibinfo{person}{Shaoning Li}, \bibinfo{person}{Yuwei Cao},
  \bibinfo{person}{Shirui Pan}, {and} \bibinfo{person}{Philip Yu}.}
  \bibinfo{year}{2022}\natexlab{}.
\newblock \showarticletitle{Reinforced, incremental and cross-lingual event
  detection from social messages}.
\newblock \bibinfo{journal}{\emph{IEEE Transactions on Pattern Analysis and
  Machine Intelligence}} (\bibinfo{year}{2022}).
\newblock


\bibitem[\protect\citeauthoryear{Pennington, Socher, and Manning}{Pennington
  et~al\mbox{.}}{2014}]%
        {DBLP:conf/emnlp/PenningtonSM14}
\bibfield{author}{\bibinfo{person}{Jeffrey Pennington},
  \bibinfo{person}{Richard Socher}, {and} \bibinfo{person}{Christopher~D.
  Manning}.} \bibinfo{year}{2014}\natexlab{}.
\newblock \showarticletitle{Glove: Global Vectors for Word Representation}. In
  \bibinfo{booktitle}{\emph{Proceedings of the 2014 Conference on Empirical
  Methods in Natural Language Processing, {EMNLP} 2014, October 25-29, 2014,
  Doha, Qatar}}. \bibinfo{pages}{1532--1543}.
\newblock
\urldef\tempurl%
\url{https://doi.org/10.3115/v1/d14-1162}
\showDOI{\tempurl}


\bibitem[\protect\citeauthoryear{Ren, Jiang, Peng, Liu, Wu, and Yu}{Ren
  et~al\mbox{.}}{2022}]%
        {DBLP:journals/corr/abs-2205-12179}
\bibfield{author}{\bibinfo{person}{Jiaqian Ren}, \bibinfo{person}{Lei Jiang},
  \bibinfo{person}{Hao Peng}, \bibinfo{person}{Zhiwei Liu},
  \bibinfo{person}{Jia Wu}, {and} \bibinfo{person}{Philip~S. Yu}.}
  \bibinfo{year}{2022}\natexlab{}.
\newblock \showarticletitle{Evidential Temporal-aware Graph-based Social Event
  Detection via Dempster-Shafer Theory}.
\newblock \bibinfo{journal}{\emph{IEEE ICWS}} (\bibinfo{year}{2022}).
\newblock


\bibitem[\protect\citeauthoryear{Righetto, Martins, Carvalho, Hattori, and
  de~Francisci}{Righetto et~al\mbox{.}}{2021}]%
        {DBLP:conf/isgt/RighettoMCHF21}
\bibfield{author}{\bibinfo{person}{Sophia~Boing Righetto},
  \bibinfo{person}{Marcos Aurelio~Izumida Martins},
  \bibinfo{person}{Edgar~Gerevini Carvalho}, \bibinfo{person}{Leandro~Takeshi
  Hattori}, {and} \bibinfo{person}{Silvia de Francisci}.}
  \bibinfo{year}{2021}\natexlab{}.
\newblock \showarticletitle{Predictive Maintenance 4.0 Applied in Electrical
  Power Systems}. In \bibinfo{booktitle}{\emph{{IEEE} Power {\&} Energy Society
  Innovative Smart Grid Technologies Conference, {ISGT} 2021, Washington, DC,
  USA, February 16-18, 2021}}. \bibinfo{pages}{1--5}.
\newblock
\urldef\tempurl%
\url{https://doi.org/10.1109/ISGT49243.2021.9372230}
\showDOI{\tempurl}


\bibitem[\protect\citeauthoryear{Scarselli, Gori, Tsoi, Hagenbuchner, and
  Monfardini}{Scarselli et~al\mbox{.}}{2009}]%
        {scarselli2008graph}
\bibfield{author}{\bibinfo{person}{Franco Scarselli}, \bibinfo{person}{Marco
  Gori}, \bibinfo{person}{Ah~Chung Tsoi}, \bibinfo{person}{Markus
  Hagenbuchner}, {and} \bibinfo{person}{Gabriele Monfardini}.}
  \bibinfo{year}{2009}\natexlab{}.
\newblock \showarticletitle{The Graph Neural Network Model}.
\newblock \bibinfo{journal}{\emph{{IEEE} Trans. Neural Networks}}
  \bibinfo{volume}{20}, \bibinfo{number}{1} (\bibinfo{year}{2009}),
  \bibinfo{pages}{61--80}.
\newblock
\urldef\tempurl%
\url{https://doi.org/10.1109/TNN.2008.2005605}
\showDOI{\tempurl}


\bibitem[\protect\citeauthoryear{Schuster and Paliwal}{Schuster and
  Paliwal}{1997}]%
        {schuster1997bidirectional}
\bibfield{author}{\bibinfo{person}{Mike Schuster} {and}
  \bibinfo{person}{Kuldip~K. Paliwal}.} \bibinfo{year}{1997}\natexlab{}.
\newblock \showarticletitle{Bidirectional recurrent neural networks}.
\newblock \bibinfo{journal}{\emph{{IEEE} Trans. Signal Process.}}
  \bibinfo{volume}{45}, \bibinfo{number}{11} (\bibinfo{year}{1997}),
  \bibinfo{pages}{2673--2681}.
\newblock
\urldef\tempurl%
\url{https://doi.org/10.1109/78.650093}
\showDOI{\tempurl}


\bibitem[\protect\citeauthoryear{Shi, Hu, Zhao, and Yu}{Shi
  et~al\mbox{.}}{2019}]%
        {DBLP:journals/tkde/ShiHZY19}
\bibfield{author}{\bibinfo{person}{Chuan Shi}, \bibinfo{person}{Binbin Hu},
  \bibinfo{person}{Wayne~Xin Zhao}, {and} \bibinfo{person}{Philip~S. Yu}.}
  \bibinfo{year}{2019}\natexlab{}.
\newblock \showarticletitle{Heterogeneous Information Network Embedding for
  Recommendation}.
\newblock \bibinfo{journal}{\emph{{IEEE} Trans. Knowl. Data Eng.}}
  \bibinfo{volume}{31}, \bibinfo{number}{2} (\bibinfo{year}{2019}),
  \bibinfo{pages}{357--370}.
\newblock
\urldef\tempurl%
\url{https://doi.org/10.1109/TKDE.2018.2833443}
\showDOI{\tempurl}


\bibitem[\protect\citeauthoryear{Smith and Wedeward}{Smith and
  Wedeward}{2009}]%
        {smith2009event}
\bibfield{author}{\bibinfo{person}{Michael~J Smith} {and}
  \bibinfo{person}{Kevin Wedeward}.} \bibinfo{year}{2009}\natexlab{}.
\newblock \showarticletitle{Event detection and location in electric power
  systems using constrained optimization}. In \bibinfo{booktitle}{\emph{2009
  IEEE Power \& Energy Society General Meeting}}. IEEE, \bibinfo{pages}{1--6}.
\newblock


\bibitem[\protect\citeauthoryear{Tang, Qu, and Mei}{Tang et~al\mbox{.}}{2015}]%
        {DBLP:conf/kdd/TangQM15}
\bibfield{author}{\bibinfo{person}{Jian Tang}, \bibinfo{person}{Meng Qu}, {and}
  \bibinfo{person}{Qiaozhu Mei}.} \bibinfo{year}{2015}\natexlab{}.
\newblock \showarticletitle{{PTE:} Predictive Text Embedding through
  Large-scale Heterogeneous Text Networks}. In
  \bibinfo{booktitle}{\emph{Proceedings of the 21th {ACM} {SIGKDD}
  International Conference on Knowledge Discovery and Data Mining, Sydney, NSW,
  Australia, August 10-13, 2015}}. \bibinfo{pages}{1165--1174}.
\newblock
\urldef\tempurl%
\url{https://doi.org/10.1145/2783258.2783307}
\showDOI{\tempurl}


\bibitem[\protect\citeauthoryear{Vashishth, Sanyal, Nitin, and
  Talukdar}{Vashishth et~al\mbox{.}}{2020}]%
        {DBLP:conf/iclr/VashishthSNT20}
\bibfield{author}{\bibinfo{person}{Shikhar Vashishth}, \bibinfo{person}{Soumya
  Sanyal}, \bibinfo{person}{Vikram Nitin}, {and} \bibinfo{person}{Partha~P.
  Talukdar}.} \bibinfo{year}{2020}\natexlab{}.
\newblock \showarticletitle{Composition-based Multi-Relational Graph
  Convolutional Networks}. In \bibinfo{booktitle}{\emph{8th International
  Conference on Learning Representations, {ICLR} 2020, Addis Ababa, Ethiopia,
  April 26-30, 2020}}.
\newblock
\urldef\tempurl%
\url{https://openreview.net/forum?id=BylA\_C4tPr}
\showURL{%
\tempurl}


\bibitem[\protect\citeauthoryear{Velickovic, Cucurull, Casanova, Romero,
  Li{\`{o}}, and Bengio}{Velickovic et~al\mbox{.}}{2018}]%
        {DBLP:conf/iclr/VelickovicCCRLB18}
\bibfield{author}{\bibinfo{person}{Petar Velickovic}, \bibinfo{person}{Guillem
  Cucurull}, \bibinfo{person}{Arantxa Casanova}, \bibinfo{person}{Adriana
  Romero}, \bibinfo{person}{Pietro Li{\`{o}}}, {and} \bibinfo{person}{Yoshua
  Bengio}.} \bibinfo{year}{2018}\natexlab{}.
\newblock \showarticletitle{Graph Attention Networks}. In
  \bibinfo{booktitle}{\emph{6th International Conference on Learning
  Representations, {ICLR} 2018, Vancouver, BC, Canada, April 30 - May 3, 2018,
  Conference Track Proceedings}}.
\newblock
\urldef\tempurl%
\url{https://openreview.net/forum?id=rJXMpikCZ}
\showURL{%
\tempurl}


\bibitem[\protect\citeauthoryear{Wang, Han, Liu, Sun, and Li}{Wang
  et~al\mbox{.}}{2019a}]%
        {DBLP:conf/naacl/WangHLSL19}
\bibfield{author}{\bibinfo{person}{Xiaozhi Wang}, \bibinfo{person}{Xu Han},
  \bibinfo{person}{Zhiyuan Liu}, \bibinfo{person}{Maosong Sun}, {and}
  \bibinfo{person}{Peng Li}.} \bibinfo{year}{2019}\natexlab{a}.
\newblock \showarticletitle{Adversarial Training for Weakly Supervised Event
  Detection}. In \bibinfo{booktitle}{\emph{Proceedings of the 2019 Conference
  of the North American Chapter of the Association for Computational
  Linguistics: Human Language Technologies, {NAACL-HLT} 2019, Minneapolis, MN,
  USA, June 2-7, 2019, Volume 1 (Long and Short Papers)}}.
  \bibinfo{pages}{998--1008}.
\newblock
\urldef\tempurl%
\url{https://doi.org/10.18653/v1/n19-1105}
\showDOI{\tempurl}


\bibitem[\protect\citeauthoryear{Wang, Ji, Shi, Wang, Ye, Cui, and Yu}{Wang
  et~al\mbox{.}}{2019b}]%
        {DBLP:conf/www/WangJSWYCY19}
\bibfield{author}{\bibinfo{person}{Xiao Wang}, \bibinfo{person}{Houye Ji},
  \bibinfo{person}{Chuan Shi}, \bibinfo{person}{Bai Wang},
  \bibinfo{person}{Yanfang Ye}, \bibinfo{person}{Peng Cui}, {and}
  \bibinfo{person}{Philip~S. Yu}.} \bibinfo{year}{2019}\natexlab{b}.
\newblock \showarticletitle{Heterogeneous Graph Attention Network}. In
  \bibinfo{booktitle}{\emph{The World Wide Web Conference, {WWW} 2019, San
  Francisco, CA, USA, May 13-17, 2019}}. \bibinfo{pages}{2022--2032}.
\newblock
\urldef\tempurl%
\url{https://doi.org/10.1145/3308558.3313562}
\showDOI{\tempurl}


\bibitem[\protect\citeauthoryear{Wang, Wang, Han, Jiang, Han, Liu, Li, Li, Lin,
  and Zhou}{Wang et~al\mbox{.}}{2020}]%
        {DBLP:conf/emnlp/WangWHJHLLLLZ20}
\bibfield{author}{\bibinfo{person}{Xiaozhi Wang}, \bibinfo{person}{Ziqi Wang},
  \bibinfo{person}{Xu Han}, \bibinfo{person}{Wangyi Jiang},
  \bibinfo{person}{Rong Han}, \bibinfo{person}{Zhiyuan Liu},
  \bibinfo{person}{Juanzi Li}, \bibinfo{person}{Peng Li},
  \bibinfo{person}{Yankai Lin}, {and} \bibinfo{person}{Jie Zhou}.}
  \bibinfo{year}{2020}\natexlab{}.
\newblock \showarticletitle{{MAVEN:} {A} Massive General Domain Event Detection
  Dataset}. In \bibinfo{booktitle}{\emph{Proceedings of the 2020 Conference on
  Empirical Methods in Natural Language Processing, {EMNLP} 2020, Online,
  November 16-20, 2020}}. \bibinfo{pages}{1652--1671}.
\newblock
\urldef\tempurl%
\url{https://doi.org/10.18653/v1/2020.emnlp-main.129}
\showDOI{\tempurl}


\bibitem[\protect\citeauthoryear{Wu, Pan, Chen, Long, Zhang, and Yu}{Wu
  et~al\mbox{.}}{2021}]%
        {DBLP:journals/tnn/WuPCLZY21}
\bibfield{author}{\bibinfo{person}{Zonghan Wu}, \bibinfo{person}{Shirui Pan},
  \bibinfo{person}{Fengwen Chen}, \bibinfo{person}{Guodong Long},
  \bibinfo{person}{Chengqi Zhang}, {and} \bibinfo{person}{Philip~S. Yu}.}
  \bibinfo{year}{2021}\natexlab{}.
\newblock \showarticletitle{A Comprehensive Survey on Graph Neural Networks}.
\newblock \bibinfo{journal}{\emph{{IEEE} Trans. Neural Networks Learn. Syst.}}
  \bibinfo{volume}{32}, \bibinfo{number}{1} (\bibinfo{year}{2021}),
  \bibinfo{pages}{4--24}.
\newblock
\urldef\tempurl%
\url{https://doi.org/10.1109/TNNLS.2020.2978386}
\showDOI{\tempurl}


\bibitem[\protect\citeauthoryear{Xie, Sun, Zhou, Qu, and Dai}{Xie
  et~al\mbox{.}}{2021}]%
        {DBLP:conf/acl/XieSZQD21}
\bibfield{author}{\bibinfo{person}{Jianye Xie}, \bibinfo{person}{Haotong Sun},
  \bibinfo{person}{Junsheng Zhou}, \bibinfo{person}{Weiguang Qu}, {and}
  \bibinfo{person}{Xinyu Dai}.} \bibinfo{year}{2021}\natexlab{}.
\newblock \showarticletitle{Event Detection as Graph Parsing}. In
  \bibinfo{booktitle}{\emph{Findings of the Association for Computational
  Linguistics: {ACL/IJCNLP} 2021, Online Event, August 1-6, 2021}}
  \emph{(\bibinfo{series}{Findings of {ACL}},
  Vol.~\bibinfo{volume}{{ACL/IJCNLP} 2021})}. \bibinfo{pages}{1630--1640}.
\newblock
\urldef\tempurl%
\url{https://doi.org/10.18653/v1/2021.findings-acl.142}
\showDOI{\tempurl}


\bibitem[\protect\citeauthoryear{Yan, Jin, Meng, Guo, and Cheng}{Yan
  et~al\mbox{.}}{2019}]%
        {DBLP:conf/emnlp/YanJMGC19}
\bibfield{author}{\bibinfo{person}{Haoran Yan}, \bibinfo{person}{Xiaolong Jin},
  \bibinfo{person}{Xiangbin Meng}, \bibinfo{person}{Jiafeng Guo}, {and}
  \bibinfo{person}{Xueqi Cheng}.} \bibinfo{year}{2019}\natexlab{}.
\newblock \showarticletitle{Event Detection with Multi-Order Graph Convolution
  and Aggregated Attention}. In \bibinfo{booktitle}{\emph{Proceedings of the
  2019 Conference on Empirical Methods in Natural Language Processing and the
  9th International Joint Conference on Natural Language Processing,
  {EMNLP-IJCNLP} 2019, Hong Kong, China, November 3-7, 2019}}.
  \bibinfo{pages}{5765--5769}.
\newblock
\urldef\tempurl%
\url{https://doi.org/10.18653/v1/D19-1582}
\showDOI{\tempurl}


\bibitem[\protect\citeauthoryear{Yun, Jeong, Kim, Kang, and Kim}{Yun
  et~al\mbox{.}}{2019}]%
        {yun2019graph}
\bibfield{author}{\bibinfo{person}{Seongjun Yun}, \bibinfo{person}{Minbyul
  Jeong}, \bibinfo{person}{Raehyun Kim}, \bibinfo{person}{Jaewoo Kang}, {and}
  \bibinfo{person}{Hyunwoo~J. Kim}.} \bibinfo{year}{2019}\natexlab{}.
\newblock \showarticletitle{Graph Transformer Networks}.
\newblock  (\bibinfo{year}{2019}), \bibinfo{pages}{11960--11970}.
\newblock
\urldef\tempurl%
\url{https://proceedings.neurips.cc/paper/2019/hash/9d63484abb477c97640154d40595a3bb-Abstract.html}
\showURL{%
\tempurl}


\bibitem[\protect\citeauthoryear{Zhang and Xu}{Zhang and Xu}{2021}]%
        {DBLP:journals/bigdatama/ZhangX21}
\bibfield{author}{\bibinfo{person}{Jintao Zhang} {and} \bibinfo{person}{Quan
  Xu}.} \bibinfo{year}{2021}\natexlab{}.
\newblock \showarticletitle{Attention-aware heterogeneous graph neural
  network}.
\newblock \bibinfo{journal}{\emph{Big Data Min. Anal.}} \bibinfo{volume}{4},
  \bibinfo{number}{4} (\bibinfo{year}{2021}), \bibinfo{pages}{233--241}.
\newblock
\urldef\tempurl%
\url{https://doi.org/10.26599/BDMA.2021.9020008}
\showDOI{\tempurl}


\bibitem[\protect\citeauthoryear{Zheng, Cai, Chen, Lei, and Chen}{Zheng
  et~al\mbox{.}}{2021}]%
        {DBLP:conf/www/ZhengCCLC21}
\bibfield{author}{\bibinfo{person}{Jianming Zheng}, \bibinfo{person}{Fei Cai},
  \bibinfo{person}{Wanyu Chen}, \bibinfo{person}{Wengqiang Lei}, {and}
  \bibinfo{person}{Honghui Chen}.} \bibinfo{year}{2021}\natexlab{}.
\newblock \showarticletitle{Taxonomy-aware Learning for Few-Shot Event
  Detection}. In \bibinfo{booktitle}{\emph{{WWW} '21: The Web Conference 2021,
  Virtual Event / Ljubljana, Slovenia, April 19-23, 2021}}.
  \bibinfo{pages}{3546--3557}.
\newblock
\urldef\tempurl%
\url{https://doi.org/10.1145/3442381.3449949}
\showDOI{\tempurl}


\bibitem[\protect\citeauthoryear{Zhou, Cui, Hu, Zhang, Yang, Liu, Wang, Li, and
  Sun}{Zhou et~al\mbox{.}}{2020}]%
        {zhou2020graph}
\bibfield{author}{\bibinfo{person}{Jie Zhou}, \bibinfo{person}{Ganqu Cui},
  \bibinfo{person}{Shengding Hu}, \bibinfo{person}{Zhengyan Zhang},
  \bibinfo{person}{Cheng Yang}, \bibinfo{person}{Zhiyuan Liu},
  \bibinfo{person}{Lifeng Wang}, \bibinfo{person}{Changcheng Li}, {and}
  \bibinfo{person}{Maosong Sun}.} \bibinfo{year}{2020}\natexlab{}.
\newblock \showarticletitle{Graph neural networks: {A} review of methods and
  applications}.
\newblock \bibinfo{journal}{\emph{{AI} Open}}  \bibinfo{volume}{1}
  (\bibinfo{year}{2020}), \bibinfo{pages}{57--81}.
\newblock
\urldef\tempurl%
\url{https://doi.org/10.1016/j.aiopen.2021.01.001}
\showDOI{\tempurl}


\bibitem[\protect\citeauthoryear{Zhou, Arghandeh, and Spanos}{Zhou
  et~al\mbox{.}}{2018}]%
        {DBLP:journals/tsg/ZhouAS18}
\bibfield{author}{\bibinfo{person}{Yuxun Zhou}, \bibinfo{person}{Reza
  Arghandeh}, {and} \bibinfo{person}{Costas~J. Spanos}.}
  \bibinfo{year}{2018}\natexlab{}.
\newblock \showarticletitle{Partial Knowledge Data-Driven Event Detection for
  Power Distribution Networks}.
\newblock \bibinfo{journal}{\emph{{IEEE} Trans. Smart Grid}}
  \bibinfo{volume}{9}, \bibinfo{number}{5} (\bibinfo{year}{2018}),
  \bibinfo{pages}{5152--5162}.
\newblock
\urldef\tempurl%
\url{https://doi.org/10.1109/TSG.2017.2681962}
\showDOI{\tempurl}


\end{thebibliography}

\appendix

\end{document}